\documentclass{acmart}

\usepackage{amsmath}
\usepackage{bbm}
\usepackage[thinc]{esdiff}
\usepackage{enumitem,kantlipsum}
\usepackage{longtable}
\usepackage{booktabs} 
\usepackage{makecell}
\usepackage{xspace}
\usepackage{fdsymbol}
\usepackage{wrapfig}

 \usepackage[suppress]{color-edits}

\addauthor{BH}{blue}
\addauthor{negar}{red}
\addauthor{christina}{green}
\addauthor{VP}{purple}

\newcommand{\classsim}{{\sc Overlap}\xspace}
\newcommand{\bleu}{\textsc{Bleu}\xspace}
\newcommand{\rouge}{\textsc{Rouge}\xspace}
\newcommand{\dice}{Dice\xspace}
\newcommand{\iid}{I.I.D.\xspace}
\newcommand{\etal}{\textit{et al}.\xspace}

\newcommand{\evaluationGaps}{Evaluation Gaps\xspace}
\newcommand{\evaluationGap}{Evaluation Gap\xspace}

\newcommand{\evaluationgaps}{evaluation gaps\xspace}
\newcommand{\evaluationgappar}[1]{\textbf{#1}. }
\newcommand{\assumption}[1]{\textsc{#1}. }

\newcommand{\papercount}{{\color{black}200}\xspace}

\AtBeginDocument{%
  \providecommand\BibTeX{{%
    \normalfont B\kern-0.5em{\scshape i\kern-0.25em b}\kern-0.8em\TeX}}}

\copyrightyear{2022}
\acmYear{2022}
\setcopyright{rightsretained}
\acmConference[FAccT '22]{2022 ACM Conference on Fairness, Accountability, and Transparency}{June 21--24, 2022}{Seoul, Republic of Korea}
\acmBooktitle{2022 ACM Conference on Fairness, Accountability, and Transparency (FAccT '22), June 21--24, 2022, Seoul, Republic of Korea}\acmDOI{10.1145/3531146.3533233}
\acmISBN{978-1-4503-9352-2/22/06}

\begin{document}

\title{\evaluationGaps in Machine Learning Practice}


\author{Ben Hutchinson}
    \affiliation{
    \institution{Google Research}
    \city{Sydney}
    \country{Australia}}
    \email{benhutch@google.com}
\author{Negar Rostamzadeh}
    \affiliation{
    \institution{Google Research}
    \city{Montreal}
    \country{Canada}} 
    \email{nrostamzadeh@google.com} 
\author{Christina Greer}
    \affiliation{
    \institution{Google Research}
    \city{Mountain View}
    \country{USA}} 
    \email{ckuhn@google.com} 
\author{Katherine Heller}
    \affiliation{
    \institution{Google Research}
    \city{Mountain View}
    \country{USA}} 
    \email{heller@google.com} 
\author{Vinodkumar Prabhakaran}
    \affiliation{
    \institution{Google Research}
    \city{San Francisco}
    \country{USA}} 
    \email{vinodkpg@google.com} 
  
  

\renewcommand{\shortauthors}{Hutchinson, Rostamzadeh, Greer, Heller, and Prabhakaran}

\begin{abstract}
Forming a reliable judgement of a machine learning (ML) model's appropriateness for an application ecosystem is critical for its responsible use, and requires considering a broad range of factors including harms, benefits, and responsibilities. In practice, however, evaluations of ML models frequently focus on only a narrow range of decontextualized predictive behaviours. We examine the \evaluationgaps between the idealized breadth of evaluation concerns and the observed narrow focus of actual evaluations.
Through an empirical study of papers from recent high-profile conferences in the Computer Vision and Natural Language Processing communities, we demonstrate a general focus on a handful of evaluation methods. By considering the metrics and test data distributions used in these methods, we draw attention to which properties of models are centered in the field, revealing the properties that are frequently neglected or sidelined during evaluation. By studying these properties, we demonstrate the machine learning discipline's implicit assumption of a range of commitments which have normative impacts; these include commitments to consequentialism, abstractability from context, the quantifiability of impacts, the limited role of model inputs in evaluation, and the equivalence of different failure modes. Shedding light on these assumptions enables us to question their appropriateness for ML system contexts, pointing the way towards more  contextualized evaluation methodologies for robustly examining the trustworthiness of ML models.
\end{abstract}

\ccsdesc[500]{Computing methodologies~Machine learning approaches}

\keywords{machine learning, evaluation, applications}


\maketitle

\section{Introduction}
\label{sec:introduction}

When evaluating a machine learning (ML) model for real-world uses, two fundamental questions arise: \emph{Is this ML model good (enough)?} and \emph{Is this ML model better than some alternative?}
Obtaining reliable answers to these questions can be consequential for safety, fairness, and justice concerns in the deployment ecosystems. To address such questions, model evaluations use a variety of methods, and in doing so make technical and normative assumptions that are not always explicit.
\BHdelete{For example, when striving to answer whether a model is desirable (or \emph{good} in the normative sense), evaluations typically focus largely on what is easily measurable (a reductive technical notion of \emph{good}).}
These implicit assumptions can obscure the presence of epistemic gaps and motivations in the model evaluations, which, if not identified, constitute risky \emph{unknown unknowns}.  

Recent scholarship has critiqued the ML community's evaluation practices, focusing on the use of evaluation benchmarks and leaderboards.
Although leaderboards support the need of the discipline to iteratively optimize for accuracy, they neglect concerns such as inference latency, robustness, and externalities \cite{ethayarajh2020utility}.
The structural incentives of the ``competition mindset'' encouraged by leaderboards can pose challenges to empirical rigor \cite{sculley2018winner}.  
For example, over-reliance on a small number of evaluation metrics can lead to gaming the metric (cf.\ Goodhart's Law ``when a measure becomes a target, it ceases to be a good measure'') \cite{thomas2020reliance}; this can  happen unintentionally as researchers pursue models with ``state of the art'' performance.
Benchmarks that encourage narrowly optimizing for test set accuracy can also lead to models relying on spurious signals \cite{carter2021overinterpretation}, while neglecting the challenge of measuring the full range of likely harms \cite{bowman2021will}. 
Birhane \etal find evidence for this in their study of the discourse of ML papers, showing that the field centers accuracy, generalization, and novelty, while marginalizing values such as safety \cite{birhane2021values}.
Given that benchmark evaluations serve as proxies for performance on underlying abstract tasks \cite{schlangen2021targeting}, evaluating against a range of diverse benchmarks for each task might help mitigate biases within each benchmark.
However, ML research disciplines seem to be trending towards relying on fewer evaluation benchmark datasets \cite{koch2021reduced},
with test set reuse potentially leading to a research community's overfitting with respect to the general task \cite{zhang2020machine,liao2021we}.
Furthermore, within each benchmark, items are weighted equally (thus focusing on the head of the data distribution), failing to capture inherent differences in difficulty across items, and hence providing poor measures of progress on task performance \cite{rodriguez-etal-2021-evaluation}.
As Raji \etal point out, the ML research discipline's decontextualized and non-systematic use of benchmark data raises serious issues with regards to the  validity of benchmarks as measures of progress on general task performance \cite{raji2021ai}.  

This paper complements and extends this range of critiques, considering the risks of application developers adopting the ML research community's standard evaluation methodologies. We seek to address challenges in measuring technology readiness (\textsc{tram}) \cite{lin2007integrating,rismani2021how}, while acknowledging this cannot be reduced to a purely technical question \cite{dahlin2021mind, rismani2021how}. By studying and analyzing the ML research community's evaluation practices, we draw attention to the \emph{\evaluationgaps} between ideal theories of evaluation and what is observed in ML research. 
By considering aspects of \emph{evaluation data} and \emph{evaluation metrics}---as well as considerations of \emph{evaluation practices} such as error analysis and reporting of error bars---we highlight the discrepancies between the model quality signals reported by the research community and what is relevant to real-world model use.
Our framework for analyzing the gaps
builds upon and complements other streams of work on ML evaluation practices, including addressing distribution shifts between development data and application data
\cite{sugiyama2007direct,koh2020wilds,chen2021mandoline}, and robustness to perturbations in test items \cite{prabhakaran2019perturbation,moradi2021evaluating,winkens2020contrastive,hendrycks2019benchmarking}. 
We situate this work alongside studies of the appropriateness of ML evaluation metrics (e.g., \cite{japkowicz2006question, derczynski2016complementarity,zhang2020machine}), noting that reliable choice of metric is often hampered by unclear goals \cite{kuwajima2020engineering,d2020underspecification}. 
In foregrounding the information needs of application developers, we are also aligned with calls for transparent reporting of ML model evaluations  \cite{mitchell2019model}, prioritizing needs of ML fairness practitioners \cite{holstein2019improving}, model auditing practices \cite{raji2020closing}, and robust practices for evaluating ML systems for production readiness \cite{breck2017ml}.

In Section~2, we consider various ideal goals that motivate why ML models are evaluated, discussing how these goals can differ between research contexts and application contexts.
We then report in Section~3  on an empirical study into how machine learning research communities report model evaluations.
By comparing the ideal goals of evaluation with the observed evaluation trends in our study, we highlight in Section~4 the \evaluationgaps that present challenges to evaluations  being good proxies for what application developers really care about. We identify six implicit evaluation assumptions that could account for the presence of these gaps. Finally, in Section~5, we discuss various techniques and methodologies that may help to mitigate these gaps.

\section{Ideals of ML Model Evaluation}
\label{sec:ideals}

\BHcomment{D's comment: Overall, section 2 feels generally true and correct to me, but also a little bit hand-wavy. Can we }

\begin{quote}
\emph{Far better an approximate answer to the right question, which is often vague, than an exact answer to the wrong question, which can always be made precise.} ---
John Tukey \cite[pp.\ 13--14]{tukey1962future}
\end{quote}

Although this paper is ultimately concerned with practical information needs when evaluating ML models for use in applications, it is useful to first step back and consider the ultimate motivations and goals of model evaluation. To evaluate is to form a judgement; however, asking \emph{Is this a good ML model?} is akin to asking such a question of other artefacts---such as \emph{Is this a good glass?}---in that it requires acknowledging the implicit semantic arguments of uses and goals \cite{pustejovsky1998generative}. For example, \emph{Is this a good glass [for my toddler to drink from, given that I want to avoid broken glass]?} is a very different question from \emph{Is this a good glass [in which to serve wine to my boss, given that I want to impress them]?}

In this paper, we will speak of a \emph{model evaluation} as a system of arbitrary structure that takes a model as an input and produces outputs of some form to judge the model. Designing a model evaluation often involves choosing one or more evaluation metrics (such as accuracy) combined with a choice of test data. The evaluation might be motivated by various stakeholder perspectives and interests \cite{jones1995evaluating}. The output might, for example, produce a single metric and an associated numeric value, or a table of such metrics and values; it might include confidence intervals and significance tests on metric values; and it might include text. By producing such an output, the evaluation helps to enable transparency by reducing the number of both unknown unknowns and known unknowns.

\begin{figure}
    \centering
    \includegraphics[width=11cm]{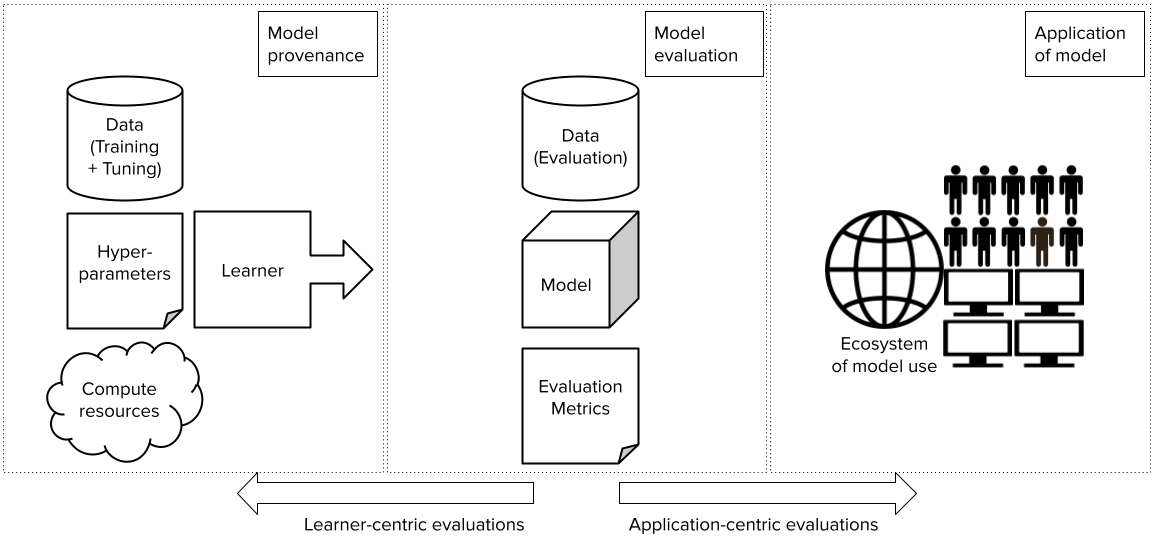}
    \caption{Learner-centric ML model evaluations are concerned with the learner and its environment. Application-centric model evaluations are concerned with how the model will interact with an ecosystem into which it is introduced.}
    \label{fig:my_label}
\end{figure}

For the purposes of this paper, it is useful to distinguish between two types of evaluations:

\begin{enumerate}[wide, labelwidth=!, labelindent=0pt]
    \item[\emph{Learner-centric}.] An ML model evaluation system useful for evaluating the learner (i.e., machine learning algorithm).
    \item[\emph{Application-centric}.] An ML model evaluation system useful for evaluating a potential application.
\end{enumerate}
Learner-centric evaluations make conclusions about the quality of the learner or its environment based on the evaluation of the learned model.
These including evaluations motivated by novel learning algorithms or model architectures, but also ones that a) aim to shed light on the training data (for example ML model evaluations can shed light on the data-generation practices used by institutions \cite{andrus2019towards}), or b) ``Green AI'' explorations of how the learner can efficiently use limited amounts of resources \cite{schwartz2020green}. 
However, when we evaluate a model without a specific application in mind, we lose the opportunity to form judgements specific to a use case. 
On the other hand, application-centric evaluations are concerned with how the model will operate within an ecosystem consisting of both human agents and technical components (Figure~\ref{fig:my_label}), sometimes described as the ``ecological validity''  \cite{de2020towards}. Applications often use scores output by the model to initiate discrete actions or decisions, by applying a specific classification threshold to the scores.\footnote{The history of this type of use case extends beyond ML models, e.g., to the use of regression models in university admissions testing \cite{hutchinson201950}.} In contrast, learner-centric evaluations sometimes care about scores output by models even in the absence of any thresholds.

This distinction between learner-centric and application-centric is related (albeit imperfectly) to the different objectives of model evaluations that concern the engineering and science disciplines \cite{wallach2018computational, mazzocchi2015could}. Note that we are not claiming (cf.\ the debate in \cite{norvig2017chomsky}) that science lies outside the bounds of statistical/ML methods, but rather that scientific-flavored pursuits have distinct uses of such methods \cite{breiman2001statistical}. Debates between AI practitioners about the relationships between AI, science, and statistical methods have a long history, for example Diana Forsythe's studies of 1980s AI labs \cite{forsythe2001invents}.
Important to this debate regarding the scientific goals of ML is the question of construct validity; that is, whether our measurements actually measure the things that we claim they do \cite{jacobs2020meaning,raji2021ai,jacobs2021measurement}. Conversely, consequential validity---which includes the real-world consequences of an evaluation's interpretation and use---is likely more important to considerations of accountability and governance of ML models in applications \cite{jacobs2021measurement}.

\begin{enumerate}[wide, labelwidth=!, labelindent=0pt]
    \item[\emph{Scientific goal}.] Evaluating the model can motivate beliefs/explanations about the world (including possibly the learner).
    \item[\emph{Engineering goal}.] Evaluating the model can tell us whether the model can be used as a means towards a goal.
\end{enumerate}
This distinction is closely related to one between ``scientific testing'' and ``competitive testing'' made by Hooker in 1995, who takes the position that competitive testing a) is unscientific, and b) does not constitute true research but merely development \cite{hooker1995testing}. However, since engineering research has its own goals, distinct from those of science \cite{bulleit2015philosophy}, a more defensible position is that evaluations in support of scientific research are distinct from evaluations in support of engineering research.

\begin{table}[]
    \centering
    \begin{tabular}{lll}\toprule
&Learner-centric evaluations&
Application-centric evaluations\\
\midrule
Typical evaluation goal&
Distinguish better learners from poorer ones&
Predict ecosystem outcomes\\
Schematic of goal&
$Understand(Learner)$&
$Understand(Ecosystem + Model)$\\
Disciplinary goals&
Science or engineering&
Primarily engineering\\
\bottomrule
    \end{tabular}
    \caption{Summary of typical goals of the idealized learner-centric and application-centric evaluations.
    }
    \label{tab:goals}
\end{table}

Table~\ref{tab:goals} summarizes the above distinctions and the relationships between them.
The distinction between learner-centric and application-centric evaluations relates to the question of internal validity and external validity that is more commonly discussed in the social sciences than in ML (see, e.g., \cite{olteanu2019social}) but also sometimes in ML \cite{liao2021we}. This is reflected in the ways in which practitioners of the two types of evaluations discuss the topic of robustness. Learner-centric evaluations pay attention to the robustness of the learner to changes in the training data (e.g., distributional shifts, outliers, perturbations, poisoning attacks; and with connections to robust estimation of statistics \cite{lecue2020robust}), while application-centric evaluations pay attention to desired behaviors such as the (in)sensitivity of the model to certain classes of perturbations of the input, or to sensitive input features (e.g., \cite{garg2019counterfactual}).

Note that nothing in the ideals of evaluation described above has stipulated whether evaluations are \emph{quantitative} or \emph{qualitative}. For example, one could imagine interrogating a chatbot model using qualitative techniques, or adopting methodologies of political critique such as \cite{crawford2021excavating}. Similarly, nothing has stipulated what combinations of empirical or deductive methods are used.

\section{ML Model Evaluations in Practice}
\label{sec:practice}

\begin{quote}
    \emph{Beneath the technical issues lie some differences in \emph{values} concerning not only the meaning but also the relative merit of ``science'' and ``artificial intelligence.''} --- Diana Forsythe \cite{forsythe2001invents}
\end{quote}

To shed light on the ML research community's norms and values around model evaluation, we looked at how these communities report their model evaluations. By examining \papercount papers from several top conferences in two research disciplines that use ML approaches extensively, we identified patterns regarding choices of metrics, evaluation data, and measurement practices. This empirical study of ML research practices complements several recent studies of ML evaluation practices. These include: a survey 144 research papers studying the properties of  models that are tested for  \cite{zhang2020machine};
a review of 107 papers from Computer Vision (CV), Natural Language Processing (NLP) and other ML disciplines to diagnose internal and external modes of evaluation failures \cite{liao2021we}; an analysis of whether 60 NLP and CV papers pay attention to accuracy or efficiency \cite{schwartz2020green}; and an analysis of the Papers With Code dataset\footnote{\url{https://paperswithcode.com}} for patterns of benchmark dataset creation and re-use \cite{koch2021reduced}. 

\subsection{Method}
\subsubsection{Data}
We sampled 200 research papers, stratified by discipline, conference and year. 100 papers were selected from each of the NLP and CV  disciplines. We selected 20 papers from the proceedings of each of the 55th to 59th Annual Meetings of the Association of Computational Linguistics (ACL'2017--ACL'2021), 25 papers at random from each of the proceedings of the 2019--2021 IEEE Conferences on Computer Vision and Pattern Recognition (CVPR'2019--CVPR'2021), and 25 papers from the 24th International Conference on Medical Image Computing and Computer Assisted Intervention (MICCAI'2021). These conferences represent the pinnacles of their respective research fields.\footnote{ACL and CVPR are rated A$^*$ (``flagship conference''), and MICCAI is rated A (``excellent conference''), by \url{core.edu.au}; all three are in the top 30 computer science conferences out of over 900 listed on \url{research.com}.}

\subsubsection{Analysis}

The authors of this paper performed this analysis, dividing the papers among themselves based on disciplinary familiarity. Using an iterative procedure of analysis and discussion, we converged on a set of labels that captured important aspects of evaluations across and within disciplines.
Recall from Section~\ref{sec:ideals} that, for our purposes, a single evaluation typically involves choosing one or more metrics and one or more datasets.
We coded each of the papers along three dimensions.
    a) \textbf{Metrics}: Which evaluation metrics were reported? After iteration, we converged on the categories of metrics shown in Table~\ref{tab:metric categories}.
    b) \textbf{Data}: Was test data drawn from the same distribution as the training data, under the Independent and Identically Distributed (\iid) assumption?
    c) \textbf{Analysis}: Was statistical significance of differences reported? Were error bars and/or confidence intervals reported? Was error analysis performed? Were examples of model performance provided to complement measurements with qualitative information?

\subsection{Results}

Although each of the disciplines and conferences does not define itself solely in terms of ML, the practice of reporting one or more model evaluations in a research paper is ubiquitous. Only five  papers did not include evaluations of ML models; of these two were published at ACL (a survey paper, a paper aimed at understanding linguistic features, and one on spanning-tree algorithms), and two at CVPR (a paper with only qualitative results, and one introducing a dataset).
Table~\ref{tab:results} summarizes the results of the other 195. Counts are non-exclusive, for example papers frequently reported multiple metrics and sometimes reported performance both on \iid test data and on non-\iid test data.

Appendix\ B contains an overview of the flavors of test data we observed.
We found  evidence to support the claim that evaluations of NLP models have ``historically involved reporting the performance (generally meaning the accuracy) of the model on a specific held-out [i.e., I.I.D.] test set''  \cite[p.\ 94]{bommasani2021opportunities}.\footnote{Two observed non-\iid evaluation patterns in NLP were: a) testing on a different linguistic ``domain'' (e.g., training on texts about earthquakes and testing on  texts about floods \cite{alam2018domain}); and b) testing a model's ability to predict properties of a manually compiled lexical resource (e.g., \cite{ustalov2017watset}). See also Appendix~B.}
CV evaluations seem to be even more likely to utilize \iid test data, and---consistent with \cite{koch2021reduced}---CV papers typically either introduce a new task (and corresponding benchmark dataset) \cite{jafarian2021learning,li2016tgif,rostamzadeh2018fashion,wu2021fashion} or present results of a new model on an existing widely-used benchmark \cite{ren2015faster,he2016deep}. An exception to this trend was CV papers which explored shared representations (e.g., in multi-task learning \cite{lacoste2018uncertainty,evci2021head2toe} or domain adaptation \cite{pinheiro2019domain,murez2018image}).

Evaluations in both disciplines showed a heavy reliance on reporting point estimates of metrics, with variance or error bars typically not reported in our sample. 
While colloquial uses of phrases like ``significantly better'' were fairly common, most papers did not report on technical calculations of statistical differences; we considered only those latter instances when coding whether a paper reported significance.
Regarding metrics, most of those that were frequently seen in our sample were somewhat insensitive to different types of errors. For example, accuracy does not distinguish between FP and FN; $F_1$ is symmetric in FP and FN (they can be swapped without affecting $F_1$); the \textsc{Overlap} metrics are similary invariant to swapping of the predicted bounding box and the reference bounding box; the \textsc{Distance} category of metrics does not distinguish over-estimation from under-estimation on regression tasks.

From our reading of the 200 papers in our sample, one qualitative observation we had was that model evaluations typically do not include concrete examples of model behavior, nor analyses of errors (for a counterexample which includes these practices, see \cite{chiril2020he}). Also, we noted the scarcity of papers whose sole contribution is a new dataset for an existing task, aligning with previous observations that dataset contributions are not valued highly within the community \cite{sambasivan2021everyone}. We hypothesise that conference reviewers place emphasis on novelty of model, task, and/or metric. We note
a general tension between disciplinary values of task novelty and demonstrating state-of-the-art performance by outperforming previous models, and the risk of overfitting from test set re-use discussed by \cite{liao2021we}.

\begin{table}
    \centering
    \begin{tabular}{p{0.14\textwidth}p{0.19\textwidth}p{0.6\textwidth}}\toprule
         Metric category&Examples&Description \\\toprule
         \sc Accuracy &Accuracy, error rate &Sensitive to the sum TP+TN and to N. Not sensitive to class imbalance. \\
         \sc Precision &Precision, \bleu &Sensitive to TP and FP. Not sensitive to FN or TN. \\
         \sc Recall & Recall, \rouge &Sensitive to TP and FN. Not sensitive to FP or TN. \\
         \sc F-score & $F_1$,  $F_\beta$&Sensitive to TP, FP and FN. Not sensitive to TN. \\
         \classsim & Dice, IoU &Sensitive to intersection and overlap of predicted and actual. \\
         \sc Likelihood &Perplexity &Sensitive to the probability that the model assigns to the test data. \\
         \sc Distance & MSE, MAE, RMSE, CD &Sensitive to the distance between the prediction and the actual value.\\
         \sc Correlation &\makecell[tl]{Pearsons $r$,\\ Spearman's $\rho$}& Sensitive to each of TP, TN, FP and FN, but unlike \textsc{Accuracy} metrics they factor in the degree of agreement that would be expected by chance.\\
         \sc AUC & MAP, AUROC& Does not rely on a specific classification threshold, but instead calculates the area under a curve parameterized by different thresholds.\\\bottomrule
    \end{tabular}
    \caption{Categories of evaluation metrics used in the analysis of the ML research literature. TP=true positives; TN=true negatives; FP=false positives; FN=false negatives; N=total number of data points. See Appendix~A for the most common metrics in our data and their categorizations.}
    \label{tab:metric categories}
\end{table}

\begin{table}[]
\small
    \centering
    \begin{tabular}{p{2.9cm}p{0.12\textwidth}p{0.12\textwidth}p{0.12\textwidth}p{0.12\textwidth}p{0.195\textwidth}}\toprule
        Discipline:Venue                          &NLP:ACL        &CV:CVPR       &CV:MICCAI     &CV:Combined     &NLP+CV:Combined  \\
        (\# papers with ML evals)                          &(97)        &(73)       &(25)     &(98)     &(195) \\
        \midrule
        \bf Most Common  Metrics         &&&&& \\
        \makecell[tl]{Metric category$^\clubsuit$\\        (num.\ of papers)}
        &\sc \makecell[tl]{Accuracy (47)\\ F-score (45) \\ Precision (43)   \\ Recall (25) }
        &\sc \makecell[tl]{AUC (32)     \\ Accuracy (25)\\ Overlap (22)     \\ Distance (10) }
        &\sc \makecell[tl]{Distance (14)\\ Overlap (9)  \\ AUC (6)          \\ Accuracy (4)}
        &\sc \makecell[tl]{AUC (38)     \\ Overlap (31) \\ Accuracy (29)    \\ Distance (24)}
        &\sc \makecell[tl]{Accuracy (76)\\ F-score + Overlap$^\spadesuit$ (74) \\ Precision (48) \\ AUC (44)} 
\\\midrule
        \bf Data         &&&&& \\
        \iid test data   & 78    &  72 & 25 &97 &175 \\
        Non-\iid test data   &28     &21   & 4 & 25 & 53 \\\midrule
        \bf  Analysis         &&&&& \\
        Reports significance        & 24    &0          &7  & 7 & 31\\
        Reports error bars$^\diamondsuit$  & 10    &6          &10  & 16 & 26\\
\bottomrule
    \end{tabular}
    \caption{Analysis of how Natural Language Processing (NLP) and Computer Vision (CV) research communities perform ML model evaluations. $^\clubsuit$Appendix\ A provides definitions of commonly observed metrics, and their mappings to categories. 
    $^\diamondsuit$Includes any form of error bars/confidence intervals/credible intervals/variation across multiple runs. 
    $^\spadesuit$Reported together here due to the equivalence of the Dice measure (in the \textsc{Overlap} category) and $F_1$ (in the \textsc{$F$-score} category) \cite{powers2011evaluation}. }
    \label{tab:results}
    \normalsize
\end{table}


\subsection{Discussion}
This small-scale quantitative study of model evaluations provides clues as to the values and goals of the ML research communities. Test data was often old (e.g., the CONLL 2003 English NER dataset \cite{sang2003introduction} used in two papers); optimizing for these static test sets fails to account for societal and linguistic change \cite{bender2021dangers}. Disaggregation of metrics was rare, and fairness analyses were absent despite our sample being from 2017 onward, concurrent with mainstream awareness of ML fairness concerns. Despite being acknowledged by influential thought-leaders in ML to be unrealistic for applications \cite{bengio2021deep}, using \iid test data is the norm. These are in alignment with the learner-centric goals of evaluations (Section~2). Similarly, with a few exceptions in our sample, there was general paucity of discussions of tradeoffs such as accuracy vs resource-efficiency that are typical of engineering disciplines \cite{bulleit2015philosophy}, suggesting that the ML research disciplines generally aspire to scientific goals concerning understanding and explaining the learner. With this lens, the disciplinary paradigm of measuring accuracy on \iid test data is not surprising: the goal is to assess a model's ability to generalize. This assessment would then give us good guarantees on the application's behavior, if the practical challenges of ascertaining the data distributions in an application ecosystem can be overcome. In practice, however, these challenges can be severe, and the research papers we surveyed do not generally tackle questions of uncertainty regarding data distributions.


\BHcomment{Compare and contrast with \url{https://github.com/hoya012/CVPR-2020-Paper-Statistics}}

\BHcomment{To Negar's point, cite a few good papers who have good practices that are worth mentioning}
\negarcomment{adding papers that have more comprehensive testing 1-2 papers in each, and highlighting them - not best models, stating the most important testing practices}


\section{Gaps and Assumptions in Common Evaluation Practices}
\label{sec:gaps}

\begin{quote}
    \emph{In theory there is no difference between theory and practice, while in practice there is.} --- Brewster (1881) \cite{brewster1881}
\end{quote}

We now consider whether the research evaluation practices observed in Section~\ref{sec:practice} are aligned with the needs of decision-makers who consider whether to use a model in an application. That is, we consider whether the typically learner-centric evaluations, which commonly use metrics such as accuracy or $F_1$ on test data \iid with the training data, meet the need of application-centric evaluations.
In doing so, we expose, in a novel way, the interplay of technical and normative considerations in model evaluation methodologies.

\subsection{Assumptions in Model Evaluation}


We introduce six assumptions in turn, describing both how they operate individually in evaluations and how they compose and compound. We also call out ``\evaluationgaps'' of concern relevant to each assumption.
Appendix\ C contains a hypothetical example from a specific application domain that illustrates the flavors of the concerns.
Our starting point is the observation from Section~\ref{sec:ideals} that the goal of application-centric model evaluations is to understand how a model will interact with its ecosystem, which we denote schematically as:
%
\begingroup
\begin{multline}
\qquad
\tag{Application-centric Evaluation Goal}
Understand(Model+Ecosystem)
\label{Application-centric Evaluation Goal}
\hfill 
\end{multline}
\endgroup
\assumption{Assumption 1: Consequentialism}
Consequentialism is the view that whether actions are good or bad depends only on their consequences  \cite{consequentialism}. The ML research literature often appeals to motivations about model utility to humans (e.g., \cite{blasi2021systematic, ethayarajh2020utility,bunescu2010utility,lo2010evaluating,idahl2021towards,neumann2020utility,hepp2018learn,orekondy2018connecting,fu2022deep, zhao2020not}, including papers on fairness in ML such as \cite{corbett2017algorithmic, card2020consequentialism, chohlas2021learning, corbett2018measure}). In adopting consequentialism as its \emph{de facto} ethical framework, ML prioritizes the greatest good for the greatest number \cite{ieeeglobal2019} and centers measurable future impacts.
Moreover, the consequences that are centered are the \emph{direct} consequences, with little attention given to motives, rules, or public acceptance \cite{consequentialism}.
This is realised as a focus on the first-order consequences of introducing the model into the ecosystem. Changes to the ecosystem itself---e.g., addressing what social change is perceived as possible and desirable \cite{green2020data, hovy2016social,eckhouse2019layers}---are assumed to be out of scope, as are concerns for setting of precedents for other ML developers. 
We denote this assumption schematically as:
 \BHcomment{cite https://arxiv.org/pdf/2001.00329.pdf and https://arxiv.org/pdf/2109.08792v1.pdf}
\begin{multline}
\qquad
\tag{Consequentialism Assumption}
Understand(Model + Ecosystem)
\approx
Utility(Model + Ecosystem) \hfill
\label{Consequentialism Assumption}
\hfill 
\end{multline}

\evaluationgappar{\evaluationGap 1: Provenance}\enskip
A focus on future consequences neglects important moral considerations regarding the construction of the model. This excludes both deontological concerns---for example, \emph{Were data consent and sovereignty handled appropriately?}  \cite{kukutai2016indigenous,andreotta2021ai,crawford2021excavating} and \emph{Were data workers treated with dignity?} \cite{gray2019ghost}---as well as questions regarding past costs of development---for example, \emph{What were the energy use externalities of model training?} \cite{strubell2019energy, garcia2019estimation, crawford2018anatomy} and \emph{Was the labour paid fairly?} \cite{silberman2018responsible}. Schwartz \etal coin the phrase ``Red AI'' to describe ML work that disregards the costs of training, noting that such work inhibits discussions of when costs might outweigh benefits \cite{schwartz2020green}.

\evaluationgappar{\evaluationGap 2: Social Responsibilities}\enskip
Another outcome of focusing primarily on direct consequences is marginalizing the assessment of a model against the social contracts that guide the ecosystem in which the model is used, such as moral values, principles, laws, and social expectations. For instance, \textit{Does the model adhere to the moral duty to treat people in ways that upholds their basic human rights?} \cite{shue2020basic}, \textit{Does it abide by legal mechanisms of accountability?} \cite{raso2018artificial,mcgregor2019international}, and \textit{Does it satisfy social expectations of inclusion, such as the ``nothing about us without us'' principle?} \cite{charlton1998nothing}.

\assumption{Assumption 2: Abstractability from Context} The model's ecosystem is reduced to a set of considerations $(X, Y)$, i.e., the inputs to the model and the ``ground truth,'' and in practice $X$ may often fail to model socially important yet sensitive aspects of the environment \cite{barocas2017fairness,andrus2019towards}. The model itself is reduced to a predicted value $\hat{Y}$, ignoring e.g., secondary model outputs such as confidence scores, or predictions on auxiliary model heads. 
\begin{multline}\qquad
Utility(Model + Ecosystem)
\approx
Utility(\hat{Y}, X, Y)
\tag{Assumption of Abstractability of Context}
\label{Black-boxing Assumption}
\hfill  \end{multline}

\evaluationgappar{\evaluationGap 3: System Considerations}\enskip
Equating a model with its prediction overlooks the potential usefulness of model interpretability and explainability. Also, reducing an ecosystem to model inputs and ``ground truth'' overlooks questions of system dynamics \cite{martin2020extending, selbst2019fairness}, such as feedback loops, ``humans-in-the-loop,'' and other effects ``due to actions of various agents changing the world'' \cite{bengio2021deep}. Also overlooked are inference-time externalities of energy use \cite{cai2017neuralpower, garcia2019estimation}, cultural aspects of the ecosystem \cite{sambasivan2021re}, and long term impacts \cite{card2020consequentialism}.

\evaluationgappar{\evaluationGap 4: Interpretive Epistemics}\enskip
By positing a variable $Y=y$ which represents the ``ground truth'' of a situation---even in situations involving social phenomena---a  positivist stance on  knowledge is implicitly adopted. That is, a ``true'' value $Y=y$ is taken to be objectively singular and knowable. This contrasts with anthropology's understanding of knowledge as socially and culturally dependent \cite{forsythe2001knowledge} and requiring interpretation \cite{geertz1973}. In the specific cases of CV and NLP discussed in Section~\ref{sec:practice}, cultural aspects of image and language interpretation are typically marginalized (cf.\ \cite{jappy2013introduction, lakoff2008metaphors, berger2008ways, barthes1977image}, for example), exemplifying what Aroyo and Welty call AI's myth of ``One Truth'' \cite{aroyo2015truth}. Furthermore, the positivist stance downplays the importance of questions of construct validity and reliability \cite{jacobs2021measurement, friedler2021possibility}.

\begin{wrapfigure}{r}{0.4\textwidth}
\centering
\vspace{-1mm}
  \includegraphics[width=0.22\textwidth]{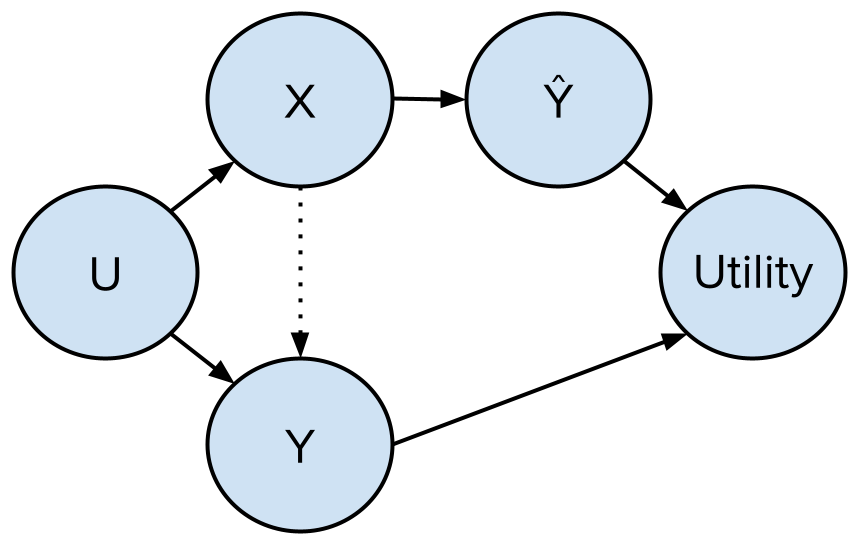}
  \caption{Causal graph illustrating the Input Myopia Assumption.}
  \label{fig:independence}
\end{wrapfigure}
\assumption{Assumption 3: Input Myopia}
Once the input variable $X$ has been used by the model to calculate the model prediction $\hat{Y}$, $X$ is typically ignored for the remainder of the evaluation.
That is, the utility of the model is assumed to depend only on the model's prediction and on the ``ground truth.'' We illustrate this with a causal graph diagram in Figure~\ref{fig:independence}, which shows  Utility as independent of $X$ once the effects of $\hat{Y}$ and $Y$ are taken into account.
\begin{multline}\qquad
Utility(\hat{Y}, X, Y)
\approx
Utility(\hat{Y} , Y)
\tag{Input Myopia Assumption}
\label{Input Myopia Assumption}
\hfill  \end{multline}

\evaluationgappar{\evaluationGap 5: Disaggregated Analyses}\enskip
By reducing the variables of interest to the evaluation to the prediction $Y$ and the ground truth $\hat{Y}$, 
the downstream evaluation is denied the potential to use $X$.
This exacerbates \evaluationGap~3 by further abstracting the evaluation statistics from their contexts. For example, $X$ could have been used to disaggregate the evaluation statistics in various dimensions---including for fairness analyses, assuming that socio-demographic data is available and appropriate \cite{andrus2021we, barocas2021designing}---or to examine regions of the input space which raise critical safety concerns (e.g., distinguishing a computer vision model's failure to recognise a pedestrian on the sidewalk from failure to recognise one crossing the road) \cite{amodei2016concrete}.
Similarly, robustness analyses which compare the model predictions for related inputs in the same neighborhood of the input space are also excluded.

\assumption{Assumption 4: Quantifiability} We have not yet described any modeling assumptions about the mathematical or topological nature of the implied $Utility$ function, which up to now has been conceived as an arbitrary procedure producing an arbitrary output.
We observe, however, that when models are evaluated, there is a social desire to produce a small number of scalar scores. This is reinforced by ``leaderboardism'' \cite{ethayarajh2020utility}, and extends to the point of averaging different types of scores such as correlation and accuracy \cite{wang2018glue}. We identify two assumptions here: first, that impacts on each individual can be reduced to a single numeric value (and thus different dimensions of impacts are commensurable\footnote{E.g., one machine learning fairness paper says ``$c$ is the cost of detention in units of crime prevented'' \cite{corbett2017algorithmic}.}); second, that impacts across individuals are similarly commensurable. We define $\hat{y}\in \hat{Y}$  and $y\in Y$  to be a specific model prediction, and a specific "ground truth" value respectively, leading to the Individual Quantifiability Assumption and the Collective Quantifiability Assumption, respectively.
\begin{multline}\qquad
Utility(\hat{y}, y) \in \mathbb{R}
\tag{Individual Quantifiability Assumption}
\label{Quantitative Assumption}
\hfill  \end{multline}
\begin{multline}\qquad
Utility(\hat{Y}, Y)
\approx
E_{(\hat{y}, y)} \{Utility(\hat{y}, y)\}
\tag{Collective Quantifiability Assumption}
\label{Averaging Assumption}
\hfill  \end{multline}
Composing these assumptions with the previous ones leads to the belief that the evaluation can be summarized as a scalar statistic:  $Utility(\hat{Y}, Y) \in \mathbb{R}$.

\evaluationgappar{\evaluationGap 6: Incommensurables}\enskip
The Quantifiability Assumptions assume that the impacts on individuals are reducible to numbers, trivializing the frequent difficulty in comparing different benefits and costs \cite{marrkula2019}. Furthermore, the harms and benefits across individuals are assumed to be comparable in the same scale. These assumptions are likely to disproportionately impact underrepresented groups, for whom model impacts might differ in qualitative ways from the well represented groups \cite{sambasivan2021re, sambasivan2018toward, heldreth2021don}. The former groups are less likely to be represented in the ML team \cite{west2019discriminating} and hence less likely to have their standpoints on harms and benefits acknowledged.

\assumption{Assumption 5: Failures Cases Are Equivalent}
For classification tasks, common evaluation metrics such as  accuracy or error rate  model the utility of $\hat{Y}$ as binary (i.e., either 1 or 0), depending entirely on whether or not it is equal to the ``ground truth'' $Y$. That is, for a binary task, 
$Utility(\hat{Y}$=$0,Y$=$0)$=$Utility(\hat{Y}$=$1, Y$=$1)$=$1$
and $Utility(\hat{Y}$=$0, Y$=$1)$=$Utility(\hat{Y}$=$1, Y$=$0)$=$0$. 
Similarly for regression tasks, common metrics such as MAE and MSE take the magnitude of error into account, yet still treat certain failures as equivalent (specifically, $Utility(\hat{Y}$=$\hat{y}, Y$=$\hat{y}+\delta)$=$Utility(\hat{Y}$=$\hat{y}, Y$=$\hat{y}-\delta)$, for all $\delta, \hat{y}$).
\begin{multline}\qquad
Utility(\hat{Y}=\hat{y}, Y=y)
\approx 
\mathbbm{1}(\hat{y}=y)
\tag{Assumption of Equivalent Failures [Classification]}
\label{Equivalent Failures Assumption 1}
\hfill  \end{multline}
\begin{multline}\qquad
Utility(\hat{y}, y)
\text{ is a function of } 
|\hat{y}-y|
\tag{Assumption of Equivalent Failures [Regression]}
\label{Equivalent Failures Assumption 2}
\hfill  \end{multline}
Taken together with the previous assumptions, this yields $Utility(\hat{y}, y)=P(\hat{y}=y)$ for classification tasks.

\evaluationgappar{\evaluationGap 7: Disparate harms and benefits}\enskip
Treating all failure cases as equivalent fails to appreciate that different classes of errors often have very different impacts \cite{provost1997analysis, challen2019artificial}. In multiclass classification, severely offensive predictions (e.g., predicting an animal in an image of a person) are given the same weight as inoffensive ones.
In regression tasks, insensitivity to either the direction of the difference $\hat{y}-y$ or the magnitude of $y$ can result in evaluations being possibly poor proxies for downstream impacts.
(One common application use case of regression models is to apply a cutoff threshold $t$ to the predicted scalar values, for which both the direction of error and the magnitude of $y$ are relevant.)

\assumption{Assumption 6: Test Data Validity}
Taken collectively, the previous assumptions might lead one to use accuracy as an evaluation metric for a classification task. Further assumptions can then be made in deciding \emph{how} to estimate accuracy.
The final assumption we discuss here is that the test data over which accuracy (or other metrics) is calculated provides a good estimate of the accuracy of the model when embedded in the ecosystem. 
\begin{multline}\qquad
P(\hat{y}=y) \approx P(\hat{y}'=y')
\tag{Assumption of Test Data Validity [Classification]}
\label{Assumption of Test Data Validity}
\hfill  \end{multline}
where $Y'=y'$ and $\hat{Y}'=\hat{y}'$ are the ground truth labels and the model predictions on the test data, respectively.

\evaluationgappar{\evaluationGap 8: Data Drifts}\enskip
A simple model of the ecosystem's data distributions is particularly risky when system feedback effects would cause the distributions of data in the ecosystem to diverge from those in the evaluation sample \cite{liu2018delayed,kannan2019downstream}. In general, this can lead to overconfidence in the system's reliability, which can be exacerbated for regions in the tail of the input distribution.

\subsection{Discussion}
\begin{table}
\renewcommand*{\arraystretch}{1.2}
\begin{tabular}{llp{0.39\textwidth}} \toprule
    &Assumption&Considerations that might be Overlooked\\\midrule
    $Understand(Model + Ecosystem)$&Application-centric evaluation&Opportunities for scientific insights. \\
    $\approx Utility(Model + Ecosystem)$                                 &\text{Consequentialism}
        &Data sourcing and processing; invisible labour; consultation with impacted communities; motives; public acceptance; human rights.\\
    $\approx Utility(\hat{Y}, X, Y)$                                    &\text{Abstractability from Context}
        &System feedback loops; humans-in-the-loop.\\   
    $\approx Utility(\hat{Y}, Y)$                                      &\text{Input Myopia}
    &Disaggregated analyses; sensitivity analyses; safety-critical edge cases.\\
    $\approx E_{(\hat{Y}=\hat{y}, Y=y)}\{Utility(\hat{y}, y)\}$ &\text{Quantitative Modeling}
    &Different flavors of impacts on a single person; different flavors of impacts across groups.\\
    $\approx P(\hat{y} = y)$                    &\text{Equivalent Failures}
        &Severe failure cases; confusion matrices; topology of the prediction space.\\
$\approx P(\hat{y}'=y')$      &\text{Test Data Validity}
    &Data sampling biases; distribution shifts.\\
\bottomrule
    \end{tabular}
\caption{Sketch of how the six assumptions of Section~\ref{sec:gaps}---when taken collectively---compose to simplify the task of evaluating a model for an application ($Understand(Model + Ecosystem)$) to one of calculating accuracy over a data sample.
A pseudo-formal notation (akin to pseudo-code) is used to enable rapid glossing of the main connections.
$Y=y$ and $\hat{Y}=\hat{y}$ denote the true (unobserved) distributions of ground truth and model predictions, respectively, while the variables $Y'=y'$ and $\hat{Y}'=\hat{y}'$ denote the samples of reference labels and model predictions over which accuracy is calculated in practice.  The order of the assumptions reflects an increasing focus on technical aspects of model evaluation, and a corresponding minimizing of non-technical aspects. Appendix~C illustrates how each of the sets of considerations might apply in a hypothetical application of a computer vision model.}
\label{fig:sketch}
\end{table}

We have described six assumptions that simplify the model evaluation task. Taken together, they would cause one to believe---with compounding risks---that a model's accuracy is a good proxy for its fitness for an application. We sketch this composition of assumptions in Figure~\ref{fig:sketch}, along with questions that illustrate the gaps raised by each assumption. Our reason for teasing apart these assumptions and their compounding effects is \emph{not} to attack the ``strawman'' of naive application-centric evaluations which rely solely on estimating model accuracy. Rather, our goal is to point out that most model evaluations, even sophisticated ones, make such assumptions to varying degrees. For example:
\begin{itemize}
    \item Some robustness evaluations (for surveys, see \cite{farahani2020brief, wang2021generalizing}) explicitly tackle the problem of distribution shifts, rejecting the Assumptions of Test Data Validity without questioning the other assumptions we have identified.
    \item Some sensitivity evaluations consider the effect on the model predictions of small changes in the input, but use accuracy as an evaluation metric, rejecting the Input Myopia Assumption without questioning the others \cite{ribeiro2020beyond}.
    \item Some fairness evaluations perform disaggregated analyses using the Recall or Precision metrics, sticking by all assumptions other than Input Myopia and Equivalent Failures \cite{chouldechova2017fair, hardt2016equality}.
\end{itemize}
It may not be possible to avoid all of the assumptions all of the time; nevertheless unavoidable assumptions should be acknowledged and critically examined. 
The six assumptions we have identified also provide a lens for assessing the consistency of some evaluation metrics with other assumptions that have been made during the evaluation, for example
\begin{itemize}
\item \emph{Is $F$-score consistent with an utilitarian evaluation framework?}  The $F$-score is mathematically a harmonic mean--- which is often appropriate for averaging pairs of rates (e.g., two speeds). When applied to Precision and Recall, however, the $F$-score constitutes a peculiar averaging of ``apples and oranges,'' since, when conceived as rates, Precision and Recall measure rates of change of different quantities,
\cite{powers2014f}. $F$-score is thus difficult to interpret within an evaluation framework that aims to maximize model utility.
\item \emph{Do threshold-free evaluations such as the Area Under the Receiver Operating Characteristic (\textsc{auroc}) abstract too much of the deployment context?} Since \textsc{auroc} is calculated by averaging over a range of possible threshold values, it ``cannot be interpreted as having any relevance to any particular classifier'' \cite{powers2012problem} (which is not saying \textsc{auroc} is irrelevant to evaluating \emph{the learner}, cf.\ Section~\ref{sec:ideals}, nor to a learned model's propensity to correctly rank positive instances above negative ones). The same argument can be made for the Mean Average Precision metric used in image classification (see Appendix~A). For useful application-centric evaluations, it is more meaningful to report pairs of $(Precision, Recall)$ values (for all classes) for a range of threshold values \cite{powers2012problemauc}.
\end{itemize}
In both cases, we  ask whether such metrics are of limited utility in application-centric evaluations and whether they are better left to learner-centric ones.

\section{Contextualizing Application-centric Model Evaluations}

\begin{quote}
    \emph{the ornithologists were forced to adapt their behavior (for the sake of ``science'') to the most primitive evaluation method which was the only one considered or known, or else throw their data away.} --- Hampel \cite{hampel1998statistics}
\end{quote}

When applications of ML models have the potential to impact human lives and livelihoods, thorough and reliable evaluations of models are critical. As discussed in Section~3, the different goals and values of academic ML research communities mean that research norms cannot be relied upon as guideposts for evaluating models for  applications. In this section, we propose steps towards evaluations that are  rigorous in their methods and  aim to be humble about their epistemic uncertainties. In doing so, we expand on the call by Raji \etal to pay more attention not just to evaluation metric values but also to the quality and reliability of the measurements themselves, including sensitivity to external factors \cite{raji2021ai}.

\subsection{Minding the Gaps between Evaluation Goals and Research Practice}

\label{sec:recommendations}

Documenting assumptions made during model evaluation is critical for transparency and enables more informed decisions.  If an assumption is difficult to avoid in practice, consider augmenting the evaluation with signals that may shed complementary light on questions of concern. For example, even a handful of insightful comments from members of impacted communities can be an invaluable complement to evaluations using quantitative  metrics. We now consider specific mitigation strategies for each of the gaps in turn.

\assumption{Minding Gap 1: Evaluate More than Consequences} To reduce the gap introduced by the Consequentialism Assumption, evaluate the processes that led to the creation of the model, including how datasets were constructed \cite{scheuerman2021datasets}. We echo calls for more reflexivity around social and intentional factors around model development \cite{miceli2021documenting}, more documentation of the complete lifecycle of model development \cite{vogelsang2019requirements,hutchinson2021towards}, and greater transparency around ML models and their datasets \cite{mitchell2019model,gebru2021datasheets,bender2018data}.
It may be appropriate to contemplate whether the model is aligned with the virtues the organization aspires to  \cite{vallor2016technology}. Consider the question of  whether any ML model could be a morally appropriate solution in this application context, e.g., whether it is appropriate to make decisions about one person on the basis of others' behaviors \cite{eckhouse2019layers}. 

\assumption{Minding Gap 2: Center Obligations} Since reasoning about uncertain future states of the world is fraught with challenges \cite{card2020consequentialism}, evaluations should consider indirect consequences and assess how the model upholds social obligations within the ecosystem. This may involve processes such as assessments of human rights, social and ethical impact \cite{mcgregor2019international,mantelero2018ai}, audits of whether the ML system upholds the organization's declared values or principles \cite{raji2020closing}, and/or assessments of the potential for privacy leakage (e.g., \cite{yeom2018privacy, carlini2021extracting}).

\assumption{Minding Gap 3: Demarginalize the Context} To address the gap introduced by the Assumption of Abstractability from Context, consider externalities such as energy consumption \cite{henderson2020towards, schwartz2020green}, as well as resource requirements \cite{ethayarajh2020utility}. It is important to think about how the human and technical parts of the system will interact \cite{selbst2019fairness, martin2020extending}.
Note that when substituting one model for another---or for displaced human labor---system stability can itself be a desirable property independent of model accuracy (and perhaps counter to tech industry discourses of ``disruption'' \cite{geiger2020silicon}), and a range of metrics exist for comparing predictions with those of a legacy model \cite{derczynski2016complementarity}. 
Care should be taken to avoid the ``portability trap'' of assuming that what is good for one context is good for another \cite{selbst2019fairness}. The more attention paid to the specifics of the application context, the better; hence, metrics which assume no particular classification threshold, such as AUC, may provide limited signal for any single context.  

\assumption{Minding Gap 4: Make Subjectivities Transparent} Acknowledge the subjectivities inherent in many tasks \cite{alm2011subjective}. An array of recent scholarship on subjectivity in ML has ``embraced disagreement'' through practices of explicitly modeling---in both the data model and the ML model---inter-subject variation in interpretations
\cite{davani2022dealing, basile2021toward, fornaciari2021beyond, diaz2019whose, aroyo2015truth}. For the purposes of ML model evaluations, disaggregating labels on test data according to the cultural and socio-demographic standpoints of their annotators enables more nuanced disaggregated evaluation statistics \cite{prabhakaran2021releasing}.

\assumption{Minding Gap 5: Respect Differences Between Inputs} A realistic ``null hypothesis'' is that misclassifications affect people in the application ecosystem disparately. For example, people may differ both in their preferences regarding model predictions $\hat{Y}$ \emph{per se}, as well as their preferences regarding model accuracy $\hat{Y}=Y$ \cite{binns2018fairness}.\footnote{Note that in many real-world applications the ``ground truth'' variable $Y$ may be a convenient counterfactual fiction, since the system's actions on the basis of the prediction $\hat{Y}$ may inhibit $Y$ from being realised---for example, a finance ML model may predict a potential customer would default on a loan if given one, and hence the system the model is deployed in may prevent the customer getting a loan in the first place.} As such---and \emph{independent of fairness considerations}---evaluations should be routinely pay attention to different parts of the input distribution, including disaggregating along social subgroups. Special attention should be paid to the tail of the distribution and outliers during evaluation, as these may require further analysis to diagnose the potential for rare but unsafe impacts. Input sensitivity testing can provide useful information about the sensitivity of the classifier to dimensions of input variation known to be of concern (e.g., gender in text \cite{zhao2018gender, borkan2019nuanced, gonen2020automatically, huang2020reducing}).

\assumption{Minding Gap 6: Think Beyond Scalar Utility} Resist the temptation to reduce a model's utility to a single scalar value, either for stack ranking \cite{ethayarajh2020utility} or to simplify the cognitive load on decision makers. Instead, include a range of different metrics and evaluation distributions in the evaluation \cite{mitchell2019model}. Acknowledge and report epistemic uncertainty, e.g., the effects of missing data or measurement and sampling error on metrics. Acknowledge qualitative impacts that are not addressed by metrics (e.g., harms to application users caused by supplanting socially meaningful human interactions), and rigorously assess the validity of attempts to measure social or emotional harms. Be conservative in aggregations: consider plotting data rather than reporting summary statistics (cf.\ Anscombe's quartet); do not aggregate unlike quantities; report multiple estimates of central tendency and variation; and don't assume that all users of an application will have equal benefits (or harms) from system outcomes. Consider applying aggregation and partial ranking techniques from the fair division literature to ML models, including techniques that give greater weight to those with the worst outcomes (e.g., in the extreme case, ``Maximin'') \cite{endriss2018lecture}.

\assumption{Minding Gap 7: Respect Differences Between Failures} If the harms of false positives and false negatives are incommensurable, report them separately. If commensurable, weight each appropriately. For multiclass classifiers, this approach generalizes to a \emph{classification cost matrix} \cite{turney1994cost}, and, more generally, including the \emph{confusion matrix} before costs are assigned; for regression tasks, report metrics such as MSE disaggregated by buckets of $Y$. 

\assumption{Minding Gap 8: Validate Quality of Test Data}
For transparency, do not assume it is obvious to others which datasets are used in training and evaluation; instead, be explicit about the provenance, distribution, and known biases of the datasets in use \cite{andrus2021we}. Consider Bayesian approaches to dealing with uncertainty about data distributions \cite{mcnair2018preventing,ji2020can,lacoste2017deep}, especially when sample sizes are small or prior work has revealed systematic biases. For example, an evaluation which uses limited data in a novel domain (or in an under-studied language) to investigate gender biases in pronoun resolution  should be tentative in drawing strong positive conclusions about ``fairness'' due to abundant evidence of gender biases in English pronoun resolution models (e.g. \cite{webster2019gendered}).

\subsection{Alternate Model Evaluation Methodologies}

More radical excursions from the disciplinary paradigm are often worth considering, especially in scenarios with high stakes or high uncertainty.

\assumption{Evaluation Remits} In 1995, Sparck Jones and Galliers called for a careful approach to NLP evaluation that is broadly applicable to ML model evaluations (see Appendix\ D) \cite{jones1995evaluating}. Their approach involves a top-down examination of the context and goal of the evaluation before the evaluation design even begins, and their call for careful documentation of the evaluation ``remit''---i.e., official  responsibilities---is in line with more recent work calling for stakeholder transparency for ML \cite{raji2020closing, hutchinson2021towards}. They advocate for establishing whose perspectives are adopted in the evaluation and whose interests prompted it. Appendix~D sketches how Sparck Jones and Galliers' framework could be adopted for ML model evaluations. 

\assumption{Active Testing} Active Testing aims to iteratively choose new items that are most informative in addressing the goals of the evaluation \cite{kossen2021active, ha2021alt} (cf.\  its cousin Active Learning, which selects items that are informative for the learner). Active Testing provides a better estimate of model performance than using the same number of test instances sampled \iid Exploring Active Testing in pursuit of fairness testing goals seems a promising direction for future research.

\assumption{Adversarial Testing} In many cases, there is great uncertainty regarding an application deployment context. One cautious and conservative approach---especially in the face of great uncertainty---is to simulate ``adversaries'' trying to provoke harmful outcomes from the system. Borrowing adversarial techniques from security testing and privacy testing, adversarial testing of models requires due diligence to trigger the most harmful model predictions, using either manually chosen or algorithmically generated test instances \cite{ruiz2022simulated, zeng2021openattack, zhang2020adversarial, ettinger2017towards}. 

\assumption{Multidimensional Comparisons} When comparing candidate models, avoid the ``Leaderboardism Trap'' of believing that a total ordering of candidates is possible. A multidimensional and nuanced evaluation may provide at best a partial ordering of candidate models, and it may require careful and accountable judgement and qualitative considerations to decide among them. The Fair Division literature on Social Welfare Orderings may be a promising direction for developing evaluation frameworks that prioritize ``egalitarian'' considerations, in which greater weighting is given to those who are worst impacted by a model \cite{endriss2018lecture}.

\BHcomment{Discuss the conceptual framework for machine learning evaluations in \url{https://datasets-benchmarks-proceedings.neurips.cc/paper/2021/file/757b505cfd34c64c85ca5b5690ee5293-Paper-round2.pdf}}

\subsection{Evaluation-driven ML Methodologies}

In this section, we follow Rostamzadeh \etal in drawing inspiration from test-driven practices, such as those of software development \cite{rostamzadeh2021thinking}.
Traditional software testing involves significant time, resources, and effort \cite{harrold2000testing}; even moderate-sized software projects spend hundreds of person-hours writing test cases, implementing them, and meticulously documenting the test results. In fact, software testing is sometimes considered an art \cite{myers2011art} requiring its own technical and non-technical skills \cite{sanchez2020beyond,matturro2013soft}, and entire career paths are built around testing \cite{cunningham2019software}. 
\emph{Test-driven development}, often associated with agile software engineering frameworks, integrates testing considerations in all parts of the development process \cite{astels2003test,george2004structured}. These processes rely on a deep understanding of software requirements and user behavior to anticipate failure modes during deployment and to expand the test suite. (In contrast, ML testing is often relegated to a small portion of the ML development cycle,
 and predominantly focuses on a static snapshot of data to provide performance guarantees.)
These software testing methodologies provide a model for ML testing. First, the model suggests anticipating, planning for, and integrating testing in all stages of the development cycle, research problem ideation, the setting of objectives, and system implementation. Second, build a practice around bringing diverse perspectives into designing the test suite. Additionally, consider participatory approaches (e.g., \cite{martin2020extending}) to ensure that the test suite accounts for societal contexts and embedded values within which the ML system will be deployed. 

An important principle in test-driven software development is visibility into the test data. Typically, engineers working on a system can not only see the details of the test suites but also often develop those test suites themselves.  In contrast, the paradigm of ML evaluation methodologies is that the ML practitioner should not inspect the test data, lest their observations result in design decisions that produce an overfitted model. How, then, can these two methodologies be reconciled? We believe that incentives are one important consideration. In the ML research community, the ``competition mindset'' might indeed lead to ``cheating'' via deliberate overfitting. In contrast, in real-world applications model developers might benefit from a healthy model ecosystem, for example when they are members of that ecosystem. (However, when developers come from a different society altogether there may be disinterest or disalignment \cite{sambasivan2021re}.)

Software testing produces artifacts such as execution traces,
and test coverage information \cite{harrold2000testing}. Developing practices for routinely sharing  testing artifacts with  stakeholders provides for more robust scrutiny and diagnosis of harmful error cases \cite{raji2020closing}. In being flexible enough to adapt to the information needs of stakeholders, software testing artifacts can be considered a form of boundary object \cite{star1989institutional}. Within an ML context, these considerations point towards adopting ML transparency mechanisms incorporating comprehensive evaluations, such as model cards \cite{mitchell2019model}. The processes that go into building test cases should be documented, so the consumer of the ML system can better understand the system's reliability. Finally, as for any high-stakes system---software, ML or otherwise---evaluation documentation constitutes an important part of the chain of auditable artifacts required for robust accountability and governance practices \cite{raji2020closing}.

\section{Conclusions}
\label{sec:conclusions}

In this paper, we compared the evaluation practices in the ML research community to the ideal information needs of those who use models in real-world applications.
The observed disconnect between the two is likely due to differences in motivations and goals, and also pressures to demonstrate ``state of the art'' performance on shared tasks, metrics and leaderboards \cite{ethayarajh2020utility, koch2021reduced, thomas2020reliance}, as well as a focus on the learner as the object upon which the researcher hopes to shed light.
One limitation of our methodology is reliance on published papers, and we encourage more human subjects research in the future, in a similar vein to e.g.\ \cite{holstein2019improving, sambasivan2021everyone, madaio2022assessing}.
We identified a range of \evaluationgaps that risk being overlooked if the ML research community's evaluation practices are uncritically adopted when for applications, and identify six assumptions that would have to be valid if these gaps are to be overlooked.
The assumptions range from a broad focus on consequentialism to technical concerns regarding distributions of evaluation data.
By presenting these assumptions as a coherent framework, we provide not just a set of mitigations for each evaluation gap, but also demonstrate the relationships between these mitigations.
We show how in the naive case these assumptions chain together, leading to the grossest assumption that calculating model accuracy on data \iid with the training data can be a reliable signal for real-world applications.
We contrast the practices of ML model evaluation with those of the mature engineering practices of software testing to draw out lessons for non-\iid testing under a variety of stress conditions and failure severities. 
One limitation of our analysis is that we are generally domain-agnostic, and we hope to stimulate investigations of assumptions and gaps for specific application domains.
We believe it is fundamental that model developers are explicit about methodological assumptions in their evaluations.
We believe that ML model evaluations have great potential to enable interpretation and use by different technical and non-technical communities \cite{star1989institutional}.
By naming each assumption we identify and exploring its technical and sociological consequences, we hope to encourage more robust interdisciplinary debate and, ultimately, to nudge model evaluation practice away from abundant opaque unknowns.

\begin{acks}
We acknowledge useful feedback from Daniel J.\ Barrett, Alexander D'Amour, Stephen Pfohl, D.\ Sculley, and the anonymous reviewers.
\end{acks}

\bibliographystyle{ACM-Reference-Format}
\bibliography{cameraready}


\begin{thebibliography}{180}


\ifx \showCODEN    \undefined \def \showCODEN     #1{\unskip}     \fi
\ifx \showDOI      \undefined \def \showDOI       #1{#1}\fi
\ifx \showISBNx    \undefined \def \showISBNx     #1{\unskip}     \fi
\ifx \showISBNxiii \undefined \def \showISBNxiii  #1{\unskip}     \fi
\ifx \showISSN     \undefined \def \showISSN      #1{\unskip}     \fi
\ifx \showLCCN     \undefined \def \showLCCN      #1{\unskip}     \fi
\ifx \shownote     \undefined \def \shownote      #1{#1}          \fi
\ifx \showarticletitle \undefined \def \showarticletitle #1{#1}   \fi
\ifx \showURL      \undefined \def \showURL       {\relax}        \fi
\providecommand\bibfield[2]{#2}
\providecommand\bibinfo[2]{#2}
\providecommand\natexlab[1]{#1}
\providecommand\showeprint[2][]{arXiv:#2}

\bibitem[\protect\citeauthoryear{Alam, Joty, and Imran}{Alam
  et~al\mbox{.}}{2018}]%
        {alam2018domain}
\bibfield{author}{\bibinfo{person}{Firoj Alam}, \bibinfo{person}{Shafiq Joty},
  {and} \bibinfo{person}{Muhammad Imran}.} \bibinfo{year}{2018}\natexlab{}.
\newblock \showarticletitle{Domain Adaptation with Adversarial Training and
  Graph Embeddings}. In \bibinfo{booktitle}{\emph{Proceedings of the 56th
  Annual Meeting of the Association for Computational Linguistics (Volume 1:
  Long Papers)}}. \bibinfo{pages}{1077--1087}.
\newblock


\bibitem[\protect\citeauthoryear{Alm}{Alm}{2011}]%
        {alm2011subjective}
\bibfield{author}{\bibinfo{person}{Cecilia~Ovesdotter Alm}.}
  \bibinfo{year}{2011}\natexlab{}.
\newblock \showarticletitle{Subjective natural language problems: Motivations,
  applications, characterizations, and implications}. In
  \bibinfo{booktitle}{\emph{Proceedings of the 49th Annual Meeting of the
  Association for Computational Linguistics: Human Language Technologies}}.
  \bibinfo{pages}{107--112}.
\newblock


\bibitem[\protect\citeauthoryear{Amodei, Olah, Steinhardt, Christiano,
  Schulman, and Man{\'e}}{Amodei et~al\mbox{.}}{2016}]%
        {amodei2016concrete}
\bibfield{author}{\bibinfo{person}{Dario Amodei}, \bibinfo{person}{Chris Olah},
  \bibinfo{person}{Jacob Steinhardt}, \bibinfo{person}{Paul Christiano},
  \bibinfo{person}{John Schulman}, {and} \bibinfo{person}{Dan Man{\'e}}.}
  \bibinfo{year}{2016}\natexlab{}.
\newblock \showarticletitle{Concrete problems in AI safety}.
\newblock \bibinfo{journal}{\emph{arXiv preprint arXiv:1606.06565}}
  (\bibinfo{year}{2016}).
\newblock


\bibitem[\protect\citeauthoryear{Andreotta, Kirkham, and Rizzi}{Andreotta
  et~al\mbox{.}}{2021}]%
        {andreotta2021ai}
\bibfield{author}{\bibinfo{person}{Adam~J Andreotta}, \bibinfo{person}{Nin
  Kirkham}, {and} \bibinfo{person}{Marco Rizzi}.}
  \bibinfo{year}{2021}\natexlab{}.
\newblock \showarticletitle{{AI}, big data, and the future of consent}.
\newblock \bibinfo{journal}{\emph{{AI} \& Society}} (\bibinfo{year}{2021}),
  \bibinfo{pages}{1--14}.
\newblock


\bibitem[\protect\citeauthoryear{Andrus and Gilbert}{Andrus and
  Gilbert}{2019}]%
        {andrus2019towards}
\bibfield{author}{\bibinfo{person}{McKane Andrus} {and}
  \bibinfo{person}{Thomas~K Gilbert}.} \bibinfo{year}{2019}\natexlab{}.
\newblock \showarticletitle{Towards a just theory of measurement: A principled
  social measurement assurance program for machine learning}. In
  \bibinfo{booktitle}{\emph{Proceedings of the 2019 AAAI/ACM Conference on
  {AI}, Ethics, and Society}}. \bibinfo{pages}{445--451}.
\newblock


\bibitem[\protect\citeauthoryear{Andrus, Spitzer, Brown, and Xiang}{Andrus
  et~al\mbox{.}}{2021}]%
        {andrus2021we}
\bibfield{author}{\bibinfo{person}{McKane Andrus}, \bibinfo{person}{Elena
  Spitzer}, \bibinfo{person}{Jeffrey Brown}, {and} \bibinfo{person}{Alice
  Xiang}.} \bibinfo{year}{2021}\natexlab{}.
\newblock \showarticletitle{What We Can't Measure, We Can't Understand:
  Challenges to Demographic Data Procurement in the Pursuit of Fairness}. In
  \bibinfo{booktitle}{\emph{Proceedings of the 2021 ACM Conference on Fairness,
  Accountability, and Transparency}}. \bibinfo{pages}{249--260}.
\newblock


\bibitem[\protect\citeauthoryear{Aroyo and Welty}{Aroyo and Welty}{2015}]%
        {aroyo2015truth}
\bibfield{author}{\bibinfo{person}{Lora Aroyo} {and} \bibinfo{person}{Chris
  Welty}.} \bibinfo{year}{2015}\natexlab{}.
\newblock \showarticletitle{Truth is a lie: Crowd truth and the seven myths of
  human annotation}.
\newblock \bibinfo{journal}{\emph{AI Magazine}} \bibinfo{volume}{36},
  \bibinfo{number}{1} (\bibinfo{year}{2015}), \bibinfo{pages}{15--24}.
\newblock


\bibitem[\protect\citeauthoryear{Astels}{Astels}{2003}]%
        {astels2003test}
\bibfield{author}{\bibinfo{person}{Dave Astels}.}
  \bibinfo{year}{2003}\natexlab{}.
\newblock \bibinfo{booktitle}{\emph{Test driven development: A practical
  guide}}.
\newblock \bibinfo{publisher}{Prentice Hall Professional Technical Reference}.
\newblock


\bibitem[\protect\citeauthoryear{Barocas, Guo, Kamar, Krones, Morris, Vaughan,
  Wadsworth, and Wallach}{Barocas et~al\mbox{.}}{2021}]%
        {barocas2021designing}
\bibfield{author}{\bibinfo{person}{Solon Barocas}, \bibinfo{person}{Anhong
  Guo}, \bibinfo{person}{Ece Kamar}, \bibinfo{person}{Jacquelyn Krones},
  \bibinfo{person}{Meredith~Ringel Morris}, \bibinfo{person}{Jennifer~Wortman
  Vaughan}, \bibinfo{person}{Duncan Wadsworth}, {and} \bibinfo{person}{Hanna
  Wallach}.} \bibinfo{year}{2021}\natexlab{}.
\newblock \showarticletitle{Designing Disaggregated Evaluations of AI Systems:
  Choices, Considerations, and Tradeoffs}.
\newblock \bibinfo{journal}{\emph{arXiv preprint arXiv:2103.06076}}
  (\bibinfo{year}{2021}).
\newblock


\bibitem[\protect\citeauthoryear{Barocas, Hardt, and Narayanan}{Barocas
  et~al\mbox{.}}{2017}]%
        {barocas2017fairness}
\bibfield{author}{\bibinfo{person}{Solon Barocas}, \bibinfo{person}{Moritz
  Hardt}, {and} \bibinfo{person}{Arvind Narayanan}.}
  \bibinfo{year}{2017}\natexlab{}.
\newblock \showarticletitle{Fairness in machine learning}.
\newblock \bibinfo{journal}{\emph{{NIPS} tutorial}}  \bibinfo{volume}{1}
  (\bibinfo{year}{2017}), \bibinfo{pages}{2017}.
\newblock


\bibitem[\protect\citeauthoryear{Barthes}{Barthes}{1977}]%
        {barthes1977image}
\bibfield{author}{\bibinfo{person}{Roland Barthes}.}
  \bibinfo{year}{1977}\natexlab{}.
\newblock \bibinfo{booktitle}{\emph{{Image-Music-Text}}}.
\newblock \bibinfo{publisher}{Macmillan}.
\newblock


\bibitem[\protect\citeauthoryear{Basile, Cabitza, Campagner, and Fell}{Basile
  et~al\mbox{.}}{2021}]%
        {basile2021toward}
\bibfield{author}{\bibinfo{person}{Valerio Basile}, \bibinfo{person}{Federico
  Cabitza}, \bibinfo{person}{Andrea Campagner}, {and} \bibinfo{person}{Michael
  Fell}.} \bibinfo{year}{2021}\natexlab{}.
\newblock \showarticletitle{Toward a Perspectivist Turn in Ground Truthing for
  Predictive Computing}.
\newblock \bibinfo{journal}{\emph{arXiv preprint arXiv:2109.04270}}
  (\bibinfo{year}{2021}).
\newblock


\bibitem[\protect\citeauthoryear{Bender and Friedman}{Bender and
  Friedman}{2018}]%
        {bender2018data}
\bibfield{author}{\bibinfo{person}{Emily~M Bender} {and} \bibinfo{person}{Batya
  Friedman}.} \bibinfo{year}{2018}\natexlab{}.
\newblock \showarticletitle{Data statements for natural language processing:
  Toward mitigating system bias and enabling better science}.
\newblock \bibinfo{journal}{\emph{Transactions of the Association for
  Computational Linguistics}}  \bibinfo{volume}{6} (\bibinfo{year}{2018}),
  \bibinfo{pages}{587--604}.
\newblock


\bibitem[\protect\citeauthoryear{Bender, Gebru, McMillan-Major, and
  Shmitchell}{Bender et~al\mbox{.}}{2021}]%
        {bender2021dangers}
\bibfield{author}{\bibinfo{person}{Emily~M Bender}, \bibinfo{person}{Timnit
  Gebru}, \bibinfo{person}{Angelina McMillan-Major}, {and}
  \bibinfo{person}{Shmargaret Shmitchell}.} \bibinfo{year}{2021}\natexlab{}.
\newblock \showarticletitle{On the Dangers of Stochastic Parrots: Can Language
  Models Be Too Big?}. In \bibinfo{booktitle}{\emph{Proceedings of the 2021 ACM
  Conference on Fairness, Accountability, and Transparency}}.
  \bibinfo{pages}{610--623}.
\newblock


\bibitem[\protect\citeauthoryear{Bengio, Lecun, and Hinton}{Bengio
  et~al\mbox{.}}{2021}]%
        {bengio2021deep}
\bibfield{author}{\bibinfo{person}{Yoshua Bengio}, \bibinfo{person}{Yann
  Lecun}, {and} \bibinfo{person}{Geoffrey Hinton}.}
  \bibinfo{year}{2021}\natexlab{}.
\newblock \showarticletitle{Deep learning for AI}.
\newblock \bibinfo{journal}{\emph{Commun. ACM}} \bibinfo{volume}{64},
  \bibinfo{number}{7} (\bibinfo{year}{2021}), \bibinfo{pages}{58--65}.
\newblock


\bibitem[\protect\citeauthoryear{Berger}{Berger}{2008}]%
        {berger2008ways}
\bibfield{author}{\bibinfo{person}{John Berger}.}
  \bibinfo{year}{2008}\natexlab{}.
\newblock \bibinfo{booktitle}{\emph{Ways of seeing}}.
\newblock \bibinfo{publisher}{Penguin {UK}}.
\newblock


\bibitem[\protect\citeauthoryear{Binns}{Binns}{2018}]%
        {binns2018fairness}
\bibfield{author}{\bibinfo{person}{Reuben Binns}.}
  \bibinfo{year}{2018}\natexlab{}.
\newblock \showarticletitle{Fairness in machine learning: Lessons from
  political philosophy}. In \bibinfo{booktitle}{\emph{Conference on Fairness,
  Accountability and Transparency}}. PMLR, \bibinfo{pages}{149--159}.
\newblock


\bibitem[\protect\citeauthoryear{Birhane, Kalluri, Card, Agnew, Dotan, and
  Bao}{Birhane et~al\mbox{.}}{2021}]%
        {birhane2021values}
\bibfield{author}{\bibinfo{person}{Abeba Birhane}, \bibinfo{person}{Pratyusha
  Kalluri}, \bibinfo{person}{Dallas Card}, \bibinfo{person}{William Agnew},
  \bibinfo{person}{Ravit Dotan}, {and} \bibinfo{person}{Michelle Bao}.}
  \bibinfo{year}{2021}\natexlab{}.
\newblock \showarticletitle{The values encoded in machine learning research}.
\newblock \bibinfo{journal}{\emph{arXiv preprint arXiv:2106.15590}}
  (\bibinfo{year}{2021}).
\newblock


\bibitem[\protect\citeauthoryear{Blasi, Anastasopoulos, and Neubig}{Blasi
  et~al\mbox{.}}{2021}]%
        {blasi2021systematic}
\bibfield{author}{\bibinfo{person}{Dami{\'a}n Blasi}, \bibinfo{person}{Antonios
  Anastasopoulos}, {and} \bibinfo{person}{Graham Neubig}.}
  \bibinfo{year}{2021}\natexlab{}.
\newblock \showarticletitle{Systematic Inequalities in Language Technology
  Performance across the World's Languages}.
\newblock \bibinfo{journal}{\emph{arXiv preprint arXiv:2110.06733}}
  (\bibinfo{year}{2021}).
\newblock


\bibitem[\protect\citeauthoryear{Bommasani, Hudson, Adeli, Altman, Arora, von
  Arx, Bernstein, Bohg, Bosselut, Brunskill, et~al\mbox{.}}{Bommasani
  et~al\mbox{.}}{2021}]%
        {bommasani2021opportunities}
\bibfield{author}{\bibinfo{person}{Rishi Bommasani}, \bibinfo{person}{Drew~A
  Hudson}, \bibinfo{person}{Ehsan Adeli}, \bibinfo{person}{Russ Altman},
  \bibinfo{person}{Simran Arora}, \bibinfo{person}{Sydney von Arx},
  \bibinfo{person}{Michael~S Bernstein}, \bibinfo{person}{Jeannette Bohg},
  \bibinfo{person}{Antoine Bosselut}, \bibinfo{person}{Emma Brunskill},
  {et~al\mbox{.}}} \bibinfo{year}{2021}\natexlab{}.
\newblock \showarticletitle{On the opportunities and risks of foundation
  models}.
\newblock \bibinfo{journal}{\emph{arXiv preprint arXiv:2108.07258}}
  (\bibinfo{year}{2021}).
\newblock


\bibitem[\protect\citeauthoryear{Borkan, Dixon, Sorensen, Thain, and
  Vasserman}{Borkan et~al\mbox{.}}{2019}]%
        {borkan2019nuanced}
\bibfield{author}{\bibinfo{person}{Daniel Borkan}, \bibinfo{person}{Lucas
  Dixon}, \bibinfo{person}{Jeffrey Sorensen}, \bibinfo{person}{Nithum Thain},
  {and} \bibinfo{person}{Lucy Vasserman}.} \bibinfo{year}{2019}\natexlab{}.
\newblock \showarticletitle{Nuanced metrics for measuring unintended bias with
  real data for text classification}. In \bibinfo{booktitle}{\emph{Companion
  proceedings of the 2019 world wide web conference}}.
  \bibinfo{pages}{491--500}.
\newblock


\bibitem[\protect\citeauthoryear{Bowman and Dahl}{Bowman and Dahl}{2021}]%
        {bowman2021will}
\bibfield{author}{\bibinfo{person}{Samuel Bowman} {and} \bibinfo{person}{George
  Dahl}.} \bibinfo{year}{2021}\natexlab{}.
\newblock \showarticletitle{What Will it Take to Fix Benchmarking in Natural
  Language Understanding?}. In \bibinfo{booktitle}{\emph{Proceedings of the
  2021 Conference of the North American Chapter of the Association for
  Computational Linguistics: Human Language Technologies}}.
  \bibinfo{pages}{4843--4855}.
\newblock


\bibitem[\protect\citeauthoryear{Breck, Cai, Nielsen, Salib, and Sculley}{Breck
  et~al\mbox{.}}{2017}]%
        {breck2017ml}
\bibfield{author}{\bibinfo{person}{Eric Breck}, \bibinfo{person}{Shanqing Cai},
  \bibinfo{person}{Eric Nielsen}, \bibinfo{person}{Michael Salib}, {and}
  \bibinfo{person}{D Sculley}.} \bibinfo{year}{2017}\natexlab{}.
\newblock \showarticletitle{The ML test score: A rubric for ML production
  readiness and technical debt reduction}. In \bibinfo{booktitle}{\emph{2017
  IEEE International Conference on Big Data (Big Data)}}. IEEE,
  \bibinfo{pages}{1123--1132}.
\newblock


\bibitem[\protect\citeauthoryear{Breiman}{Breiman}{2001}]%
        {breiman2001statistical}
\bibfield{author}{\bibinfo{person}{Leo Breiman}.}
  \bibinfo{year}{2001}\natexlab{}.
\newblock \showarticletitle{Statistical modeling: The two cultures (with
  comments and a rejoinder by the author)}.
\newblock \bibinfo{journal}{\emph{Statistical science}} \bibinfo{volume}{16},
  \bibinfo{number}{3} (\bibinfo{year}{2001}), \bibinfo{pages}{199--231}.
\newblock


\bibitem[\protect\citeauthoryear{Brewster}{Brewster}{1881}]%
        {brewster1881}
\bibfield{author}{\bibinfo{person}{Benjamin Brewster}.}
  \bibinfo{year}{1881}\natexlab{}.
\newblock \bibinfo{journal}{\emph{The Yale Literary Magazine}}
  \bibinfo{volume}{October 1881--June 1882} (\bibinfo{year}{1881}).
\newblock


\bibitem[\protect\citeauthoryear{Bulleit, Schmidt, Alvi, Nelson, and
  Rodriguez-Nikl}{Bulleit et~al\mbox{.}}{2015}]%
        {bulleit2015philosophy}
\bibfield{author}{\bibinfo{person}{William Bulleit}, \bibinfo{person}{Jon
  Schmidt}, \bibinfo{person}{Irfan Alvi}, \bibinfo{person}{Erik Nelson}, {and}
  \bibinfo{person}{Tonatiuh Rodriguez-Nikl}.} \bibinfo{year}{2015}\natexlab{}.
\newblock \showarticletitle{Philosophy of engineering: What it is and why it
  matters}.
\newblock \bibinfo{journal}{\emph{Journal of Professional Issues in Engineering
  Education and Practice}} \bibinfo{volume}{141}, \bibinfo{number}{3}
  (\bibinfo{year}{2015}), \bibinfo{pages}{02514003}.
\newblock


\bibitem[\protect\citeauthoryear{Bunescu and Huang}{Bunescu and Huang}{2010}]%
        {bunescu2010utility}
\bibfield{author}{\bibinfo{person}{Razvan Bunescu} {and}
  \bibinfo{person}{Yunfeng Huang}.} \bibinfo{year}{2010}\natexlab{}.
\newblock \showarticletitle{A utility-driven approach to question ranking in
  social QA}. In \bibinfo{booktitle}{\emph{Proceedings of The 23rd
  International Conference on Computational Linguistics (COLING 2010)}}.
  \bibinfo{pages}{125--133}.
\newblock


\bibitem[\protect\citeauthoryear{Cai, Juan, Stamoulis, and Marculescu}{Cai
  et~al\mbox{.}}{2017}]%
        {cai2017neuralpower}
\bibfield{author}{\bibinfo{person}{Ermao Cai}, \bibinfo{person}{Da-Cheng Juan},
  \bibinfo{person}{Dimitrios Stamoulis}, {and} \bibinfo{person}{Diana
  Marculescu}.} \bibinfo{year}{2017}\natexlab{}.
\newblock \showarticletitle{{NeuralPower}: Predict and deploy energy-efficient
  convolutional neural networks}. In \bibinfo{booktitle}{\emph{Asian Conference
  on Machine Learning}}. PMLR, \bibinfo{pages}{622--637}.
\newblock


\bibitem[\protect\citeauthoryear{Card and Smith}{Card and Smith}{2020}]%
        {card2020consequentialism}
\bibfield{author}{\bibinfo{person}{Dallas Card} {and} \bibinfo{person}{Noah~A
  Smith}.} \bibinfo{year}{2020}\natexlab{}.
\newblock \showarticletitle{On Consequentialism and Fairness}.
\newblock \bibinfo{journal}{\emph{Frontiers in Artificial Intelligence}}
  \bibinfo{volume}{3} (\bibinfo{year}{2020}), \bibinfo{pages}{34}.
\newblock


\bibitem[\protect\citeauthoryear{Carlini, Tramer, Wallace, Jagielski,
  Herbert-Voss, Lee, Roberts, Brown, Song, Erlingsson, et~al\mbox{.}}{Carlini
  et~al\mbox{.}}{2021}]%
        {carlini2021extracting}
\bibfield{author}{\bibinfo{person}{Nicholas Carlini}, \bibinfo{person}{Florian
  Tramer}, \bibinfo{person}{Eric Wallace}, \bibinfo{person}{Matthew Jagielski},
  \bibinfo{person}{Ariel Herbert-Voss}, \bibinfo{person}{Katherine Lee},
  \bibinfo{person}{Adam Roberts}, \bibinfo{person}{Tom Brown},
  \bibinfo{person}{Dawn Song}, \bibinfo{person}{Ulfar Erlingsson},
  {et~al\mbox{.}}} \bibinfo{year}{2021}\natexlab{}.
\newblock \showarticletitle{Extracting training data from large language
  models}. In \bibinfo{booktitle}{\emph{30th USENIX Security Symposium (USENIX
  Security 21)}}. \bibinfo{pages}{2633--2650}.
\newblock


\bibitem[\protect\citeauthoryear{Carter, Jain, Mueller, and Gifford}{Carter
  et~al\mbox{.}}{2021}]%
        {carter2021overinterpretation}
\bibfield{author}{\bibinfo{person}{Brandon Carter}, \bibinfo{person}{Siddhartha
  Jain}, \bibinfo{person}{Jonas~W Mueller}, {and} \bibinfo{person}{David
  Gifford}.} \bibinfo{year}{2021}\natexlab{}.
\newblock \showarticletitle{Overinterpretation reveals image classification
  model pathologies}.
\newblock \bibinfo{journal}{\emph{Advances in Neural Information Processing
  Systems}}  \bibinfo{volume}{34} (\bibinfo{year}{2021}).
\newblock


\bibitem[\protect\citeauthoryear{Challen, Denny, Pitt, Gompels, Edwards, and
  Tsaneva-Atanasova}{Challen et~al\mbox{.}}{2019}]%
        {challen2019artificial}
\bibfield{author}{\bibinfo{person}{Robert Challen}, \bibinfo{person}{Joshua
  Denny}, \bibinfo{person}{Martin Pitt}, \bibinfo{person}{Luke Gompels},
  \bibinfo{person}{Tom Edwards}, {and} \bibinfo{person}{Krasimira
  Tsaneva-Atanasova}.} \bibinfo{year}{2019}\natexlab{}.
\newblock \showarticletitle{Artificial intelligence, bias and clinical safety}.
\newblock \bibinfo{journal}{\emph{BMJ Quality \& Safety}} \bibinfo{volume}{28},
  \bibinfo{number}{3} (\bibinfo{year}{2019}), \bibinfo{pages}{231--237}.
\newblock


\bibitem[\protect\citeauthoryear{Charlton}{Charlton}{1998}]%
        {charlton1998nothing}
\bibfield{author}{\bibinfo{person}{James~I Charlton}.}
  \bibinfo{year}{1998}\natexlab{}.
\newblock \bibinfo{booktitle}{\emph{Nothing about us without us}}.
\newblock \bibinfo{publisher}{University of California Press}.
\newblock


\bibitem[\protect\citeauthoryear{Chen, Goel, Sohoni, Poms, Fatahalian, and
  R{\'e}}{Chen et~al\mbox{.}}{2021}]%
        {chen2021mandoline}
\bibfield{author}{\bibinfo{person}{Mayee Chen}, \bibinfo{person}{Karan Goel},
  \bibinfo{person}{Nimit~S Sohoni}, \bibinfo{person}{Fait Poms},
  \bibinfo{person}{Kayvon Fatahalian}, {and} \bibinfo{person}{Christopher
  R{\'e}}.} \bibinfo{year}{2021}\natexlab{}.
\newblock \showarticletitle{Mandoline: Model Evaluation under Distribution
  Shift}. In \bibinfo{booktitle}{\emph{International Conference on Machine
  Learning}}. PMLR, \bibinfo{pages}{1617--1629}.
\newblock


\bibitem[\protect\citeauthoryear{Chiril, Moriceau, Benamara, Mari, Origgi, and
  Coulomb-Gully}{Chiril et~al\mbox{.}}{2020}]%
        {chiril2020he}
\bibfield{author}{\bibinfo{person}{Patricia Chiril},
  \bibinfo{person}{V{\'e}ronique Moriceau}, \bibinfo{person}{Farah Benamara},
  \bibinfo{person}{Alda Mari}, \bibinfo{person}{Gloria Origgi}, {and}
  \bibinfo{person}{Marl{\`e}ne Coulomb-Gully}.}
  \bibinfo{year}{2020}\natexlab{}.
\newblock \showarticletitle{He said “who’s gonna take care of your children
  when you are at ACL?”: Reported Sexist Acts are Not Sexist}. In
  \bibinfo{booktitle}{\emph{Proceedings of the 58th Annual Meeting of the
  Association for Computational Linguistics}}. \bibinfo{pages}{4055--4066}.
\newblock


\bibitem[\protect\citeauthoryear{Chohlas-Wood, Coots, Brunskill, and
  Goel}{Chohlas-Wood et~al\mbox{.}}{2021}]%
        {chohlas2021learning}
\bibfield{author}{\bibinfo{person}{Alex Chohlas-Wood}, \bibinfo{person}{Madison
  Coots}, \bibinfo{person}{Emma Brunskill}, {and} \bibinfo{person}{Sharad
  Goel}.} \bibinfo{year}{2021}\natexlab{}.
\newblock \showarticletitle{Learning to be Fair: A Consequentialist Approach to
  Equitable Decision-Making}.
\newblock \bibinfo{journal}{\emph{arXiv preprint arXiv:2109.08792}}
  (\bibinfo{year}{2021}).
\newblock


\bibitem[\protect\citeauthoryear{Chouldechova}{Chouldechova}{2017}]%
        {chouldechova2017fair}
\bibfield{author}{\bibinfo{person}{Alexandra Chouldechova}.}
  \bibinfo{year}{2017}\natexlab{}.
\newblock \showarticletitle{Fair prediction with disparate impact: A study of
  bias in recidivism prediction instruments}.
\newblock \bibinfo{journal}{\emph{Big data}} \bibinfo{volume}{5},
  \bibinfo{number}{2} (\bibinfo{year}{2017}), \bibinfo{pages}{153--163}.
\newblock


\bibitem[\protect\citeauthoryear{Corbett-Davies and Goel}{Corbett-Davies and
  Goel}{2018}]%
        {corbett2018measure}
\bibfield{author}{\bibinfo{person}{Sam Corbett-Davies} {and}
  \bibinfo{person}{Sharad Goel}.} \bibinfo{year}{2018}\natexlab{}.
\newblock \showarticletitle{The measure and mismeasure of fairness: A critical
  review of fair machine learning}.
\newblock \bibinfo{journal}{\emph{arXiv preprint arXiv:1808.00023}}
  (\bibinfo{year}{2018}).
\newblock


\bibitem[\protect\citeauthoryear{Corbett-Davies, Pierson, Feller, Goel, and
  Huq}{Corbett-Davies et~al\mbox{.}}{2017}]%
        {corbett2017algorithmic}
\bibfield{author}{\bibinfo{person}{Sam Corbett-Davies}, \bibinfo{person}{Emma
  Pierson}, \bibinfo{person}{Avi Feller}, \bibinfo{person}{Sharad Goel}, {and}
  \bibinfo{person}{Aziz Huq}.} \bibinfo{year}{2017}\natexlab{}.
\newblock \showarticletitle{Algorithmic decision making and the cost of
  fairness}. In \bibinfo{booktitle}{\emph{Proceedings of the 23rd acm sigkdd
  international conference on knowledge discovery and data mining}}.
  \bibinfo{pages}{797--806}.
\newblock


\bibitem[\protect\citeauthoryear{Crawford and Joler}{Crawford and
  Joler}{2018}]%
        {crawford2018anatomy}
\bibfield{author}{\bibinfo{person}{Kate Crawford} {and} \bibinfo{person}{Vladan
  Joler}.} \bibinfo{year}{2018}\natexlab{}.
\newblock \bibinfo{title}{Anatomy of an AI System}.
\newblock
\newblock
\newblock
\shownote{(Accessed January, 2022)}.


\bibitem[\protect\citeauthoryear{Crawford and Paglen}{Crawford and
  Paglen}{2021}]%
        {crawford2021excavating}
\bibfield{author}{\bibinfo{person}{Kate Crawford} {and} \bibinfo{person}{Trevor
  Paglen}.} \bibinfo{year}{2021}\natexlab{}.
\newblock \showarticletitle{Excavating AI: The politics of images in machine
  learning training sets}.
\newblock \bibinfo{journal}{\emph{AI \& SOCIETY}} (\bibinfo{year}{2021}),
  \bibinfo{pages}{1--12}.
\newblock


\bibitem[\protect\citeauthoryear{Cunningham, Gambo, Lawless, Moore, Yilmaz,
  Clarke, and O’Connor}{Cunningham et~al\mbox{.}}{2019}]%
        {cunningham2019software}
\bibfield{author}{\bibinfo{person}{Sean Cunningham}, \bibinfo{person}{Jemil
  Gambo}, \bibinfo{person}{Aidan Lawless}, \bibinfo{person}{Declan Moore},
  \bibinfo{person}{Murat Yilmaz}, \bibinfo{person}{Paul~M Clarke}, {and}
  \bibinfo{person}{Rory~V O’Connor}.} \bibinfo{year}{2019}\natexlab{}.
\newblock \showarticletitle{Software testing: a changing career}. In
  \bibinfo{booktitle}{\emph{European Conference on Software Process
  Improvement}}. Springer, \bibinfo{pages}{731--742}.
\newblock


\bibitem[\protect\citeauthoryear{Dahlin}{Dahlin}{2021}]%
        {dahlin2021mind}
\bibfield{author}{\bibinfo{person}{Emma Dahlin}.}
  \bibinfo{year}{2021}\natexlab{}.
\newblock \showarticletitle{Mind the gap! On the future of {AI} research}.
\newblock \bibinfo{journal}{\emph{Humanities and Social Sciences
  Communications}} \bibinfo{volume}{8}, \bibinfo{number}{1}
  (\bibinfo{year}{2021}), \bibinfo{pages}{1--4}.
\newblock


\bibitem[\protect\citeauthoryear{D'Amour, Heller, Moldovan, Adlam, Alipanahi,
  Beutel, Chen, Deaton, Eisenstein, Hoffman, et~al\mbox{.}}{D'Amour
  et~al\mbox{.}}{2020}]%
        {d2020underspecification}
\bibfield{author}{\bibinfo{person}{Alexander D'Amour},
  \bibinfo{person}{Katherine Heller}, \bibinfo{person}{Dan Moldovan},
  \bibinfo{person}{Ben Adlam}, \bibinfo{person}{Babak Alipanahi},
  \bibinfo{person}{Alex Beutel}, \bibinfo{person}{Christina Chen},
  \bibinfo{person}{Jonathan Deaton}, \bibinfo{person}{Jacob Eisenstein},
  \bibinfo{person}{Matthew~D Hoffman}, {et~al\mbox{.}}}
  \bibinfo{year}{2020}\natexlab{}.
\newblock \showarticletitle{Underspecification presents challenges for
  credibility in modern machine learning}.
\newblock \bibinfo{journal}{\emph{arXiv preprint arXiv:2011.03395}}
  (\bibinfo{year}{2020}).
\newblock


\bibitem[\protect\citeauthoryear{Davani, D{\'\i}az, and Prabhakaran}{Davani
  et~al\mbox{.}}{2022}]%
        {davani2022dealing}
\bibfield{author}{\bibinfo{person}{Aida~Mostafazadeh Davani},
  \bibinfo{person}{Mark D{\'\i}az}, {and} \bibinfo{person}{Vinodkumar
  Prabhakaran}.} \bibinfo{year}{2022}\natexlab{}.
\newblock \showarticletitle{Dealing with disagreements: Looking beyond the
  majority vote in subjective annotations}.
\newblock \bibinfo{journal}{\emph{Transactions of the Association for
  Computational Linguistics}}  \bibinfo{volume}{10} (\bibinfo{year}{2022}),
  \bibinfo{pages}{92--110}.
\newblock


\bibitem[\protect\citeauthoryear{De~Vries, Bahdanau, and Manning}{De~Vries
  et~al\mbox{.}}{2020}]%
        {de2020towards}
\bibfield{author}{\bibinfo{person}{Harm De~Vries}, \bibinfo{person}{Dzmitry
  Bahdanau}, {and} \bibinfo{person}{Christopher Manning}.}
  \bibinfo{year}{2020}\natexlab{}.
\newblock \showarticletitle{Towards ecologically valid research on language
  user interfaces}.
\newblock \bibinfo{journal}{\emph{arXiv preprint arXiv:2007.14435}}
  (\bibinfo{year}{2020}).
\newblock


\bibitem[\protect\citeauthoryear{Derczynski}{Derczynski}{2016}]%
        {derczynski2016complementarity}
\bibfield{author}{\bibinfo{person}{Leon Derczynski}.}
  \bibinfo{year}{2016}\natexlab{}.
\newblock \showarticletitle{Complementarity, F-score, and NLP Evaluation}. In
  \bibinfo{booktitle}{\emph{Proceedings of the Tenth International Conference
  on Language Resources and Evaluation (LREC'16)}}. \bibinfo{pages}{261--266}.
\newblock


\bibitem[\protect\citeauthoryear{D{\'\i}az and Diakopoulos}{D{\'\i}az and
  Diakopoulos}{2019}]%
        {diaz2019whose}
\bibfield{author}{\bibinfo{person}{Mark D{\'\i}az} {and}
  \bibinfo{person}{Nicholas Diakopoulos}.} \bibinfo{year}{2019}\natexlab{}.
\newblock \showarticletitle{Whose walkability?: {Challenges} in algorithmically
  measuring subjective experience}.
\newblock \bibinfo{journal}{\emph{Proceedings of the ACM on Human-Computer
  Interaction}} \bibinfo{volume}{3}, \bibinfo{number}{CSCW}
  (\bibinfo{year}{2019}), \bibinfo{pages}{1--22}.
\newblock


\bibitem[\protect\citeauthoryear{Eckhouse, Lum, Conti-Cook, and
  Ciccolini}{Eckhouse et~al\mbox{.}}{2019}]%
        {eckhouse2019layers}
\bibfield{author}{\bibinfo{person}{Laurel Eckhouse}, \bibinfo{person}{Kristian
  Lum}, \bibinfo{person}{Cynthia Conti-Cook}, {and} \bibinfo{person}{Julie
  Ciccolini}.} \bibinfo{year}{2019}\natexlab{}.
\newblock \showarticletitle{Layers of bias: A unified approach for
  understanding problems with risk assessment}.
\newblock \bibinfo{journal}{\emph{Criminal Justice and Behavior}}
  \bibinfo{volume}{46}, \bibinfo{number}{2} (\bibinfo{year}{2019}),
  \bibinfo{pages}{185--209}.
\newblock


\bibitem[\protect\citeauthoryear{Endriss}{Endriss}{2018}]%
        {endriss2018lecture}
\bibfield{author}{\bibinfo{person}{Ulle Endriss}.}
  \bibinfo{year}{2018}\natexlab{}.
\newblock \showarticletitle{Lecture notes on fair division}.
\newblock \bibinfo{journal}{\emph{arXiv preprint arXiv:1806.04234}}
  (\bibinfo{year}{2018}).
\newblock


\bibitem[\protect\citeauthoryear{Ethayarajh and Jurafsky}{Ethayarajh and
  Jurafsky}{2020}]%
        {ethayarajh2020utility}
\bibfield{author}{\bibinfo{person}{Kawin Ethayarajh} {and} \bibinfo{person}{Dan
  Jurafsky}.} \bibinfo{year}{2020}\natexlab{}.
\newblock \showarticletitle{Utility is in the Eye of the User: A Critique of
  NLP Leaderboards}. In \bibinfo{booktitle}{\emph{Proceedings of the 2020
  Conference on Empirical Methods in Natural Language Processing (EMNLP)}}.
  \bibinfo{pages}{4846--4853}.
\newblock


\bibitem[\protect\citeauthoryear{Ettinger, Rao, Daum{\'e}~III, and
  Bender}{Ettinger et~al\mbox{.}}{2017}]%
        {ettinger2017towards}
\bibfield{author}{\bibinfo{person}{Allyson Ettinger}, \bibinfo{person}{Sudha
  Rao}, \bibinfo{person}{Hal Daum{\'e}~III}, {and} \bibinfo{person}{Emily~M
  Bender}.} \bibinfo{year}{2017}\natexlab{}.
\newblock \showarticletitle{Towards linguistically generalizable NLP systems: A
  workshop and shared task}.
\newblock \bibinfo{journal}{\emph{arXiv preprint arXiv:1711.01505}}
  (\bibinfo{year}{2017}).
\newblock


\bibitem[\protect\citeauthoryear{Evci, Dumoulin, Larochelle, and Mozer}{Evci
  et~al\mbox{.}}{2021}]%
        {evci2021head2toe}
\bibfield{author}{\bibinfo{person}{Utku Evci}, \bibinfo{person}{Vincent
  Dumoulin}, \bibinfo{person}{Hugo Larochelle}, {and}
  \bibinfo{person}{Michael~Curtis Mozer}.} \bibinfo{year}{2021}\natexlab{}.
\newblock \showarticletitle{Head2Toe: Utilizing Intermediate Representations
  for Better OOD Generalization}.
\newblock  (\bibinfo{year}{2021}).
\newblock


\bibitem[\protect\citeauthoryear{Farahani, Voghoei, Rasheed, and
  Arabnia}{Farahani et~al\mbox{.}}{2020}]%
        {farahani2020brief}
\bibfield{author}{\bibinfo{person}{Abolfazl Farahani}, \bibinfo{person}{Sahar
  Voghoei}, \bibinfo{person}{Khaled Rasheed}, {and} \bibinfo{person}{Hamid~R
  Arabnia}.} \bibinfo{year}{2020}\natexlab{}.
\newblock \showarticletitle{A brief review of domain adaptation}.
\newblock \bibinfo{journal}{\emph{arXiv preprint arXiv:2010.03978}}
  (\bibinfo{year}{2020}).
\newblock


\bibitem[\protect\citeauthoryear{Fornaciari, Uma, Paun, Plank, Hovy, and
  Poesio}{Fornaciari et~al\mbox{.}}{2021}]%
        {fornaciari2021beyond}
\bibfield{author}{\bibinfo{person}{Tommaso Fornaciari},
  \bibinfo{person}{Alexandra Uma}, \bibinfo{person}{Silviu Paun},
  \bibinfo{person}{Barbara Plank}, \bibinfo{person}{Dirk Hovy}, {and}
  \bibinfo{person}{Massimo Poesio}.} \bibinfo{year}{2021}\natexlab{}.
\newblock \showarticletitle{Beyond Black \& White: Leveraging Annotator
  Disagreement via Soft-Label Multi-Task Learning}. In
  \bibinfo{booktitle}{\emph{Proceedings of the 2021 Conference of the North
  American Chapter of the Association for Computational Linguistics: Human
  Language Technologies}}. \bibinfo{pages}{2591--2597}.
\newblock


\bibitem[\protect\citeauthoryear{Forsythe}{Forsythe}{2001a}]%
        {forsythe2001invents}
\bibfield{author}{\bibinfo{person}{Diana Forsythe}.}
  \bibinfo{year}{2001}\natexlab{a}.
\newblock \bibinfo{booktitle}{\emph{Studying those who study us: An
  anthropologist in the world of Artificial Intelligence}}.
\newblock \bibinfo{publisher}{Stanford University Press}, Chapter Artificial
  intelligence invents itself: Collective identity and boundary maintenance in
  an emergent scientific discipline.
\newblock


\bibitem[\protect\citeauthoryear{Forsythe}{Forsythe}{2001b}]%
        {forsythe2001knowledge}
\bibfield{author}{\bibinfo{person}{Diana Forsythe}.}
  \bibinfo{year}{2001}\natexlab{b}.
\newblock \bibinfo{booktitle}{\emph{Studying those who study us: An
  anthropologist in the world of Artificial Intelligence}}.
\newblock \bibinfo{publisher}{Stanford University Press}, Chapter The
  Construction of Knowledge in Artificial Intelligence.
\newblock


\bibitem[\protect\citeauthoryear{Friedler, Scheidegger, and
  Venkatasubramanian}{Friedler et~al\mbox{.}}{2021}]%
        {friedler2021possibility}
\bibfield{author}{\bibinfo{person}{Sorelle~A Friedler}, \bibinfo{person}{Carlos
  Scheidegger}, {and} \bibinfo{person}{Suresh Venkatasubramanian}.}
  \bibinfo{year}{2021}\natexlab{}.
\newblock \showarticletitle{The (im) possibility of fairness: Different value
  systems require different mechanisms for fair decision making}.
\newblock \bibinfo{journal}{\emph{Commun. ACM}} \bibinfo{volume}{64},
  \bibinfo{number}{4} (\bibinfo{year}{2021}), \bibinfo{pages}{136--143}.
\newblock


\bibitem[\protect\citeauthoryear{Fu, Chen, Henniger, and Damer}{Fu
  et~al\mbox{.}}{2022}]%
        {fu2022deep}
\bibfield{author}{\bibinfo{person}{Biying Fu}, \bibinfo{person}{Cong Chen},
  \bibinfo{person}{Olaf Henniger}, {and} \bibinfo{person}{Naser Damer}.}
  \bibinfo{year}{2022}\natexlab{}.
\newblock \showarticletitle{A deep insight into measuring face image utility
  with general and face-specific image quality metrics}. In
  \bibinfo{booktitle}{\emph{Proceedings of the IEEE/CVF Winter Conference on
  Applications of Computer Vision}}. \bibinfo{pages}{905--914}.
\newblock


\bibitem[\protect\citeauthoryear{Garc{\'\i}a-Mart{\'\i}n, Rodrigues, Riley, and
  Grahn}{Garc{\'\i}a-Mart{\'\i}n et~al\mbox{.}}{2019}]%
        {garcia2019estimation}
\bibfield{author}{\bibinfo{person}{Eva Garc{\'\i}a-Mart{\'\i}n},
  \bibinfo{person}{Crefeda~Faviola Rodrigues}, \bibinfo{person}{Graham Riley},
  {and} \bibinfo{person}{H{\aa}kan Grahn}.} \bibinfo{year}{2019}\natexlab{}.
\newblock \showarticletitle{Estimation of energy consumption in machine
  learning}.
\newblock \bibinfo{journal}{\emph{J. Parallel and Distrib. Comput.}}
  \bibinfo{volume}{134} (\bibinfo{year}{2019}), \bibinfo{pages}{75--88}.
\newblock


\bibitem[\protect\citeauthoryear{Garg, Perot, Limtiaco, Taly, Chi, and
  Beutel}{Garg et~al\mbox{.}}{2019}]%
        {garg2019counterfactual}
\bibfield{author}{\bibinfo{person}{Sahaj Garg}, \bibinfo{person}{Vincent
  Perot}, \bibinfo{person}{Nicole Limtiaco}, \bibinfo{person}{Ankur Taly},
  \bibinfo{person}{Ed~H Chi}, {and} \bibinfo{person}{Alex Beutel}.}
  \bibinfo{year}{2019}\natexlab{}.
\newblock \showarticletitle{Counterfactual fairness in text classification
  through robustness}. In \bibinfo{booktitle}{\emph{Proceedings of the 2019
  AAAI/ACM Conference on AI, Ethics, and Society}}. \bibinfo{pages}{219--226}.
\newblock


\bibitem[\protect\citeauthoryear{Gebru, Morgenstern, Vecchione, Vaughan,
  Wallach, Iii, and Crawford}{Gebru et~al\mbox{.}}{2021}]%
        {gebru2021datasheets}
\bibfield{author}{\bibinfo{person}{Timnit Gebru}, \bibinfo{person}{Jamie
  Morgenstern}, \bibinfo{person}{Briana Vecchione},
  \bibinfo{person}{Jennifer~Wortman Vaughan}, \bibinfo{person}{Hanna Wallach},
  \bibinfo{person}{Hal~Daum{\'e} Iii}, {and} \bibinfo{person}{Kate Crawford}.}
  \bibinfo{year}{2021}\natexlab{}.
\newblock \showarticletitle{Datasheets for datasets}.
\newblock \bibinfo{journal}{\emph{Commun. ACM}} \bibinfo{volume}{64},
  \bibinfo{number}{12} (\bibinfo{year}{2021}), \bibinfo{pages}{86--92}.
\newblock


\bibitem[\protect\citeauthoryear{Geertz}{Geertz}{1973}]%
        {geertz1973}
\bibfield{author}{\bibinfo{person}{Clifford Geertz}.}
  \bibinfo{year}{1973}\natexlab{}.
\newblock \bibinfo{booktitle}{\emph{The Interpretation of Cultures}}.
\newblock \bibinfo{publisher}{Basic Books}.
\newblock


\bibitem[\protect\citeauthoryear{Geiger}{Geiger}{2020}]%
        {geiger2020silicon}
\bibfield{author}{\bibinfo{person}{Susi Geiger}.}
  \bibinfo{year}{2020}\natexlab{}.
\newblock \showarticletitle{Silicon Valley, disruption, and the end of
  uncertainty}.
\newblock \bibinfo{journal}{\emph{Journal of cultural economy}}
  \bibinfo{volume}{13}, \bibinfo{number}{2} (\bibinfo{year}{2020}),
  \bibinfo{pages}{169--184}.
\newblock


\bibitem[\protect\citeauthoryear{George and Williams}{George and
  Williams}{2004}]%
        {george2004structured}
\bibfield{author}{\bibinfo{person}{Boby George} {and} \bibinfo{person}{Laurie
  Williams}.} \bibinfo{year}{2004}\natexlab{}.
\newblock \showarticletitle{A structured experiment of test-driven
  development}.
\newblock \bibinfo{journal}{\emph{Information and software Technology}}
  \bibinfo{volume}{46}, \bibinfo{number}{5} (\bibinfo{year}{2004}),
  \bibinfo{pages}{337--342}.
\newblock


\bibitem[\protect\citeauthoryear{Gonen and Webster}{Gonen and Webster}{2020}]%
        {gonen2020automatically}
\bibfield{author}{\bibinfo{person}{Hila Gonen} {and} \bibinfo{person}{Kellie
  Webster}.} \bibinfo{year}{2020}\natexlab{}.
\newblock \showarticletitle{Automatically Identifying Gender Issues in Machine
  Translation using Perturbations}. In \bibinfo{booktitle}{\emph{Findings of
  the Association for Computational Linguistics: EMNLP 2020}}.
  \bibinfo{pages}{1991--1995}.
\newblock


\bibitem[\protect\citeauthoryear{Gray and Suri}{Gray and Suri}{2019}]%
        {gray2019ghost}
\bibfield{author}{\bibinfo{person}{Mary~L Gray} {and}
  \bibinfo{person}{Siddharth Suri}.} \bibinfo{year}{2019}\natexlab{}.
\newblock \bibinfo{booktitle}{\emph{Ghost work: How to stop Silicon Valley from
  building a new global underclass}}.
\newblock \bibinfo{publisher}{Eamon Dolan Books}.
\newblock


\bibitem[\protect\citeauthoryear{Green}{Green}{2020}]%
        {green2020data}
\bibfield{author}{\bibinfo{person}{Ben Green}.}
  \bibinfo{year}{2020}\natexlab{}.
\newblock \showarticletitle{Data science as political action: grounding data
  science in a politics of justice}.
\newblock \bibinfo{journal}{\emph{Available at SSRN 3658431}}
  (\bibinfo{year}{2020}).
\newblock


\bibitem[\protect\citeauthoryear{Ha, Gupta, Rana, and Venkatesh}{Ha
  et~al\mbox{.}}{2021}]%
        {ha2021alt}
\bibfield{author}{\bibinfo{person}{Huong Ha}, \bibinfo{person}{Sunil Gupta},
  \bibinfo{person}{Santu Rana}, {and} \bibinfo{person}{Svetha Venkatesh}.}
  \bibinfo{year}{2021}\natexlab{}.
\newblock \showarticletitle{ALT-MAS: A Data-Efficient Framework for Active
  Testing of Machine Learning Algorithms}.
\newblock \bibinfo{journal}{\emph{arXiv preprint arXiv:2104.04999}}
  (\bibinfo{year}{2021}).
\newblock


\bibitem[\protect\citeauthoryear{Hampel and Zurich}{Hampel and Zurich}{1998}]%
        {hampel1998statistics}
\bibfield{author}{\bibinfo{person}{Frank Hampel} {and} \bibinfo{person}{Eth
  Zurich}.} \bibinfo{year}{1998}\natexlab{}.
\newblock \showarticletitle{Is statistics too difficult?}
\newblock \bibinfo{journal}{\emph{Canadian Journal of Statistics}}
  \bibinfo{volume}{26}, \bibinfo{number}{3} (\bibinfo{year}{1998}),
  \bibinfo{pages}{497--513}.
\newblock


\bibitem[\protect\citeauthoryear{Hardt, Price, and Srebro}{Hardt
  et~al\mbox{.}}{2016}]%
        {hardt2016equality}
\bibfield{author}{\bibinfo{person}{Moritz Hardt}, \bibinfo{person}{Eric Price},
  {and} \bibinfo{person}{Nati Srebro}.} \bibinfo{year}{2016}\natexlab{}.
\newblock \showarticletitle{Equality of opportunity in supervised learning}.
\newblock \bibinfo{journal}{\emph{Advances in neural information processing
  systems}}  \bibinfo{volume}{29} (\bibinfo{year}{2016}),
  \bibinfo{pages}{3315--3323}.
\newblock


\bibitem[\protect\citeauthoryear{Harrold}{Harrold}{2000}]%
        {harrold2000testing}
\bibfield{author}{\bibinfo{person}{Mary~Jean Harrold}.}
  \bibinfo{year}{2000}\natexlab{}.
\newblock \showarticletitle{Testing: a roadmap}. In
  \bibinfo{booktitle}{\emph{Proceedings of the Conference on the Future of
  Software Engineering}}. \bibinfo{pages}{61--72}.
\newblock


\bibitem[\protect\citeauthoryear{He, Zhang, Ren, and Sun}{He
  et~al\mbox{.}}{2016}]%
        {he2016deep}
\bibfield{author}{\bibinfo{person}{Kaiming He}, \bibinfo{person}{Xiangyu
  Zhang}, \bibinfo{person}{Shaoqing Ren}, {and} \bibinfo{person}{Jian Sun}.}
  \bibinfo{year}{2016}\natexlab{}.
\newblock \showarticletitle{Deep residual learning for image recognition}. In
  \bibinfo{booktitle}{\emph{Proceedings of the IEEE conference on computer
  vision and pattern recognition}}. \bibinfo{pages}{770--778}.
\newblock


\bibitem[\protect\citeauthoryear{Heldreth, Lahav, Mengesha, Sublewski, and
  Tuennerman}{Heldreth et~al\mbox{.}}{2021}]%
        {heldreth2021don}
\bibfield{author}{\bibinfo{person}{Courtney Heldreth}, \bibinfo{person}{Michal
  Lahav}, \bibinfo{person}{Zion Mengesha}, \bibinfo{person}{Juliana Sublewski},
  {and} \bibinfo{person}{Elyse Tuennerman}.} \bibinfo{year}{2021}\natexlab{}.
\newblock \showarticletitle{``I don't think these devices are very culturally
  sensitive.''---The impact of errors on African Americans in Automated Speech
  Recognition}.
\newblock \bibinfo{journal}{\emph{Frontiers in Artificial Intelligence}}
  \bibinfo{volume}{26} (\bibinfo{year}{2021}).
\newblock


\bibitem[\protect\citeauthoryear{Henderson, Hu, Romoff, Brunskill, Jurafsky,
  and Pineau}{Henderson et~al\mbox{.}}{2020}]%
        {henderson2020towards}
\bibfield{author}{\bibinfo{person}{Peter Henderson}, \bibinfo{person}{Jieru
  Hu}, \bibinfo{person}{Joshua Romoff}, \bibinfo{person}{Emma Brunskill},
  \bibinfo{person}{Dan Jurafsky}, {and} \bibinfo{person}{Joelle Pineau}.}
  \bibinfo{year}{2020}\natexlab{}.
\newblock \showarticletitle{Towards the systematic reporting of the energy and
  carbon footprints of machine learning}.
\newblock \bibinfo{journal}{\emph{Journal of Machine Learning Research}}
  \bibinfo{volume}{21}, \bibinfo{number}{248} (\bibinfo{year}{2020}),
  \bibinfo{pages}{1--43}.
\newblock


\bibitem[\protect\citeauthoryear{Hepp, Dey, Sinha, Kapoor, Joshi, and
  Hilliges}{Hepp et~al\mbox{.}}{2018}]%
        {hepp2018learn}
\bibfield{author}{\bibinfo{person}{Benjamin Hepp}, \bibinfo{person}{Debadeepta
  Dey}, \bibinfo{person}{Sudipta~N Sinha}, \bibinfo{person}{Ashish Kapoor},
  \bibinfo{person}{Neel Joshi}, {and} \bibinfo{person}{Otmar Hilliges}.}
  \bibinfo{year}{2018}\natexlab{}.
\newblock \showarticletitle{Learn-to-score: Efficient {3D} scene exploration by
  predicting view utility}. In \bibinfo{booktitle}{\emph{Proceedings of the
  European conference on computer vision (ECCV)}}. \bibinfo{pages}{437--452}.
\newblock


\bibitem[\protect\citeauthoryear{Holstein, Wortman~Vaughan, Daum{\'e}~III,
  Dudik, and Wallach}{Holstein et~al\mbox{.}}{2019}]%
        {holstein2019improving}
\bibfield{author}{\bibinfo{person}{Kenneth Holstein}, \bibinfo{person}{Jennifer
  Wortman~Vaughan}, \bibinfo{person}{Hal Daum{\'e}~III}, \bibinfo{person}{Miro
  Dudik}, {and} \bibinfo{person}{Hanna Wallach}.}
  \bibinfo{year}{2019}\natexlab{}.
\newblock \showarticletitle{Improving fairness in machine learning systems:
  What do industry practitioners need?}. In
  \bibinfo{booktitle}{\emph{Proceedings of the 2019 CHI conference on human
  factors in computing systems}}. \bibinfo{pages}{1--16}.
\newblock


\bibitem[\protect\citeauthoryear{Hooker}{Hooker}{1995}]%
        {hooker1995testing}
\bibfield{author}{\bibinfo{person}{John~N Hooker}.}
  \bibinfo{year}{1995}\natexlab{}.
\newblock \showarticletitle{Testing heuristics: We have it all wrong}.
\newblock \bibinfo{journal}{\emph{Journal of heuristics}} \bibinfo{volume}{1},
  \bibinfo{number}{1} (\bibinfo{year}{1995}), \bibinfo{pages}{33--42}.
\newblock


\bibitem[\protect\citeauthoryear{Hovy and Spruit}{Hovy and Spruit}{2016}]%
        {hovy2016social}
\bibfield{author}{\bibinfo{person}{Dirk Hovy} {and} \bibinfo{person}{Shannon~L
  Spruit}.} \bibinfo{year}{2016}\natexlab{}.
\newblock \showarticletitle{The social impact of natural language processing}.
  In \bibinfo{booktitle}{\emph{Proceedings of the 54th Annual Meeting of the
  Association for Computational Linguistics (Volume 2: Short Papers)}}.
  \bibinfo{pages}{591--598}.
\newblock


\bibitem[\protect\citeauthoryear{Huang, Zhang, Jiang, Stanforth, Welbl, Rae,
  Maini, Yogatama, and Kohli}{Huang et~al\mbox{.}}{2020}]%
        {huang2020reducing}
\bibfield{author}{\bibinfo{person}{Po-Sen Huang}, \bibinfo{person}{Huan Zhang},
  \bibinfo{person}{Ray Jiang}, \bibinfo{person}{Robert Stanforth},
  \bibinfo{person}{Johannes Welbl}, \bibinfo{person}{Jack Rae},
  \bibinfo{person}{Vishal Maini}, \bibinfo{person}{Dani Yogatama}, {and}
  \bibinfo{person}{Pushmeet Kohli}.} \bibinfo{year}{2020}\natexlab{}.
\newblock \showarticletitle{Reducing Sentiment Bias in Language Models via
  Counterfactual Evaluation}. In \bibinfo{booktitle}{\emph{Findings of the
  Association for Computational Linguistics: EMNLP 2020}}.
  \bibinfo{pages}{65--83}.
\newblock


\bibitem[\protect\citeauthoryear{Hutchinson and Mitchell}{Hutchinson and
  Mitchell}{2019}]%
        {hutchinson201950}
\bibfield{author}{\bibinfo{person}{Ben Hutchinson} {and}
  \bibinfo{person}{Margaret Mitchell}.} \bibinfo{year}{2019}\natexlab{}.
\newblock \showarticletitle{50 years of test (un) fairness: Lessons for machine
  learning}. In \bibinfo{booktitle}{\emph{Proceedings of the Conference on
  Fairness, Accountability, and Transparency}}. \bibinfo{pages}{49--58}.
\newblock


\bibitem[\protect\citeauthoryear{Hutchinson, Smart, Hanna, Denton, Greer,
  Kjartansson, Barnes, and Mitchell}{Hutchinson et~al\mbox{.}}{2021}]%
        {hutchinson2021towards}
\bibfield{author}{\bibinfo{person}{Ben Hutchinson}, \bibinfo{person}{Andrew
  Smart}, \bibinfo{person}{Alex Hanna}, \bibinfo{person}{Emily Denton},
  \bibinfo{person}{Christina Greer}, \bibinfo{person}{Oddur Kjartansson},
  \bibinfo{person}{Parker Barnes}, {and} \bibinfo{person}{Margaret Mitchell}.}
  \bibinfo{year}{2021}\natexlab{}.
\newblock \showarticletitle{Towards accountability for machine learning
  datasets: Practices from software engineering and infrastructure}. In
  \bibinfo{booktitle}{\emph{Proceedings of the 2021 ACM Conference on Fairness,
  Accountability, and Transparency}}. \bibinfo{pages}{560--575}.
\newblock


\bibitem[\protect\citeauthoryear{Idahl, Lyu, Gadiraju, and Anand}{Idahl
  et~al\mbox{.}}{2021}]%
        {idahl2021towards}
\bibfield{author}{\bibinfo{person}{Maximilian Idahl}, \bibinfo{person}{Lijun
  Lyu}, \bibinfo{person}{Ujwal Gadiraju}, {and} \bibinfo{person}{Avishek
  Anand}.} \bibinfo{year}{2021}\natexlab{}.
\newblock \showarticletitle{Towards Benchmarking the Utility of Explanations
  for Model Debugging}. In \bibinfo{booktitle}{\emph{Proceedings of the First
  Workshop on Trustworthy Natural Language Processing}}.
  \bibinfo{pages}{68--73}.
\newblock


\bibitem[\protect\citeauthoryear{{IEEE}}{{IEEE}}{2019}]%
        {ieeeglobal2019}
\bibfield{author}{\bibinfo{person}{{IEEE}}.} \bibinfo{year}{2019}\natexlab{}.
\newblock \showarticletitle{{The IEEE Global Initiative on Ethics of Autonomous
  and Intelligent Systems. ``Classical Ethics in A/IS''}}.
\newblock In \bibinfo{booktitle}{\emph{{Ethically Aligned Design: A Vision for
  Prioritizing Human Well-being with Autonomous and Intelligent Systems, First
  Edition}}}. \bibinfo{pages}{36--67}.
\newblock


\bibitem[\protect\citeauthoryear{Jacobs, Blodgett, Barocas, Daum{\'e}~III, and
  Wallach}{Jacobs et~al\mbox{.}}{2020}]%
        {jacobs2020meaning}
\bibfield{author}{\bibinfo{person}{Abigail~Z Jacobs}, \bibinfo{person}{Su~Lin
  Blodgett}, \bibinfo{person}{Solon Barocas}, \bibinfo{person}{Hal
  Daum{\'e}~III}, {and} \bibinfo{person}{Hanna Wallach}.}
  \bibinfo{year}{2020}\natexlab{}.
\newblock \showarticletitle{The meaning and measurement of bias: lessons from
  natural language processing}. In \bibinfo{booktitle}{\emph{Proceedings of the
  2020 Conference on Fairness, Accountability, and Transparency}}.
  \bibinfo{pages}{706--706}.
\newblock


\bibitem[\protect\citeauthoryear{Jacobs and Wallach}{Jacobs and
  Wallach}{2021}]%
        {jacobs2021measurement}
\bibfield{author}{\bibinfo{person}{Abigail~Z Jacobs} {and}
  \bibinfo{person}{Hanna Wallach}.} \bibinfo{year}{2021}\natexlab{}.
\newblock \showarticletitle{Measurement and fairness}. In
  \bibinfo{booktitle}{\emph{Proceedings of the 2021 ACM Conference on Fairness,
  Accountability, and Transparency}}. \bibinfo{pages}{375--385}.
\newblock


\bibitem[\protect\citeauthoryear{Jafarian and Park}{Jafarian and Park}{2021}]%
        {jafarian2021learning}
\bibfield{author}{\bibinfo{person}{Yasamin Jafarian} {and}
  \bibinfo{person}{Hyun~Soo Park}.} \bibinfo{year}{2021}\natexlab{}.
\newblock \showarticletitle{Learning high fidelity depths of dressed humans by
  watching social media dance videos}. In \bibinfo{booktitle}{\emph{Proceedings
  of the IEEE/CVF Conference on Computer Vision and Pattern Recognition}}.
  \bibinfo{pages}{12753--12762}.
\newblock


\bibitem[\protect\citeauthoryear{Japkowicz}{Japkowicz}{2006}]%
        {japkowicz2006question}
\bibfield{author}{\bibinfo{person}{Nathalie Japkowicz}.}
  \bibinfo{year}{2006}\natexlab{}.
\newblock \showarticletitle{Why question machine learning evaluation methods}.
  In \bibinfo{booktitle}{\emph{AAAI workshop on evaluation methods for machine
  learning}}. \bibinfo{pages}{6--11}.
\newblock


\bibitem[\protect\citeauthoryear{Jappy}{Jappy}{2013}]%
        {jappy2013introduction}
\bibfield{author}{\bibinfo{person}{Tony Jappy}.}
  \bibinfo{year}{2013}\natexlab{}.
\newblock \bibinfo{booktitle}{\emph{Introduction to Peircean visual
  semiotics}}.
\newblock \bibinfo{publisher}{A\&C Black}.
\newblock


\bibitem[\protect\citeauthoryear{Ji, Smyth, and Steyvers}{Ji
  et~al\mbox{.}}{2020}]%
        {ji2020can}
\bibfield{author}{\bibinfo{person}{Disi Ji}, \bibinfo{person}{Padhraic Smyth},
  {and} \bibinfo{person}{Mark Steyvers}.} \bibinfo{year}{2020}\natexlab{}.
\newblock \showarticletitle{Can i trust my fairness metric? assessing fairness
  with unlabeled data and bayesian inference}.
\newblock \bibinfo{journal}{\emph{Advances in Neural Information Processing
  Systems}}  \bibinfo{volume}{33} (\bibinfo{year}{2020}),
  \bibinfo{pages}{18600--18612}.
\newblock


\bibitem[\protect\citeauthoryear{Jones and Galliers}{Jones and
  Galliers}{1995}]%
        {jones1995evaluating}
\bibfield{author}{\bibinfo{person}{Karen~Sparck Jones} {and}
  \bibinfo{person}{Julia~R Galliers}.} \bibinfo{year}{1995}\natexlab{}.
\newblock \bibinfo{booktitle}{\emph{Evaluating natural language processing
  systems: An analysis and review}}. Vol.~\bibinfo{volume}{1083}.
\newblock \bibinfo{publisher}{Springer Science \& Business Media}.
\newblock


\bibitem[\protect\citeauthoryear{Kannan, Roth, and Ziani}{Kannan
  et~al\mbox{.}}{2019}]%
        {kannan2019downstream}
\bibfield{author}{\bibinfo{person}{Sampath Kannan}, \bibinfo{person}{Aaron
  Roth}, {and} \bibinfo{person}{Juba Ziani}.} \bibinfo{year}{2019}\natexlab{}.
\newblock \showarticletitle{Downstream effects of affirmative action}. In
  \bibinfo{booktitle}{\emph{Proceedings of the Conference on Fairness,
  Accountability, and Transparency}}. \bibinfo{pages}{240--248}.
\newblock


\bibitem[\protect\citeauthoryear{Koch, Denton, Hanna, and Foster}{Koch
  et~al\mbox{.}}{2021}]%
        {koch2021reduced}
\bibfield{author}{\bibinfo{person}{Bernard Koch}, \bibinfo{person}{Emily
  Denton}, \bibinfo{person}{Alex Hanna}, {and} \bibinfo{person}{Jacob~G
  Foster}.} \bibinfo{year}{2021}\natexlab{}.
\newblock \showarticletitle{Reduced, Reused and Recycled: The Life of a Dataset
  in Machine Learning Research}.
\newblock \bibinfo{journal}{\emph{{NeurIPS} Dataset \& Benchmark track}}
  (\bibinfo{year}{2021}).
\newblock


\bibitem[\protect\citeauthoryear{Koh, Sagawa, Marklund, Xie, Zhang,
  Balsubramani, Hu, Yasunaga, Phillips, Beery, Leskovec, Kundaje, Pierson,
  Levine, Finn, and Liang}{Koh et~al\mbox{.}}{2020}]%
        {koh2020wilds}
\bibfield{author}{\bibinfo{person}{Pang~Wei Koh}, \bibinfo{person}{Shiori
  Sagawa}, \bibinfo{person}{Henrik Marklund}, \bibinfo{person}{Sang~Michael
  Xie}, \bibinfo{person}{Marvin Zhang}, \bibinfo{person}{Akshay Balsubramani},
  \bibinfo{person}{Weihua Hu}, \bibinfo{person}{Michihiro Yasunaga},
  \bibinfo{person}{Richard~Lanas Phillips}, \bibinfo{person}{Sara Beery},
  \bibinfo{person}{Jure Leskovec}, \bibinfo{person}{Anshul Kundaje},
  \bibinfo{person}{Emma Pierson}, \bibinfo{person}{Sergey Levine},
  \bibinfo{person}{Chelsea Finn}, {and} \bibinfo{person}{Percy Liang}.}
  \bibinfo{year}{2020}\natexlab{}.
\newblock \showarticletitle{WILDS: A Benchmark of in-the-Wild Distribution
  Shifts}.
\newblock \bibinfo{journal}{\emph{CoRR}}  \bibinfo{volume}{abs/2012.07421}
  (\bibinfo{year}{2020}).
\newblock
\urldef\tempurl%
\url{https://arxiv.org/abs/2012.07421}
\showURL{%
\tempurl}


\bibitem[\protect\citeauthoryear{Kossen, Farquhar, Gal, and Rainforth}{Kossen
  et~al\mbox{.}}{2021}]%
        {kossen2021active}
\bibfield{author}{\bibinfo{person}{Jannik Kossen}, \bibinfo{person}{Sebastian
  Farquhar}, \bibinfo{person}{Yarin Gal}, {and} \bibinfo{person}{Tom
  Rainforth}.} \bibinfo{year}{2021}\natexlab{}.
\newblock \showarticletitle{Active testing: Sample-efficient model evaluation}.
  In \bibinfo{booktitle}{\emph{International Conference on Machine Learning}}.
  PMLR, \bibinfo{pages}{5753--5763}.
\newblock


\bibitem[\protect\citeauthoryear{Kukutai and Taylor}{Kukutai and
  Taylor}{2016}]%
        {kukutai2016indigenous}
\bibfield{author}{\bibinfo{person}{Tahu Kukutai} {and} \bibinfo{person}{John
  Taylor}.} \bibinfo{year}{2016}\natexlab{}.
\newblock \bibinfo{booktitle}{\emph{Indigenous data sovereignty: Toward an
  agenda}}.
\newblock \bibinfo{publisher}{ANU press}.
\newblock


\bibitem[\protect\citeauthoryear{Kuwajima, Yasuoka, and Nakae}{Kuwajima
  et~al\mbox{.}}{2020}]%
        {kuwajima2020engineering}
\bibfield{author}{\bibinfo{person}{Hiroshi Kuwajima},
  \bibinfo{person}{Hirotoshi Yasuoka}, {and} \bibinfo{person}{Toshihiro
  Nakae}.} \bibinfo{year}{2020}\natexlab{}.
\newblock \showarticletitle{Engineering problems in machine learning systems}.
\newblock \bibinfo{journal}{\emph{Machine Learning}} \bibinfo{volume}{109},
  \bibinfo{number}{5} (\bibinfo{year}{2020}), \bibinfo{pages}{1103--1126}.
\newblock


\bibitem[\protect\citeauthoryear{Lacoste, Boquet, Rostamzadeh, Oreshkin, Chung,
  and Krueger}{Lacoste et~al\mbox{.}}{2017}]%
        {lacoste2017deep}
\bibfield{author}{\bibinfo{person}{Alexandre Lacoste}, \bibinfo{person}{Thomas
  Boquet}, \bibinfo{person}{Negar Rostamzadeh}, \bibinfo{person}{Boris
  Oreshkin}, \bibinfo{person}{Wonchang Chung}, {and} \bibinfo{person}{David
  Krueger}.} \bibinfo{year}{2017}\natexlab{}.
\newblock \showarticletitle{Deep prior}.
\newblock \bibinfo{journal}{\emph{arXiv preprint arXiv:1712.05016}}
  (\bibinfo{year}{2017}).
\newblock


\bibitem[\protect\citeauthoryear{Lacoste, Oreshkin, Chung, Boquet, Rostamzadeh,
  and Krueger}{Lacoste et~al\mbox{.}}{2018}]%
        {lacoste2018uncertainty}
\bibfield{author}{\bibinfo{person}{Alexandre Lacoste}, \bibinfo{person}{Boris
  Oreshkin}, \bibinfo{person}{Wonchang Chung}, \bibinfo{person}{Thomas Boquet},
  \bibinfo{person}{Negar Rostamzadeh}, {and} \bibinfo{person}{David Krueger}.}
  \bibinfo{year}{2018}\natexlab{}.
\newblock \showarticletitle{Uncertainty in multitask transfer learning}.
\newblock \bibinfo{journal}{\emph{arXiv preprint arXiv:1806.07528}}
  (\bibinfo{year}{2018}).
\newblock


\bibitem[\protect\citeauthoryear{Lakoff and Johnson}{Lakoff and
  Johnson}{2008}]%
        {lakoff2008metaphors}
\bibfield{author}{\bibinfo{person}{George Lakoff} {and} \bibinfo{person}{Mark
  Johnson}.} \bibinfo{year}{2008}\natexlab{}.
\newblock \bibinfo{booktitle}{\emph{Metaphors we live by}}.
\newblock \bibinfo{publisher}{University of Chicago press}.
\newblock


\bibitem[\protect\citeauthoryear{Lecu{\'e} and Lerasle}{Lecu{\'e} and
  Lerasle}{2020}]%
        {lecue2020robust}
\bibfield{author}{\bibinfo{person}{Guillaume Lecu{\'e}} {and}
  \bibinfo{person}{Matthieu Lerasle}.} \bibinfo{year}{2020}\natexlab{}.
\newblock \showarticletitle{Robust machine learning by median-of-means: theory
  and practice}.
\newblock \bibinfo{journal}{\emph{The Annals of Statistics}}
  \bibinfo{volume}{48}, \bibinfo{number}{2} (\bibinfo{year}{2020}),
  \bibinfo{pages}{906--931}.
\newblock


\bibitem[\protect\citeauthoryear{Li, Song, Cao, Tetreault, Goldberg, Jaimes,
  and Luo}{Li et~al\mbox{.}}{2016}]%
        {li2016tgif}
\bibfield{author}{\bibinfo{person}{Yuncheng Li}, \bibinfo{person}{Yale Song},
  \bibinfo{person}{Liangliang Cao}, \bibinfo{person}{Joel Tetreault},
  \bibinfo{person}{Larry Goldberg}, \bibinfo{person}{Alejandro Jaimes}, {and}
  \bibinfo{person}{Jiebo Luo}.} \bibinfo{year}{2016}\natexlab{}.
\newblock \showarticletitle{TGIF: A new dataset and benchmark on animated GIF
  description}. In \bibinfo{booktitle}{\emph{Proceedings of the IEEE Conference
  on Computer Vision and Pattern Recognition}}. \bibinfo{pages}{4641--4650}.
\newblock


\bibitem[\protect\citeauthoryear{Liao, Taori, Raji, and Schmidt}{Liao
  et~al\mbox{.}}{2021}]%
        {liao2021we}
\bibfield{author}{\bibinfo{person}{Thomas Liao}, \bibinfo{person}{Rohan Taori},
  \bibinfo{person}{Inioluwa~Deborah Raji}, {and} \bibinfo{person}{Ludwig
  Schmidt}.} \bibinfo{year}{2021}\natexlab{}.
\newblock \showarticletitle{Are We Learning Yet? A Meta Review of Evaluation
  Failures Across Machine Learning}. In \bibinfo{booktitle}{\emph{Thirty-fifth
  Conference on Neural Information Processing Systems Datasets and Benchmarks
  Track (Round 2)}}.
\newblock


\bibitem[\protect\citeauthoryear{Lin, Shih, and Sher}{Lin
  et~al\mbox{.}}{2007}]%
        {lin2007integrating}
\bibfield{author}{\bibinfo{person}{Chien-Hsin Lin}, \bibinfo{person}{Hsin-Yu
  Shih}, {and} \bibinfo{person}{Peter~J Sher}.}
  \bibinfo{year}{2007}\natexlab{}.
\newblock \showarticletitle{Integrating technology readiness into technology
  acceptance: The TRAM model}.
\newblock \bibinfo{journal}{\emph{Psychology \& Marketing}}
  \bibinfo{volume}{24}, \bibinfo{number}{7} (\bibinfo{year}{2007}),
  \bibinfo{pages}{641--657}.
\newblock


\bibitem[\protect\citeauthoryear{Lin}{Lin}{2004}]%
        {lin2004rouge}
\bibfield{author}{\bibinfo{person}{Chin-Yew Lin}.}
  \bibinfo{year}{2004}\natexlab{}.
\newblock \showarticletitle{Rouge: A package for automatic evaluation of
  summaries}. In \bibinfo{booktitle}{\emph{Text summarization branches out}}.
  \bibinfo{pages}{74--81}.
\newblock


\bibitem[\protect\citeauthoryear{Liu, Dean, Rolf, Simchowitz, and Hardt}{Liu
  et~al\mbox{.}}{2018}]%
        {liu2018delayed}
\bibfield{author}{\bibinfo{person}{Lydia~T Liu}, \bibinfo{person}{Sarah Dean},
  \bibinfo{person}{Esther Rolf}, \bibinfo{person}{Max Simchowitz}, {and}
  \bibinfo{person}{Moritz Hardt}.} \bibinfo{year}{2018}\natexlab{}.
\newblock \showarticletitle{Delayed impact of fair machine learning}. In
  \bibinfo{booktitle}{\emph{International Conference on Machine Learning}}.
  PMLR, \bibinfo{pages}{3150--3158}.
\newblock


\bibitem[\protect\citeauthoryear{Lo and Wu}{Lo and Wu}{2010}]%
        {lo2010evaluating}
\bibfield{author}{\bibinfo{person}{Chi-kiu Lo} {and} \bibinfo{person}{Dekai
  Wu}.} \bibinfo{year}{2010}\natexlab{}.
\newblock \showarticletitle{Evaluating Machine Translation Utility via Semantic
  Role Labels.}. In \bibinfo{booktitle}{\emph{LREC}}. Citeseer.
\newblock


\bibitem[\protect\citeauthoryear{Madaio, Egede, Subramonyam, Wortman~Vaughan,
  and Wallach}{Madaio et~al\mbox{.}}{2022}]%
        {madaio2022assessing}
\bibfield{author}{\bibinfo{person}{Michael Madaio}, \bibinfo{person}{Lisa
  Egede}, \bibinfo{person}{Hariharan Subramonyam}, \bibinfo{person}{Jennifer
  Wortman~Vaughan}, {and} \bibinfo{person}{Hanna Wallach}.}
  \bibinfo{year}{2022}\natexlab{}.
\newblock \showarticletitle{Assessing the Fairness of AI Systems: AI
  Practitioners' Processes, Challenges, and Needs for Support}.
\newblock \bibinfo{journal}{\emph{Proceedings of the ACM on Human-Computer
  Interaction}} \bibinfo{volume}{6}, \bibinfo{number}{CSCW1}
  (\bibinfo{year}{2022}), \bibinfo{pages}{1--26}.
\newblock


\bibitem[\protect\citeauthoryear{Mantelero}{Mantelero}{2018}]%
        {mantelero2018ai}
\bibfield{author}{\bibinfo{person}{Alessandro Mantelero}.}
  \bibinfo{year}{2018}\natexlab{}.
\newblock \showarticletitle{AI and Big Data: A blueprint for a human rights,
  social and ethical impact assessment}.
\newblock \bibinfo{journal}{\emph{Computer Law \& Security Review}}
  \bibinfo{volume}{34}, \bibinfo{number}{4} (\bibinfo{year}{2018}),
  \bibinfo{pages}{754--772}.
\newblock


\bibitem[\protect\citeauthoryear{{Marrkula Center}}{{Marrkula Center}}{2019}]%
        {marrkula2019}
\bibfield{author}{\bibinfo{person}{{Marrkula Center}}.}
  \bibinfo{year}{2019}\natexlab{}.
\newblock \bibinfo{title}{Approaches to Ethical Decision-making}.
\newblock
\newblock
\urldef\tempurl%
\url{https://www.scu.edu/ethics/ethics-resources/ethical-decision-making/}
\showURL{%
\tempurl}


\bibitem[\protect\citeauthoryear{Martin, Prabhakaran, Kuhlberg, Smart, and
  Isaac}{Martin et~al\mbox{.}}{2020}]%
        {martin2020extending}
\bibfield{author}{\bibinfo{person}{Donald Martin, Jr.},
  \bibinfo{person}{Vinodkumar Prabhakaran}, \bibinfo{person}{Jill Kuhlberg},
  \bibinfo{person}{Andrew Smart}, {and} \bibinfo{person}{William~S. Isaac}.}
  \bibinfo{year}{2020}\natexlab{}.
\newblock \bibinfo{title}{Extending the Machine Learning Abstraction Boundary:
  A Complex Systems Approach to Incorporate Societal Context}.
\newblock
\newblock
\showeprint[arxiv]{2006.09663}~[cs.CY]


\bibitem[\protect\citeauthoryear{Matturro}{Matturro}{2013}]%
        {matturro2013soft}
\bibfield{author}{\bibinfo{person}{Gerardo Matturro}.}
  \bibinfo{year}{2013}\natexlab{}.
\newblock \showarticletitle{Soft skills in software engineering: A study of its
  demand by software companies in Uruguay}. In \bibinfo{booktitle}{\emph{2013
  6th international workshop on cooperative and human aspects of software
  engineering (CHASE)}}. IEEE, \bibinfo{pages}{133--136}.
\newblock


\bibitem[\protect\citeauthoryear{Mazzocchi}{Mazzocchi}{2015}]%
        {mazzocchi2015could}
\bibfield{author}{\bibinfo{person}{Fulvio Mazzocchi}.}
  \bibinfo{year}{2015}\natexlab{}.
\newblock \showarticletitle{Could Big Data be the end of theory in science? A
  few remarks on the epistemology of data-driven science}.
\newblock \bibinfo{journal}{\emph{EMBO reports}} \bibinfo{volume}{16},
  \bibinfo{number}{10} (\bibinfo{year}{2015}), \bibinfo{pages}{1250--1255}.
\newblock


\bibitem[\protect\citeauthoryear{McGregor, Murray, and Ng}{McGregor
  et~al\mbox{.}}{2019}]%
        {mcgregor2019international}
\bibfield{author}{\bibinfo{person}{Lorna McGregor}, \bibinfo{person}{Daragh
  Murray}, {and} \bibinfo{person}{Vivian Ng}.} \bibinfo{year}{2019}\natexlab{}.
\newblock \showarticletitle{International human rights law as a framework for
  algorithmic accountability}.
\newblock \bibinfo{journal}{\emph{International \& Comparative Law Quarterly}}
  \bibinfo{volume}{68}, \bibinfo{number}{2} (\bibinfo{year}{2019}),
  \bibinfo{pages}{309--343}.
\newblock


\bibitem[\protect\citeauthoryear{McNair}{McNair}{2018}]%
        {mcnair2018preventing}
\bibfield{author}{\bibinfo{person}{Douglas~S McNair}.}
  \bibinfo{year}{2018}\natexlab{}.
\newblock \showarticletitle{Preventing disparities: Bayesian and frequentist
  methods for assessing fairness in machine learning decision-support models}.
\newblock \bibinfo{journal}{\emph{New Insights into Bayesian Inference}}
  (\bibinfo{year}{2018}), \bibinfo{pages}{71}.
\newblock


\bibitem[\protect\citeauthoryear{Miceli, Yang, Naudts, Schuessler, Serbanescu,
  and Hanna}{Miceli et~al\mbox{.}}{2021}]%
        {miceli2021documenting}
\bibfield{author}{\bibinfo{person}{Milagros Miceli}, \bibinfo{person}{Tianling
  Yang}, \bibinfo{person}{Laurens Naudts}, \bibinfo{person}{Martin Schuessler},
  \bibinfo{person}{Diana Serbanescu}, {and} \bibinfo{person}{Alex Hanna}.}
  \bibinfo{year}{2021}\natexlab{}.
\newblock \showarticletitle{Documenting Computer Vision Datasets: An Invitation
  to Reflexive Data Practices}. In \bibinfo{booktitle}{\emph{Proceedings of the
  2021 ACM Conference on Fairness, Accountability, and Transparency}}.
  \bibinfo{pages}{161--172}.
\newblock


\bibitem[\protect\citeauthoryear{Mitchell, Wu, Zaldivar, Barnes, Vasserman,
  Hutchinson, Spitzer, Raji, and Gebru}{Mitchell et~al\mbox{.}}{2019}]%
        {mitchell2019model}
\bibfield{author}{\bibinfo{person}{Margaret Mitchell}, \bibinfo{person}{Simone
  Wu}, \bibinfo{person}{Andrew Zaldivar}, \bibinfo{person}{Parker Barnes},
  \bibinfo{person}{Lucy Vasserman}, \bibinfo{person}{Ben Hutchinson},
  \bibinfo{person}{Elena Spitzer}, \bibinfo{person}{Inioluwa~Deborah Raji},
  {and} \bibinfo{person}{Timnit Gebru}.} \bibinfo{year}{2019}\natexlab{}.
\newblock \showarticletitle{Model cards for model reporting}. In
  \bibinfo{booktitle}{\emph{Proceedings of the conference on fairness,
  accountability, and transparency}}. \bibinfo{pages}{220--229}.
\newblock


\bibitem[\protect\citeauthoryear{Moradi and Samwald}{Moradi and
  Samwald}{2021}]%
        {moradi2021evaluating}
\bibfield{author}{\bibinfo{person}{Milad Moradi} {and}
  \bibinfo{person}{Matthias Samwald}.} \bibinfo{year}{2021}\natexlab{}.
\newblock \showarticletitle{Evaluating the Robustness of Neural Language Models
  to Input Perturbations}. In \bibinfo{booktitle}{\emph{Proceedings of the 2021
  Conference on Empirical Methods in Natural Language Processing}}.
  \bibinfo{pages}{1558--1570}.
\newblock


\bibitem[\protect\citeauthoryear{Murez, Kolouri, Kriegman, Ramamoorthi, and
  Kim}{Murez et~al\mbox{.}}{2018}]%
        {murez2018image}
\bibfield{author}{\bibinfo{person}{Zak Murez}, \bibinfo{person}{Soheil
  Kolouri}, \bibinfo{person}{David Kriegman}, \bibinfo{person}{Ravi
  Ramamoorthi}, {and} \bibinfo{person}{Kyungnam Kim}.}
  \bibinfo{year}{2018}\natexlab{}.
\newblock \showarticletitle{Image to image translation for domain adaptation}.
  In \bibinfo{booktitle}{\emph{Proceedings of the IEEE Conference on Computer
  Vision and Pattern Recognition}}. \bibinfo{pages}{4500--4509}.
\newblock


\bibitem[\protect\citeauthoryear{Myers, Sandler, and Badgett}{Myers
  et~al\mbox{.}}{2011}]%
        {myers2011art}
\bibfield{author}{\bibinfo{person}{Glenford~J Myers}, \bibinfo{person}{Corey
  Sandler}, {and} \bibinfo{person}{Tom Badgett}.}
  \bibinfo{year}{2011}\natexlab{}.
\newblock \bibinfo{booktitle}{\emph{The art of software testing}}.
\newblock \bibinfo{publisher}{John Wiley \& Sons}.
\newblock


\bibitem[\protect\citeauthoryear{Neumann, Roessler, Suendermann-Oeft, and
  Ramanarayanan}{Neumann et~al\mbox{.}}{2020}]%
        {neumann2020utility}
\bibfield{author}{\bibinfo{person}{Michael Neumann}, \bibinfo{person}{Oliver
  Roessler}, \bibinfo{person}{David Suendermann-Oeft}, {and}
  \bibinfo{person}{Vikram Ramanarayanan}.} \bibinfo{year}{2020}\natexlab{}.
\newblock \showarticletitle{On the utility of audiovisual dialog technologies
  and signal analytics for real-time remote monitoring of depression
  biomarkers}. In \bibinfo{booktitle}{\emph{Proceedings of the First Workshop
  on Natural Language Processing for Medical Conversations}}.
  \bibinfo{pages}{47--52}.
\newblock


\bibitem[\protect\citeauthoryear{Norvig}{Norvig}{2017}]%
        {norvig2017chomsky}
\bibfield{author}{\bibinfo{person}{Peter Norvig}.}
  \bibinfo{year}{2017}\natexlab{}.
\newblock \showarticletitle{On Chomsky and the two cultures of statistical
  learning}.
\newblock In \bibinfo{booktitle}{\emph{Berechenbarkeit der Welt?}}
  \bibinfo{publisher}{Springer}, \bibinfo{pages}{61--83}.
\newblock


\bibitem[\protect\citeauthoryear{Olteanu, Castillo, Diaz, and
  K{\i}c{\i}man}{Olteanu et~al\mbox{.}}{2019}]%
        {olteanu2019social}
\bibfield{author}{\bibinfo{person}{Alexandra Olteanu}, \bibinfo{person}{Carlos
  Castillo}, \bibinfo{person}{Fernando Diaz}, {and} \bibinfo{person}{Emre
  K{\i}c{\i}man}.} \bibinfo{year}{2019}\natexlab{}.
\newblock \showarticletitle{Social data: Biases, methodological pitfalls, and
  ethical boundaries}.
\newblock \bibinfo{journal}{\emph{Frontiers in Big Data}}  \bibinfo{volume}{2}
  (\bibinfo{year}{2019}), \bibinfo{pages}{13}.
\newblock


\bibitem[\protect\citeauthoryear{Orekondy, Fritz, and Schiele}{Orekondy
  et~al\mbox{.}}{2018}]%
        {orekondy2018connecting}
\bibfield{author}{\bibinfo{person}{Tribhuvanesh Orekondy},
  \bibinfo{person}{Mario Fritz}, {and} \bibinfo{person}{Bernt Schiele}.}
  \bibinfo{year}{2018}\natexlab{}.
\newblock \showarticletitle{Connecting pixels to privacy and utility: Automatic
  redaction of private information in images}. In
  \bibinfo{booktitle}{\emph{Proceedings of the IEEE Conference on Computer
  Vision and Pattern Recognition}}. \bibinfo{pages}{8466--8475}.
\newblock


\bibitem[\protect\citeauthoryear{Papineni, Roukos, Ward, and Zhu}{Papineni
  et~al\mbox{.}}{2002}]%
        {papineni2002bleu}
\bibfield{author}{\bibinfo{person}{Kishore Papineni}, \bibinfo{person}{Salim
  Roukos}, \bibinfo{person}{Todd Ward}, {and} \bibinfo{person}{Wei-Jing Zhu}.}
  \bibinfo{year}{2002}\natexlab{}.
\newblock \showarticletitle{\textsc{Bleu}: a method for automatic evaluation of
  machine translation}. In \bibinfo{booktitle}{\emph{Proceedings of the 40th
  annual meeting of the Association for Computational Linguistics}}.
  \bibinfo{pages}{311--318}.
\newblock


\bibitem[\protect\citeauthoryear{Pinheiro, Rostamzadeh, and Ahn}{Pinheiro
  et~al\mbox{.}}{2019}]%
        {pinheiro2019domain}
\bibfield{author}{\bibinfo{person}{Pedro~O Pinheiro}, \bibinfo{person}{Negar
  Rostamzadeh}, {and} \bibinfo{person}{Sungjin Ahn}.}
  \bibinfo{year}{2019}\natexlab{}.
\newblock \showarticletitle{Domain-adaptive single-view 3d reconstruction}. In
  \bibinfo{booktitle}{\emph{Proceedings of the IEEE/CVF International
  Conference on Computer Vision}}. \bibinfo{pages}{7638--7647}.
\newblock


\bibitem[\protect\citeauthoryear{Powers}{Powers}{2011}]%
        {powers2011evaluation}
\bibfield{author}{\bibinfo{person}{David Martin~Ward Powers}.}
  \bibinfo{year}{2011}\natexlab{}.
\newblock \showarticletitle{Evaluation: From Precision, Recall and {F}-Factor
  to {ROC}, Informedness, Markedness \& Correlation}.
\newblock \bibinfo{journal}{\emph{Journal of Machine Learning Technologies}}
  \bibinfo{volume}{2}, \bibinfo{number}{1} (\bibinfo{year}{2011}),
  \bibinfo{pages}{37--63}.
\newblock


\bibitem[\protect\citeauthoryear{Powers}{Powers}{2012a}]%
        {powers2012problemauc}
\bibfield{author}{\bibinfo{person}{David Martin~Ward Powers}.}
  \bibinfo{year}{2012}\natexlab{a}.
\newblock \showarticletitle{The problem of area under the curve}. In
  \bibinfo{booktitle}{\emph{2012 IEEE International conference on information
  science and technology}}. IEEE, \bibinfo{pages}{567--573}.
\newblock


\bibitem[\protect\citeauthoryear{Powers}{Powers}{2012b}]%
        {powers2012problem}
\bibfield{author}{\bibinfo{person}{David Martin~Ward Powers}.}
  \bibinfo{year}{2012}\natexlab{b}.
\newblock \showarticletitle{The problem with kappa}. In
  \bibinfo{booktitle}{\emph{Proceedings of the 13th Conference of the European
  Chapter of the Association for Computational Linguistics}}.
  \bibinfo{pages}{345--355}.
\newblock


\bibitem[\protect\citeauthoryear{Powers}{Powers}{2014}]%
        {powers2014f}
\bibfield{author}{\bibinfo{person}{David Martin~Ward Powers}.}
  \bibinfo{year}{2014}\natexlab{}.
\newblock \showarticletitle{What the F-measure doesn’t measure: Features,
  Flaws, Fallacies and Fixes}.
\newblock \bibinfo{journal}{\emph{Technical report, Beijing University of
  Technology, China \& Flinders University, Australia, Tech. Rep.}}
  (\bibinfo{year}{2014}).
\newblock


\bibitem[\protect\citeauthoryear{Prabhakaran, Davani, and Diaz}{Prabhakaran
  et~al\mbox{.}}{2021}]%
        {prabhakaran2021releasing}
\bibfield{author}{\bibinfo{person}{Vinodkumar Prabhakaran},
  \bibinfo{person}{Aida~Mostafazadeh Davani}, {and} \bibinfo{person}{Mark
  Diaz}.} \bibinfo{year}{2021}\natexlab{}.
\newblock \showarticletitle{On Releasing Annotator-Level Labels and Information
  in Datasets}. In \bibinfo{booktitle}{\emph{Proceedings of The Joint 15th
  Linguistic Annotation Workshop (LAW) and 3rd Designing Meaning
  Representations (DMR) Workshop}}. \bibinfo{pages}{133--138}.
\newblock


\bibitem[\protect\citeauthoryear{Prabhakaran, Hutchinson, and
  Mitchell}{Prabhakaran et~al\mbox{.}}{2019}]%
        {prabhakaran2019perturbation}
\bibfield{author}{\bibinfo{person}{Vinodkumar Prabhakaran},
  \bibinfo{person}{Ben Hutchinson}, {and} \bibinfo{person}{Margaret Mitchell}.}
  \bibinfo{year}{2019}\natexlab{}.
\newblock \showarticletitle{Perturbation Sensitivity Analysis to Detect
  Unintended Model Biases}. In \bibinfo{booktitle}{\emph{Proceedings of the
  2019 Conference on Empirical Methods in Natural Language Processing and the
  9th International Joint Conference on Natural Language Processing
  (EMNLP-IJCNLP)}}. \bibinfo{pages}{5740--5745}.
\newblock


\bibitem[\protect\citeauthoryear{Provost and Fawcett}{Provost and
  Fawcett}{1997}]%
        {provost1997analysis}
\bibfield{author}{\bibinfo{person}{Foster Provost} {and} \bibinfo{person}{Tom
  Fawcett}.} \bibinfo{year}{1997}\natexlab{}.
\newblock \showarticletitle{Analysis and visualization of classifier
  performance with nonuniform class and cost distributions}. In
  \bibinfo{booktitle}{\emph{Proceedings of AAAI-97 Workshop on {AI} Approaches
  to Fraud Detection \& Risk Management}}. \bibinfo{pages}{57--63}.
\newblock


\bibitem[\protect\citeauthoryear{Pustejovsky}{Pustejovsky}{1998}]%
        {pustejovsky1998generative}
\bibfield{author}{\bibinfo{person}{James Pustejovsky}.}
  \bibinfo{year}{1998}\natexlab{}.
\newblock \bibinfo{booktitle}{\emph{The generative lexicon}}.
\newblock \bibinfo{publisher}{MIT press}.
\newblock


\bibitem[\protect\citeauthoryear{Raji, Denton, Bender, Hanna, and
  Paullada}{Raji et~al\mbox{.}}{2021}]%
        {raji2021ai}
\bibfield{author}{\bibinfo{person}{Inioluwa~Deborah Raji},
  \bibinfo{person}{Emily Denton}, \bibinfo{person}{Emily~M Bender},
  \bibinfo{person}{Alex Hanna}, {and} \bibinfo{person}{Amandalynne Paullada}.}
  \bibinfo{year}{2021}\natexlab{}.
\newblock \showarticletitle{{AI} and the Everything in the Whole Wide World
  Benchmark}. In \bibinfo{booktitle}{\emph{Thirty-fifth Conference on Neural
  Information Processing Systems Datasets and Benchmarks Track (Round 2)}}.
\newblock


\bibitem[\protect\citeauthoryear{Raji, Smart, White, Mitchell, Gebru,
  Hutchinson, Smith-Loud, Theron, and Barnes}{Raji et~al\mbox{.}}{2020}]%
        {raji2020closing}
\bibfield{author}{\bibinfo{person}{Inioluwa~Deborah Raji},
  \bibinfo{person}{Andrew Smart}, \bibinfo{person}{Rebecca~N White},
  \bibinfo{person}{Margaret Mitchell}, \bibinfo{person}{Timnit Gebru},
  \bibinfo{person}{Ben Hutchinson}, \bibinfo{person}{Jamila Smith-Loud},
  \bibinfo{person}{Daniel Theron}, {and} \bibinfo{person}{Parker Barnes}.}
  \bibinfo{year}{2020}\natexlab{}.
\newblock \showarticletitle{Closing the {AI} accountability gap: Defining an
  end-to-end framework for internal algorithmic auditing}. In
  \bibinfo{booktitle}{\emph{Proceedings of the 2020 conference on fairness,
  accountability, and transparency}}. \bibinfo{pages}{33--44}.
\newblock


\bibitem[\protect\citeauthoryear{Raso, Hilligoss, Krishnamurthy, Bavitz, and
  Kim}{Raso et~al\mbox{.}}{2018}]%
        {raso2018artificial}
\bibfield{author}{\bibinfo{person}{Filippo~A Raso}, \bibinfo{person}{Hannah
  Hilligoss}, \bibinfo{person}{Vivek Krishnamurthy},
  \bibinfo{person}{Christopher Bavitz}, {and} \bibinfo{person}{Levin Kim}.}
  \bibinfo{year}{2018}\natexlab{}.
\newblock \showarticletitle{Artificial intelligence \& human rights:
  Opportunities \& risks}.
\newblock \bibinfo{journal}{\emph{Berkman Klein Center Research Publication}}
  \bibinfo{number}{2018-6} (\bibinfo{year}{2018}).
\newblock


\bibitem[\protect\citeauthoryear{Ren, He, Girshick, and Sun}{Ren
  et~al\mbox{.}}{2015}]%
        {ren2015faster}
\bibfield{author}{\bibinfo{person}{Shaoqing Ren}, \bibinfo{person}{Kaiming He},
  \bibinfo{person}{Ross Girshick}, {and} \bibinfo{person}{Jian Sun}.}
  \bibinfo{year}{2015}\natexlab{}.
\newblock \showarticletitle{Faster r-cnn: Towards real-time object detection
  with region proposal networks}.
\newblock \bibinfo{journal}{\emph{Advances in neural information processing
  systems}}  \bibinfo{volume}{28} (\bibinfo{year}{2015}).
\newblock


\bibitem[\protect\citeauthoryear{Ribeiro, Wu, Guestrin, and Singh}{Ribeiro
  et~al\mbox{.}}{2020}]%
        {ribeiro2020beyond}
\bibfield{author}{\bibinfo{person}{Marco~Tulio Ribeiro},
  \bibinfo{person}{Tongshuang Wu}, \bibinfo{person}{Carlos Guestrin}, {and}
  \bibinfo{person}{Sameer Singh}.} \bibinfo{year}{2020}\natexlab{}.
\newblock \showarticletitle{Beyond accuracy: Behavioral testing of NLP models
  with CheckList}.
\newblock \bibinfo{journal}{\emph{arXiv preprint arXiv:2005.04118}}
  (\bibinfo{year}{2020}).
\newblock


\bibitem[\protect\citeauthoryear{Rismani and Moon}{Rismani and Moon}{2021}]%
        {rismani2021how}
\bibfield{author}{\bibinfo{person}{Shalaleh Rismani} {and}
  \bibinfo{person}{Ajung Moon}.} \bibinfo{year}{2021}\natexlab{}.
\newblock \showarticletitle{How do AI systems fail socially?: an engineering
  risk analysis approach}. In \bibinfo{booktitle}{\emph{2021 IEEE International
  Symposium on Ethics in Engineering, Science and Technology (ETHICS)}}.
  \bibinfo{pages}{1--8}.
\newblock
\urldef\tempurl%
\url{https://doi.org/10.1109/ETHICS53270.2021.9632769}
\showDOI{\tempurl}


\bibitem[\protect\citeauthoryear{Rodriguez, Barrow, Hoyle, Lalor, Jia, and
  Boyd-Graber}{Rodriguez et~al\mbox{.}}{2021}]%
        {rodriguez-etal-2021-evaluation}
\bibfield{author}{\bibinfo{person}{Pedro Rodriguez}, \bibinfo{person}{Joe
  Barrow}, \bibinfo{person}{Alexander~Miserlis Hoyle}, \bibinfo{person}{John~P.
  Lalor}, \bibinfo{person}{Robin Jia}, {and} \bibinfo{person}{Jordan
  Boyd-Graber}.} \bibinfo{year}{2021}\natexlab{}.
\newblock \showarticletitle{Evaluation Examples are not Equally Informative:
  How should that change {NLP} Leaderboards?}. In
  \bibinfo{booktitle}{\emph{Proceedings of the 59th Annual Meeting of the
  Association for Computational Linguistics and the 11th International Joint
  Conference on Natural Language Processing (Volume 1: Long Papers)}}.
  \bibinfo{publisher}{Association for Computational Linguistics},
  \bibinfo{address}{Online}, \bibinfo{pages}{4486--4503}.
\newblock
\urldef\tempurl%
\url{https://doi.org/10.18653/v1/2021.acl-long.346}
\showDOI{\tempurl}


\bibitem[\protect\citeauthoryear{Rostamzadeh, Hosseini, Boquet, Stokowiec,
  Zhang, Jauvin, and Pal}{Rostamzadeh et~al\mbox{.}}{2018}]%
        {rostamzadeh2018fashion}
\bibfield{author}{\bibinfo{person}{Negar Rostamzadeh},
  \bibinfo{person}{Seyedarian Hosseini}, \bibinfo{person}{Thomas Boquet},
  \bibinfo{person}{Wojciech Stokowiec}, \bibinfo{person}{Ying Zhang},
  \bibinfo{person}{Christian Jauvin}, {and} \bibinfo{person}{Chris Pal}.}
  \bibinfo{year}{2018}\natexlab{}.
\newblock \showarticletitle{Fashion-gen: The generative fashion dataset and
  challenge}.
\newblock \bibinfo{journal}{\emph{arXiv preprint arXiv:1806.08317}}
  (\bibinfo{year}{2018}).
\newblock


\bibitem[\protect\citeauthoryear{Rostamzadeh, Hutchinson, Greer, and
  Prabhakaran}{Rostamzadeh et~al\mbox{.}}{2021}]%
        {rostamzadeh2021thinking}
\bibfield{author}{\bibinfo{person}{Negar Rostamzadeh}, \bibinfo{person}{Ben
  Hutchinson}, \bibinfo{person}{Christina Greer}, {and}
  \bibinfo{person}{Vinodkumar Prabhakaran}.} \bibinfo{year}{2021}\natexlab{}.
\newblock \showarticletitle{Thinking Beyond Distributions in Testing Machine
  Learned Models}. In \bibinfo{booktitle}{\emph{NeurIPS 2021 Workshop on
  Distribution Shifts: Connecting Methods and Applications}}.
\newblock


\bibitem[\protect\citeauthoryear{Ruiz, Kortylewski, Qiu, Xie, Bargal, Yuille,
  and Sclaroff}{Ruiz et~al\mbox{.}}{2022}]%
        {ruiz2022simulated}
\bibfield{author}{\bibinfo{person}{Nataniel Ruiz}, \bibinfo{person}{Adam
  Kortylewski}, \bibinfo{person}{Weichao Qiu}, \bibinfo{person}{Cihang Xie},
  \bibinfo{person}{Sarah~Adel Bargal}, \bibinfo{person}{Alan Yuille}, {and}
  \bibinfo{person}{Stan Sclaroff}.} \bibinfo{year}{2022}\natexlab{}.
\newblock \showarticletitle{Simulated Adversarial Testing of Face Recognition
  Models}.
\newblock \bibinfo{journal}{\emph{CVPR}} (\bibinfo{year}{2022}).
\newblock


\bibitem[\protect\citeauthoryear{Sambasivan, Arnesen, Hutchinson, Doshi, and
  Prabhakaran}{Sambasivan et~al\mbox{.}}{2021a}]%
        {sambasivan2021re}
\bibfield{author}{\bibinfo{person}{Nithya Sambasivan}, \bibinfo{person}{Erin
  Arnesen}, \bibinfo{person}{Ben Hutchinson}, \bibinfo{person}{Tulsee Doshi},
  {and} \bibinfo{person}{Vinodkumar Prabhakaran}.}
  \bibinfo{year}{2021}\natexlab{a}.
\newblock \showarticletitle{Re-Imagining Algorithmic Fairness in India and
  Beyond}. In \bibinfo{booktitle}{\emph{Proceedings of the 2021 ACM Conference
  on Fairness, Accountability, and Transparency}} (Virtual Event, Canada)
  \emph{(\bibinfo{series}{FAccT '21})}. \bibinfo{publisher}{Association for
  Computing Machinery}, \bibinfo{address}{New York, NY, USA},
  \bibinfo{pages}{315–328}.
\newblock
\showISBNx{9781450383097}
\urldef\tempurl%
\url{https://doi.org/10.1145/3442188.3445896}
\showDOI{\tempurl}


\bibitem[\protect\citeauthoryear{Sambasivan and Holbrook}{Sambasivan and
  Holbrook}{2018}]%
        {sambasivan2018toward}
\bibfield{author}{\bibinfo{person}{Nithya Sambasivan} {and}
  \bibinfo{person}{Jess Holbrook}.} \bibinfo{year}{2018}\natexlab{}.
\newblock \showarticletitle{Toward responsible AI for the next billion users}.
\newblock \bibinfo{journal}{\emph{Interactions}} \bibinfo{volume}{26},
  \bibinfo{number}{1} (\bibinfo{year}{2018}), \bibinfo{pages}{68--71}.
\newblock


\bibitem[\protect\citeauthoryear{Sambasivan, Kapania, Highfill, Akrong,
  Paritosh, and Aroyo}{Sambasivan et~al\mbox{.}}{2021b}]%
        {sambasivan2021everyone}
\bibfield{author}{\bibinfo{person}{Nithya Sambasivan}, \bibinfo{person}{Shivani
  Kapania}, \bibinfo{person}{Hannah Highfill}, \bibinfo{person}{Diana Akrong},
  \bibinfo{person}{Praveen Paritosh}, {and} \bibinfo{person}{Lora~M Aroyo}.}
  \bibinfo{year}{2021}\natexlab{b}.
\newblock \showarticletitle{“Everyone wants to do the model work, not the
  data work”: Data Cascades in High-Stakes AI}. In
  \bibinfo{booktitle}{\emph{proceedings of the 2021 CHI Conference on Human
  Factors in Computing Systems}}. \bibinfo{pages}{1--15}.
\newblock


\bibitem[\protect\citeauthoryear{S{\'a}nchez-Gord{\'o}n, Rijal, and
  Colomo-Palacios}{S{\'a}nchez-Gord{\'o}n et~al\mbox{.}}{2020}]%
        {sanchez2020beyond}
\bibfield{author}{\bibinfo{person}{Mary S{\'a}nchez-Gord{\'o}n},
  \bibinfo{person}{Laxmi Rijal}, {and} \bibinfo{person}{Ricardo
  Colomo-Palacios}.} \bibinfo{year}{2020}\natexlab{}.
\newblock \showarticletitle{Beyond Technical Skills in Software Testing:
  Automated versus Manual Testing}. In \bibinfo{booktitle}{\emph{Proceedings of
  the IEEE/ACM 42nd International Conference on Software Engineering
  Workshops}}. \bibinfo{pages}{161--164}.
\newblock


\bibitem[\protect\citeauthoryear{Sang and De~Meulder}{Sang and
  De~Meulder}{2003}]%
        {sang2003introduction}
\bibfield{author}{\bibinfo{person}{Erik Tjong~Kim Sang} {and}
  \bibinfo{person}{Fien De~Meulder}.} \bibinfo{year}{2003}\natexlab{}.
\newblock \showarticletitle{Introduction to the CoNLL-2003 Shared Task:
  Language-Independent Named Entity Recognition}. In
  \bibinfo{booktitle}{\emph{Proceedings of the Seventh Conference on Natural
  Language Learning at HLT-NAACL 2003}}. \bibinfo{pages}{142--147}.
\newblock


\bibitem[\protect\citeauthoryear{Scheuerman, Hanna, and Denton}{Scheuerman
  et~al\mbox{.}}{2021}]%
        {scheuerman2021datasets}
\bibfield{author}{\bibinfo{person}{Morgan~Klaus Scheuerman},
  \bibinfo{person}{Alex Hanna}, {and} \bibinfo{person}{Emily Denton}.}
  \bibinfo{year}{2021}\natexlab{}.
\newblock \showarticletitle{Do datasets have politics? Disciplinary values in
  computer vision dataset development}.
\newblock \bibinfo{journal}{\emph{Proceedings of the ACM on Human-Computer
  Interaction}} \bibinfo{volume}{5}, \bibinfo{number}{CSCW2}
  (\bibinfo{year}{2021}), \bibinfo{pages}{1--37}.
\newblock


\bibitem[\protect\citeauthoryear{Schlangen}{Schlangen}{2021}]%
        {schlangen2021targeting}
\bibfield{author}{\bibinfo{person}{David Schlangen}.}
  \bibinfo{year}{2021}\natexlab{}.
\newblock \showarticletitle{Targeting the Benchmark: On Methodology in Current
  Natural Language Processing Research}. In
  \bibinfo{booktitle}{\emph{Proceedings of the 59th Annual Meeting of the
  Association for Computational Linguistics and the 11th International Joint
  Conference on Natural Language Processing (Volume 2: Short Papers)}}.
  \bibinfo{pages}{670--674}.
\newblock


\bibitem[\protect\citeauthoryear{Schwartz, Dodge, Smith, and Etzioni}{Schwartz
  et~al\mbox{.}}{2020}]%
        {schwartz2020green}
\bibfield{author}{\bibinfo{person}{Roy Schwartz}, \bibinfo{person}{Jesse
  Dodge}, \bibinfo{person}{Noah~A Smith}, {and} \bibinfo{person}{Oren
  Etzioni}.} \bibinfo{year}{2020}\natexlab{}.
\newblock \showarticletitle{Green {AI}}.
\newblock \bibinfo{journal}{\emph{Commun. ACM}} \bibinfo{volume}{63},
  \bibinfo{number}{12} (\bibinfo{year}{2020}), \bibinfo{pages}{54--63}.
\newblock


\bibitem[\protect\citeauthoryear{Sculley, Snoek, Wiltschko, and Rahimi}{Sculley
  et~al\mbox{.}}{2018}]%
        {sculley2018winner}
\bibfield{author}{\bibinfo{person}{David Sculley}, \bibinfo{person}{Jasper
  Snoek}, \bibinfo{person}{Alex Wiltschko}, {and} \bibinfo{person}{Ali
  Rahimi}.} \bibinfo{year}{2018}\natexlab{}.
\newblock \showarticletitle{Winner's curse? On pace, progress, and empirical
  rigor}. In \bibinfo{booktitle}{\emph{Proceedings of ICLR 2018}}.
\newblock


\bibitem[\protect\citeauthoryear{Selbst, Boyd, Friedler, Venkatasubramanian,
  and Vertesi}{Selbst et~al\mbox{.}}{2019}]%
        {selbst2019fairness}
\bibfield{author}{\bibinfo{person}{Andrew~D Selbst}, \bibinfo{person}{Danah
  Boyd}, \bibinfo{person}{Sorelle~A Friedler}, \bibinfo{person}{Suresh
  Venkatasubramanian}, {and} \bibinfo{person}{Janet Vertesi}.}
  \bibinfo{year}{2019}\natexlab{}.
\newblock \showarticletitle{Fairness and abstraction in sociotechnical
  systems}. In \bibinfo{booktitle}{\emph{Proceedings of the conference on
  fairness, accountability, and transparency}}. \bibinfo{pages}{59--68}.
\newblock


\bibitem[\protect\citeauthoryear{Shue}{Shue}{2020}]%
        {shue2020basic}
\bibfield{author}{\bibinfo{person}{Henry Shue}.}
  \bibinfo{year}{2020}\natexlab{}.
\newblock \bibinfo{booktitle}{\emph{Basic rights: Subsistence, affluence, and
  US foreign policy}}.
\newblock \bibinfo{publisher}{Princeton University Press}.
\newblock


\bibitem[\protect\citeauthoryear{Silberman, Tomlinson, LaPlante, Ross, Irani,
  and Zaldivar}{Silberman et~al\mbox{.}}{2018}]%
        {silberman2018responsible}
\bibfield{author}{\bibinfo{person}{M~Six Silberman}, \bibinfo{person}{Bill
  Tomlinson}, \bibinfo{person}{Rochelle LaPlante}, \bibinfo{person}{Joel Ross},
  \bibinfo{person}{Lilly Irani}, {and} \bibinfo{person}{Andrew Zaldivar}.}
  \bibinfo{year}{2018}\natexlab{}.
\newblock \showarticletitle{Responsible research with crowds: pay crowdworkers
  at least minimum wage}.
\newblock \bibinfo{journal}{\emph{Commun. ACM}} \bibinfo{volume}{61},
  \bibinfo{number}{3} (\bibinfo{year}{2018}), \bibinfo{pages}{39--41}.
\newblock


\bibitem[\protect\citeauthoryear{Sinnott-Armstrong}{Sinnott-Armstrong}{2021}]%
        {consequentialism}
\bibfield{author}{\bibinfo{person}{Walter Sinnott-Armstrong}.}
  \bibinfo{year}{2021}\natexlab{}.
\newblock \showarticletitle{Consequentialism}.
\newblock \bibinfo{journal}{\emph{The Stanford Encyclopedia of Philosophy}}
  \bibinfo{volume}{Winter 2021 Edition} (\bibinfo{year}{2021}).
\newblock
\urldef\tempurl%
\url{https://plato.stanford.edu/archives/win2021/entries/consequentialism/}
\showURL{%
\tempurl}


\bibitem[\protect\citeauthoryear{Star and Griesemer}{Star and
  Griesemer}{1989}]%
        {star1989institutional}
\bibfield{author}{\bibinfo{person}{Susan~Leigh Star} {and}
  \bibinfo{person}{James~R Griesemer}.} \bibinfo{year}{1989}\natexlab{}.
\newblock \showarticletitle{Institutional ecology, `translations' and boundary
  objects: Amateurs and professionals in Berkeley's Museum of Vertebrate
  Zoology, 1907-39}.
\newblock \bibinfo{journal}{\emph{Social studies of science}}
  \bibinfo{volume}{19}, \bibinfo{number}{3} (\bibinfo{year}{1989}),
  \bibinfo{pages}{387--420}.
\newblock


\bibitem[\protect\citeauthoryear{Strubell, Ganesh, and McCallum}{Strubell
  et~al\mbox{.}}{2019}]%
        {strubell2019energy}
\bibfield{author}{\bibinfo{person}{Emma Strubell}, \bibinfo{person}{Ananya
  Ganesh}, {and} \bibinfo{person}{Andrew McCallum}.}
  \bibinfo{year}{2019}\natexlab{}.
\newblock \showarticletitle{Energy and Policy Considerations for Deep Learning
  in NLP}. In \bibinfo{booktitle}{\emph{Proceedings of the 57th Annual Meeting
  of the Association for Computational Linguistics}}.
  \bibinfo{pages}{3645--3650}.
\newblock


\bibitem[\protect\citeauthoryear{Sugiyama, Nakajima, Kashima, Buenau, and
  Kawanabe}{Sugiyama et~al\mbox{.}}{2007}]%
        {sugiyama2007direct}
\bibfield{author}{\bibinfo{person}{Masashi Sugiyama}, \bibinfo{person}{Shinichi
  Nakajima}, \bibinfo{person}{Hisashi Kashima}, \bibinfo{person}{Paul Buenau},
  {and} \bibinfo{person}{Motoaki Kawanabe}.} \bibinfo{year}{2007}\natexlab{}.
\newblock \showarticletitle{Direct Importance Estimation with Model Selection
  and Its Application to Covariate Shift Adaptation}.
\newblock \bibinfo{journal}{\emph{Advances in Neural Information Processing
  Systems}}  \bibinfo{volume}{20} (\bibinfo{year}{2007}).
\newblock


\bibitem[\protect\citeauthoryear{Thomas and Uminsky}{Thomas and
  Uminsky}{2020}]%
        {thomas2020reliance}
\bibfield{author}{\bibinfo{person}{RL Thomas} {and} \bibinfo{person}{D
  Uminsky}.} \bibinfo{year}{2020}\natexlab{}.
\newblock \showarticletitle{Reliance on metrics is a fundamental challenge for
  {AI}}. In \bibinfo{booktitle}{\emph{Proceedings of the Ethics of Data Science
  Conference}}.
\newblock


\bibitem[\protect\citeauthoryear{Tukey}{Tukey}{1962}]%
        {tukey1962future}
\bibfield{author}{\bibinfo{person}{John~W Tukey}.}
  \bibinfo{year}{1962}\natexlab{}.
\newblock \showarticletitle{The future of data analysis}.
\newblock \bibinfo{journal}{\emph{The annals of mathematical statistics}}
  \bibinfo{volume}{33}, \bibinfo{number}{1} (\bibinfo{year}{1962}),
  \bibinfo{pages}{1--67}.
\newblock


\bibitem[\protect\citeauthoryear{Turney}{Turney}{1994}]%
        {turney1994cost}
\bibfield{author}{\bibinfo{person}{Peter~D Turney}.}
  \bibinfo{year}{1994}\natexlab{}.
\newblock \showarticletitle{Cost-sensitive classification: Empirical evaluation
  of a hybrid genetic decision tree induction algorithm}.
\newblock \bibinfo{journal}{\emph{Journal of artificial intelligence research}}
   \bibinfo{volume}{2} (\bibinfo{year}{1994}), \bibinfo{pages}{369--409}.
\newblock


\bibitem[\protect\citeauthoryear{Ustalov, Panchenko, and Biemann}{Ustalov
  et~al\mbox{.}}{2017}]%
        {ustalov2017watset}
\bibfield{author}{\bibinfo{person}{Dmitry Ustalov}, \bibinfo{person}{Alexander
  Panchenko}, {and} \bibinfo{person}{Chris Biemann}.}
  \bibinfo{year}{2017}\natexlab{}.
\newblock \showarticletitle{Watset: Automatic induction of synsets from a graph
  of synonyms}. In \bibinfo{booktitle}{\emph{55th Annual Meeting of the
  Association for Computational Linguistics, ACL 2017}}. Association for
  Computational Linguistics, \bibinfo{pages}{1579--1590}.
\newblock


\bibitem[\protect\citeauthoryear{Vallor}{Vallor}{2016}]%
        {vallor2016technology}
\bibfield{author}{\bibinfo{person}{Shannon Vallor}.}
  \bibinfo{year}{2016}\natexlab{}.
\newblock \bibinfo{booktitle}{\emph{Technology and the virtues: A philosophical
  guide to a future worth wanting}}.
\newblock \bibinfo{publisher}{Oxford University Press}.
\newblock


\bibitem[\protect\citeauthoryear{Van~Rijsbergen}{Van~Rijsbergen}{1974}]%
        {van1974foundation}
\bibfield{author}{\bibinfo{person}{Cornelis~Joost Van~Rijsbergen}.}
  \bibinfo{year}{1974}\natexlab{}.
\newblock \showarticletitle{Foundation of evaluation}.
\newblock \bibinfo{journal}{\emph{Journal of documentation}}
  (\bibinfo{year}{1974}).
\newblock


\bibitem[\protect\citeauthoryear{Vogelsang and Borg}{Vogelsang and
  Borg}{2019}]%
        {vogelsang2019requirements}
\bibfield{author}{\bibinfo{person}{Andreas Vogelsang} {and}
  \bibinfo{person}{Markus Borg}.} \bibinfo{year}{2019}\natexlab{}.
\newblock \showarticletitle{Requirements engineering for machine learning:
  Perspectives from data scientists}. In \bibinfo{booktitle}{\emph{2019 IEEE
  27th International Requirements Engineering Conference Workshops (REW)}}.
  IEEE, \bibinfo{pages}{245--251}.
\newblock


\bibitem[\protect\citeauthoryear{Wallach}{Wallach}{2018}]%
        {wallach2018computational}
\bibfield{author}{\bibinfo{person}{Hanna Wallach}.}
  \bibinfo{year}{2018}\natexlab{}.
\newblock \showarticletitle{Computational social science$\ne$ computer science+
  social data}.
\newblock \bibinfo{journal}{\emph{Commun. ACM}} \bibinfo{volume}{61},
  \bibinfo{number}{3} (\bibinfo{year}{2018}), \bibinfo{pages}{42--44}.
\newblock


\bibitem[\protect\citeauthoryear{Wang, Singh, Michael, Hill, Levy, and
  Bowman}{Wang et~al\mbox{.}}{2018}]%
        {wang2018glue}
\bibfield{author}{\bibinfo{person}{Alex Wang}, \bibinfo{person}{Amanpreet
  Singh}, \bibinfo{person}{Julian Michael}, \bibinfo{person}{Felix Hill},
  \bibinfo{person}{Omer Levy}, {and} \bibinfo{person}{Samuel Bowman}.}
  \bibinfo{year}{2018}\natexlab{}.
\newblock \showarticletitle{GLUE: A Multi-Task Benchmark and Analysis Platform
  for Natural Language Understanding}. In \bibinfo{booktitle}{\emph{Proceedings
  of the 2018 EMNLP Workshop BlackboxNLP: Analyzing and Interpreting Neural
  Networks for NLP}}. \bibinfo{pages}{353--355}.
\newblock


\bibitem[\protect\citeauthoryear{Wang, Lan, Liu, Ouyang, Zeng, and Qin}{Wang
  et~al\mbox{.}}{2021}]%
        {wang2021generalizing}
\bibfield{author}{\bibinfo{person}{Jindong Wang}, \bibinfo{person}{Cuiling
  Lan}, \bibinfo{person}{Chang Liu}, \bibinfo{person}{Yidong Ouyang},
  \bibinfo{person}{Wenjun Zeng}, {and} \bibinfo{person}{Tao Qin}.}
  \bibinfo{year}{2021}\natexlab{}.
\newblock \showarticletitle{Generalizing to Unseen Domains: A Survey on Domain
  Generalization}. In \bibinfo{booktitle}{\emph{Proceedings of IJCAI 2021}}.
\newblock


\bibitem[\protect\citeauthoryear{Webster, Costa-juss{\`a}, Hardmeier, and
  Radford}{Webster et~al\mbox{.}}{2019}]%
        {webster2019gendered}
\bibfield{author}{\bibinfo{person}{Kellie Webster}, \bibinfo{person}{Marta~R
  Costa-juss{\`a}}, \bibinfo{person}{Christian Hardmeier}, {and}
  \bibinfo{person}{Will Radford}.} \bibinfo{year}{2019}\natexlab{}.
\newblock \showarticletitle{Gendered ambiguous pronoun (GAP) shared task at the
  Gender Bias in NLP Workshop 2019}. In \bibinfo{booktitle}{\emph{Proceedings
  of the First Workshop on Gender Bias in Natural Language Processing}}.
  \bibinfo{pages}{1--7}.
\newblock


\bibitem[\protect\citeauthoryear{West, Whittaker, and Crawford}{West
  et~al\mbox{.}}{2019}]%
        {west2019discriminating}
\bibfield{author}{\bibinfo{person}{Sarah~Myers West}, \bibinfo{person}{Meredith
  Whittaker}, {and} \bibinfo{person}{Kate Crawford}.}
  \bibinfo{year}{2019}\natexlab{}.
\newblock \showarticletitle{Discriminating systems}.
\newblock \bibinfo{journal}{\emph{AI Now}} (\bibinfo{year}{2019}).
\newblock


\bibitem[\protect\citeauthoryear{Winkens, Bunel, Guha~Roy, Stanforth,
  Natarajan, Ledsam, MacWilliams, Kohli, Karthikesalingam, Kohl,
  et~al\mbox{.}}{Winkens et~al\mbox{.}}{2020}]%
        {winkens2020contrastive}
\bibfield{author}{\bibinfo{person}{Jim Winkens}, \bibinfo{person}{Rudy Bunel},
  \bibinfo{person}{Abhijit Guha~Roy}, \bibinfo{person}{Robert Stanforth},
  \bibinfo{person}{Vivek Natarajan}, \bibinfo{person}{Joseph~R Ledsam},
  \bibinfo{person}{Patricia MacWilliams}, \bibinfo{person}{Pushmeet Kohli},
  \bibinfo{person}{Alan Karthikesalingam}, \bibinfo{person}{Simon Kohl},
  {et~al\mbox{.}}} \bibinfo{year}{2020}\natexlab{}.
\newblock \showarticletitle{Contrastive Training for Improved
  Out-of-Distribution Detection}.
\newblock \bibinfo{journal}{\emph{arXiv e-prints}} (\bibinfo{year}{2020}),
  \bibinfo{pages}{arXiv--2007}.
\newblock


\bibitem[\protect\citeauthoryear{Wu, Gao, Guo, Al-Halah, Rennie, Grauman, and
  Feris}{Wu et~al\mbox{.}}{2021}]%
        {wu2021fashion}
\bibfield{author}{\bibinfo{person}{Hui Wu}, \bibinfo{person}{Yupeng Gao},
  \bibinfo{person}{Xiaoxiao Guo}, \bibinfo{person}{Ziad Al-Halah},
  \bibinfo{person}{Steven Rennie}, \bibinfo{person}{Kristen Grauman}, {and}
  \bibinfo{person}{Rogerio Feris}.} \bibinfo{year}{2021}\natexlab{}.
\newblock \showarticletitle{Fashion iq: A new dataset towards retrieving images
  by natural language feedback}. In \bibinfo{booktitle}{\emph{Proceedings of
  the IEEE/CVF Conference on Computer Vision and Pattern Recognition}}.
  \bibinfo{pages}{11307--11317}.
\newblock


\bibitem[\protect\citeauthoryear{Yeom, Giacomelli, Fredrikson, and Jha}{Yeom
  et~al\mbox{.}}{2018}]%
        {yeom2018privacy}
\bibfield{author}{\bibinfo{person}{Samuel Yeom}, \bibinfo{person}{Irene
  Giacomelli}, \bibinfo{person}{Matt Fredrikson}, {and} \bibinfo{person}{Somesh
  Jha}.} \bibinfo{year}{2018}\natexlab{}.
\newblock \showarticletitle{Privacy risk in machine learning: Analyzing the
  connection to overfitting}. In \bibinfo{booktitle}{\emph{2018 IEEE 31st
  Computer Security Foundations Symposium (CSF)}}. IEEE,
  \bibinfo{pages}{268--282}.
\newblock


\bibitem[\protect\citeauthoryear{Zeng, Qi, Zhou, Zhang, Ma, Hou, Zang, Liu, and
  Sun}{Zeng et~al\mbox{.}}{2021}]%
        {zeng2021openattack}
\bibfield{author}{\bibinfo{person}{Guoyang Zeng}, \bibinfo{person}{Fanchao Qi},
  \bibinfo{person}{Qianrui Zhou}, \bibinfo{person}{Tingji Zhang},
  \bibinfo{person}{Zixian Ma}, \bibinfo{person}{Bairu Hou},
  \bibinfo{person}{Yuan Zang}, \bibinfo{person}{Zhiyuan Liu}, {and}
  \bibinfo{person}{Maosong Sun}.} \bibinfo{year}{2021}\natexlab{}.
\newblock \showarticletitle{{OpenAttack}: An Open-source Textual Adversarial
  Attack Toolkit}. In \bibinfo{booktitle}{\emph{Proceedings of the 59th Annual
  Meeting of the Association for Computational Linguistics and the 11th
  International Joint Conference on Natural Language Processing: System
  Demonstrations}}. \bibinfo{pages}{363--371}.
\newblock


\bibitem[\protect\citeauthoryear{Zhang, Harman, Ma, and Liu}{Zhang
  et~al\mbox{.}}{2020a}]%
        {zhang2020machine}
\bibfield{author}{\bibinfo{person}{Jie~M Zhang}, \bibinfo{person}{Mark Harman},
  \bibinfo{person}{Lei Ma}, {and} \bibinfo{person}{Yang Liu}.}
  \bibinfo{year}{2020}\natexlab{a}.
\newblock \showarticletitle{Machine learning testing: Survey, landscapes and
  horizons}.
\newblock \bibinfo{journal}{\emph{IEEE Transactions on Software Engineering}}
  (\bibinfo{year}{2020}).
\newblock


\bibitem[\protect\citeauthoryear{Zhang, Sheng, Alhazmi, and Li}{Zhang
  et~al\mbox{.}}{2020b}]%
        {zhang2020adversarial}
\bibfield{author}{\bibinfo{person}{Wei~Emma Zhang}, \bibinfo{person}{Quan~Z
  Sheng}, \bibinfo{person}{Ahoud Alhazmi}, {and} \bibinfo{person}{Chenliang
  Li}.} \bibinfo{year}{2020}\natexlab{b}.
\newblock \showarticletitle{Adversarial attacks on deep-learning models in
  natural language processing: A survey}.
\newblock \bibinfo{journal}{\emph{ACM Transactions on Intelligent Systems and
  Technology (TIST)}} \bibinfo{volume}{11}, \bibinfo{number}{3}
  (\bibinfo{year}{2020}), \bibinfo{pages}{1--41}.
\newblock


\bibitem[\protect\citeauthoryear{Zhao, Kaafar, and Kourtellis}{Zhao
  et~al\mbox{.}}{2020}]%
        {zhao2020not}
\bibfield{author}{\bibinfo{person}{Benjamin Zi~Hao Zhao},
  \bibinfo{person}{Mohamed~Ali Kaafar}, {and} \bibinfo{person}{Nicolas
  Kourtellis}.} \bibinfo{year}{2020}\natexlab{}.
\newblock \showarticletitle{Not one but many tradeoffs: Privacy vs. utility in
  differentially private machine learning}. In
  \bibinfo{booktitle}{\emph{Proceedings of the 2020 ACM SIGSAC Conference on
  Cloud Computing Security Workshop}}. \bibinfo{pages}{15--26}.
\newblock


\bibitem[\protect\citeauthoryear{Zhao, Wang, Yatskar, Ordonez, and Chang}{Zhao
  et~al\mbox{.}}{2018}]%
        {zhao2018gender}
\bibfield{author}{\bibinfo{person}{Jieyu Zhao}, \bibinfo{person}{Tianlu Wang},
  \bibinfo{person}{Mark Yatskar}, \bibinfo{person}{Vicente Ordonez}, {and}
  \bibinfo{person}{Kai-Wei Chang}.} \bibinfo{year}{2018}\natexlab{}.
\newblock \showarticletitle{Gender Bias in Coreference Resolution: Evaluation
  and Debiasing Methods}. In \bibinfo{booktitle}{\emph{Proceedings of the 2018
  Conference of the North American Chapter of the Association for Computational
  Linguistics: Human Language Technologies}}, Vol.~\bibinfo{volume}{2}.
\newblock


\end{thebibliography}

\newpage

\section*{Appendix A: Metrics in ML Model Evaluations}

Here we give definitions and categorizations of some of the metrics reported in the study in Section~\ref{sec:practice}. 
In practice, there was a long tail since many metrics were used in only a single paper. Here we include only the metrics which were most frequently observed in our study.

\begin{longtable}{p{0.18\textwidth}p{0.18\textwidth}p{0.18\textwidth}p{0.36\textwidth}}\toprule
         Metric&Example Task(s)&Metric category&Definition\\   \toprule \endhead
Accuracy& Classification  &\sc Accuracy & A metric that penalizes system predictions that do not agree with the reference data ($\frac{TP+TN}{TP+TN+FP+FN}$).\\
         AUC&  Classification&\sc AUC
         & The area under the curve parameterized by classification threshold $t$, typically with $y$-axis representing recall and $x$-axis representing false positive rate ($\frac{FP}{FP+TN}$).\\
\bleu &Machine translation &\sc Precision & A form of ``$n$-gram precision,'' originally designed for machine translation but also sometimes used for other text generation tasks, which measures whether sequences of words in the system output are also present in the reference texts  \cite{papineni2002bleu}.\\
         \dice &Image segmentation &\classsim& Equivalent to $F_1$ ($\frac{2TP}{2TP+FP+FN}$). More commonly used for medical image segmentation.\\
         Error rate&Classification  &\sc Accuracy & The inverse of accuracy ($1-accuracy=\frac{FP+FN}{TP+TN+FP+FN}$).\\
                  
        $F$ (or $F_1$) &Text classification&\classsim
         & The harmonic mean of recall and precision ($\frac{2PR}{P+R}$), originally developed for information retrieval \cite{van1974foundation} but now widely used in NLP.\\
         $F_{0.5}$& Text classification &\classsim
         & A weighted harmonic mean of recall and precision, with greater weight given to recall ($(1+\beta^2)\frac{PR}{\beta^2 P+R}$ with $\beta=0.5$).\\
         Hausdorff distance &Medical Image Segmentation &\sc Distance & A measure of distance between two sets in a metric space. Two sets have a low Hausdorff distance if every point in each set is close to a point in the other set. \\
         
         IoU &Image segmentation   &\classsim& $\frac{TP}{TP+FP+FN}$. Equivalent to Jaccard. \\
         Matthew's Correlation Coefficient & & \sc Correlation & Has been argued to address shortcomings in $F_1$'s asymmetry with respect to classes ($\frac{TP.TN - FP.FN}{\sqrt{(TP+TN)(TP+FN)(TN+FN)(TN+FP)}}$). \\
         Mean absolute error &Regression &\sc Distance & $\frac{1}{N}\sum_{i=1}^{N}|\hat{y}_i-y_i|$\\

         Mean Average Precision (MAP)&Information retrieval (NLP)& \sc AUC &
         In information retrieval, the average over information needs of the average precision of the documents retrieved for that need.\\
         Mean average precision (mAP)&Object detection (CV)  &\sc AUC
         & The area under the Precision-Recall tradeoff curve, averaged over multiple IoU (intersection over union) threshold values, then averaged across all categories (\url{https://cocodataset.org/#detection-eval}).\\
Mean reciprocal rank &Information retrieval &\sc Other & A measure for evaluating processes that produces an ordered list of possible responses. The average of the inverse rank of the first relevant item retrieved.  \\

         MSE &Image Decomposition & \sc Distance & Mean squared error (MSE) measures the average of the squared difference between estimated and actual values.\\
         Normalized Discounted Cumulative Gain (NDCG) & Recommendation or ranking tasks & Other &
         A measure of ranking quality which takes into account the usefulness of items based on their ranking in the result list.\\

         Pearson's $r$ &Quality Estimation &\sc Correlation & A measure of linear correlation between two sets of data. \\

         Perplexity&Language modeling  &\sc Perplexity & Information-theoretic metric (measured in bits-per-unit, e.g., bits-per-character or bits-per-sentence) often used for language models, inversely related to the probability assigned to the test data by the model. Closely related to the cross-entropy between the model and the test data. Can be thought of as how efficiently does the language model encode the test data.\\

         Precision&Classification &\sc Precision & A metric that penalizes the system for predicting a class (if class is unspecified, by default the ``positive'' class) when the reference data did not belong to this class ($\frac{TP}{TP+FP}$).\\
         
         PSNR & Super Resolution&\sc Distance & Peak Signal-to-Noise ratio (PSNR) is the ratio between the maximum possible value of a signal and the power of distorting noise (Mean Squared Error) that impacts the quality of its representation. \\
         Recall& Classification &\sc Recall & Also known as ``sensitivity", this metric that penalizes the system for failing to predict a class (if class is unspecified, by default the ``positive'' class) when the reference data did belong to this class ($\frac{TP}{TP+FN}$); a.k.a.\ true positive rate.\\
RMSE &Depth Estimation &\sc Distance & Root Mean Square Error (RMSE) is the square root of the MSE. \\
         
         \rouge &Text summarization &\sc Recall & A form of ``$n$-gram recall,'' originally designed for text summarization but also sometimes used for other text generation tasks, which measures whether sequences of words in the reference texts are also present in the system output\cite{lin2004rouge}.\\

Spearman's $\rho$ &Graph Edit Distance &\sc Correlation &  A measure of monotonic association between two variables--less restrictive than linear correlations.\\

        Specificity& Classification &\sc Other & Like Precision, this metric that penalizes the system for failing to predict a class (if class is unspecified, by default the ``positive'' class) when the reference data did belong to this class; unlike Precision it rewards true negatives rather than true positives ($\frac{TN}{TN+FN}$).\\
        
         SSIM & Super Resolution &\sc Distance & The Structural Similarity Method (SSIM) is a perception-based method for measuring the similarity between two images. The formula is based on comparison measurements of luminance, contrast, and structure.\\
         
        Top-$n$ accuracy&Face recognition  &\sc Accuracy & A metric for systems that return ranked lists, which calculates accuracy over the top $n$ entries in each list.\\

        Word error rate&Speech recognition  &\sc Accuracy
         & The inverse of word accuracy: $1-word\ accuracy$ (which is not technically always in $[0, 1]$ due to the way \textit{word accuracy} is defined but which is categorized as ``Accuracy'' here because both insertions and deletions are penalized).\\

    \caption{Definitions and categorizations of metrics reported in Section~\ref{sec:practice}. TP, TN, FP and FN indicate the number of true positives, true negatives, false positives and false negatives, respectively. $y$ and $\hat{y}$ represent actual values and values predicted by the system, respectively.}
    \label{tab:all_metrics}
\end{longtable}

\newpage
\section*{Appendix B: Types of Evaluation Data used in ML Model Evaluations}

\begin{longtable}{p{0.18\textwidth}p{0.18\textwidth}p{0.18\textwidth}p{0.36\textwidth}}\toprule
    Type of Test Data &Example Task(s)&\iid with training data?&Definition  \\\toprule
    Test split      &Classification  &yes  &Typically, labeled data is partitioned into training and test splits (and often a dev split too), drawn randomly from the same dataset.\\
    Manual resource &Lexical acquisition  &no  &A manually compiled resource (in NLP, often a word-based resource such as a lexicon or thesaurus), against which knowledge acquired from a dataset is compared.\\
    References &Machine translation  &no  &Reference outputs (typically obtained prior to building the system) which a generative system is trying to reproduce, typically obtained from humans (e.g., manual translations of input sentences in the case of evaluations using \bleu for machine translation tasks).\\
    Training data&Keyword extraction &yes  &Training data that contains labels is used to evaluate an unsupervised algorithm that did not have access to the labels during learning.\\
    Novel distribution&Domain transfer &no  &Test data that has the same form as the training data but is drawn from a different distribution (e.g., in the case of NLP training on labeled newspaper data and testing on labeled Wikipedia data).\\\bottomrule    
    
    \caption{Types of datasets used in ML model evaluations.}
    \label{tab:all_datasets}
\end{longtable}

\newpage
\section*{Appendix C: Example of Assumptions and Gaps for a Hypothetical Application}

Suppose we are evaluating a hypothetical image classification model for use in an application for assisting blind people in identifying groceries in their pantries. Then some application-specific questions related to the assumptions in Section~\ref{sec:gaps} might be:

    \subsubsection*{Consequentialism}      Was data ethically sourced and labeled? Were blind people involved in the design process? Does this use of this model encourage high-risk uses of other similar models, such as identifying pharmaceutical products?
    \subsubsection*{Abstractability from Context} Does the application have a human-in-the-loop feature available when the model is uncertain? Will the system nudge purchasing behaviors towards products on which the model performs well?
    \subsubsection*{Input Myopia}   Are uncommon grocery products misclassified more often? Does this disproportionately impact home cooks who don't stick to the dominant cuisines, or who have food requirements due to medical conditions?
    \subsubsection*{Quantitative Modeling}        Does measuring predictive accuracy fail to take into account dignitary consequences associated with increased independence? Should each user be weighted equally in the evaluation (cf.\ each image)? 
    \subsubsection*{Equivalent Failures}          Are there severe risks in confusing certain pairs of products, e.g., food products associated with dangerous allergies? Are some errors only minimally inconvenient, such as confusing different shapes of pasta?
    \subsubsection*{Test Data Validity}              Is the evaluation data representative of what the application's users have in their pantries? Are the image qualities (lighting, focus, framing, etc.) representative of images taken by blind photographers?

\newpage
\section*{Appendix D: Model Evaluation Remits and Design}

\begin{table}[H]
\centering
    \begin{tabular}{p{\textwidth}}\toprule
    \textsc{Model Evaluation Remit}\\
    \quad To establish:\\
    \qquad motivation --- why evaluate the model? \\
    \quad \qquad what is the perspective being adopted --- task/financial/administrative/scientific/... \\
    \quad \qquad whose interests prompted the evaluation --- developer/funder/... \\
    \quad \qquad who are the consumers of the model evaluation results --- manager/user/researcher/... \\
    \qquad goal --- what do we want/need to discover? \\
    \qquad orientation --- intrinsic/extrinsic \\
    \qquad kind --- investigation/experiment \\
    \qquad type --- black box/glass box \\
    \qquad form (of yardstick) --- ideal/attainable/exemplar/given/judged \\
    \qquad style --- suggestive/indicative/exhaustive \\
    \qquad mode --- quantitative/qualitative/hybrid \\
    \midrule
    \textsc{Model Evaluation Design}\\
    \quad To identify:\\
    \qquad ends --- what is the model for? what is its objective or function? \\
    \qquad context --- what is the ecosystem the model is in? what are the animate and inanimate factors? \\
    \qquad constitution --- what is the structure of the model? what was the training data? \\  
    \quad To determine: \\
    \qquad factors that will be tested \\
    \quad \qquad environment variables \\
    \quad \qquad `system' parameters \\
    \qquad evaluation criteria \\
    \quad \qquad metrics/measures \\
    \quad \qquad methods \\
    \quad Evaluation data  --- what type, status and nature? \\
    \quad Evaluation procedure \\
    \bottomrule
    \end{tabular}
    \caption{A sketch of how Karen Sparck Jones and Julia Galliers' 1995 NLP evaluation framework questionnaire \cite{jones1995evaluating} can be adapted for the evaluation of ML models. The output of the remit and the design is a strategy for conducting the model evaluation. For a related but simpler framework based on model requirements analysis, see also the ``7-step Recipe'' for NLP system evaluation (\url{https://www.issco.unige.ch/en/research/projects/eagles/ewg99/7steps.html}) developed by the \textsc{eagles} Evaluation Working Group in 1999, which considers whether different parties have a shared understanding of the evaluation's purpose.}
    \label{tab:eval framework}
\end{table}

\end{document}